\documentclass[10pt,twocolumn,letterpaper]{article}

\usepackage[pagenumbers]{cvpr} % To force page numbers, e.g. for an arXiv version

\usepackage{graphicx}
\usepackage{amsmath}
\usepackage{amssymb}
\usepackage{booktabs}
\usepackage{multirow}
\usepackage[pagebackref,breaklinks,colorlinks]{hyperref}
\usepackage[capitalize]{cleveref}
\crefname{section}{Sec.}{Secs.}
\Crefname{section}{Section}{Sections}
\Crefname{table}{Table}{Tables}
\crefname{table}{Tab.}{Tabs.}

\def\cvprPaperID{4528}
\def\confName{CVPR}
\def\confYear{2022}

\newcommand\blfootnote[1]{%	
  \begingroup
  \renewcommand\thefootnote{}\footnote{#1}%
  \addtocounter{footnote}{-1}%
  \endgroup
}

\begin{document}

\title{CamLiFlow: Bidirectional Camera-LiDAR Fusion for Joint Optical Flow and Scene Flow Estimation}

\author{
Haisong Liu* \quad Tao Lu \quad Yihui Xu \quad Jia Liu \quad Wenjie Li \quad Lijun Chen\\
State Key Laboratory for Novel Software Technology, Nanjing University, China\\
\small\texttt{\{liuhs, taolu, yhxu, wenjielee\}@smail.nju.edu.cn, \{jialiu, chenlj\}@nju.edu.cn}
}


\maketitle

\begin{abstract}
In this paper, we study the problem of jointly estimating the optical flow and scene flow from synchronized 2D and 3D data. Previous methods either employ a complex pipeline that splits the joint task into independent stages, or fuse 2D and 3D information in an ``early-fusion'' or ``late-fusion'' manner. Such one-size-fits-all approaches suffer from a dilemma of failing to fully utilize the characteristic of each modality or to maximize the inter-modality complementarity. To address the problem, we propose a novel end-to-end framework, called CamLiFlow. It consists of 2D and 3D branches with multiple bidirectional connections between them in specific layers. Different from previous work, we apply a point-based 3D branch to better extract the geometric features and design a symmetric learnable operator to fuse dense image features and sparse point features. Experiments show that CamLiFlow achieves better performance with fewer parameters. Our method ranks 1st on the KITTI Scene Flow benchmark, outperforming the previous art with 1/7 parameters. Code is available at \url{https://github.com/MCG-NJU/CamLiFlow}.
\end{abstract}

\blfootnote{* Corresponding author (liuhs@smail.nju.edu.cn).}

\vspace{-20pt}

\section{Introduction}

\begin{figure}[t]
    \hspace*{0.45cm}
    \includegraphics[width=0.79\linewidth]{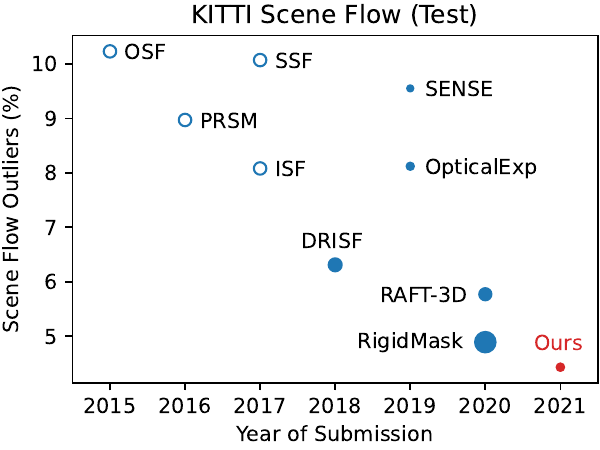}
    \vspace{-5pt}
    \caption{Results of the KITTI Scene Flow Benchmark. Marker size indicates model size. Models with unknown sizes and conventional approaches are marked as hollow. Our method outperforms all existing approaches \cite{menze2015osf,ren2017ssf,vogel2015prsm,behl2017isf,ma2019drisf,jiang2019sense,yang2020opticalexp,teed2021raft3d,yang2021rigidmask} with much fewer parameters.}
    \vspace{-16pt}
    \label{fig:kitti-leaderboard}
\end{figure}

Optical flow and scene flow are the motion field in 2D and 3D space respectively. Through them, we can gain insights into the dynamics of the scene, which are critical to some high-level scene understanding tasks. In this work, we focus on the joint estimation of optical flow and scene flow, which addresses monocular camera frames with sparse depth measurements from LiDAR.

Previous methods \cite{yang2021rigidmask, ma2019drisf, yang2020opticalexp, behl2017isf} construct a modular network that decomposes the estimation of flow into multiple subtasks. These submodules are independent of each other, making it impossible for utilizing their complementarity. Moreover, the limitations of any submodule will hurt the overall performance, since the whole pipeline depends on its results.

\begin{figure}[t]
    \centering
    \begin{subfigure}[b]{0.49\linewidth}
        \includegraphics[width=\linewidth]{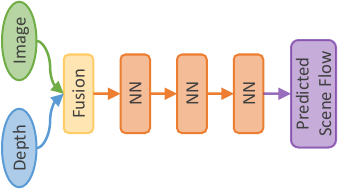}%
        \caption{Early Fusion}
        \label{fig:fusion-early}
    \end{subfigure}%
    \hfill
    \begin{subfigure}[b]{0.49\linewidth}
        \includegraphics[width=\linewidth]{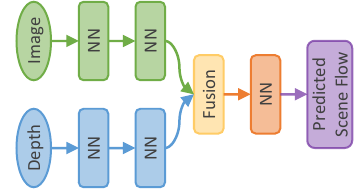}%
        \caption{Late Fusion}
        \label{fig:fusion-late}
    \end{subfigure}%
    \hfill
    \begin{subfigure}[b]{0.98\linewidth}
        \includegraphics[width=\linewidth]{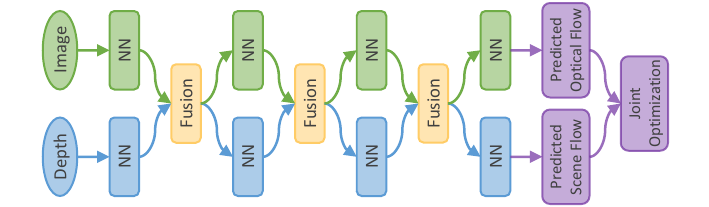}%
        \caption{Our Fusion}
        \label{fig:fusion-ours}
    \end{subfigure}

    \vspace{-5pt}
    \caption{Architectures for feature-level fusion. Different from previous work which adopt an early/late fusion manner, we propose a multi-stage and bidirectional fusion pipeline.}
    \vspace{-12pt}
\end{figure}

Other methods \cite{teed2021raft3d, rishav2020deeplidarflow} build an end-to-end architecture, consisting of multiple stages (including feature extraction, computing correlation, feature decoding, etc). RAFT-3D \cite{teed2021raft3d} concatenates images and dense depth maps to RGB-D frames and feeds them into a unified 2D network to predict pixel-wise 3D motion. This kind of ``early fusion'' (Fig. \ref{fig:fusion-early}) makes 2D CNNs fail to use most 3D structural information provided by depth. DeepLiDARFlow \cite{rishav2020deeplidarflow} takes images and LiDAR point clouds as input, where points are projected onto the image plane for densification and are fused with images in a ``late-fusion'' manner (Fig. \ref{fig:fusion-late}). However, some errors generated in the early stage lack the opportunity to be corrected by the other modality and are accumulated to the subsequent stages. Thus the complementarity between the two modalities is not fully exploited.

In general, single-stage fusion suffers from a dilemma of failing to fully utilize the characteristic of each modality or to maximize the inter-modality complementarity. To address the problem, we propose a \textbf{multi-stage} and \textbf{bidirectional} fusion pipeline (see Fig. \ref{fig:fusion-ours}), which achieves better performance with fewer parameters. Within each stage, the two modalities are learned in separate branches using modality-specific architecture. At the end of each stage, a learnable bidirectional bridge connects the two branches to pass complementary information. Moreover, recent point-based methods \cite{qi2017pointnet, qi2017pointnet++, wu2019pointconv, liu2019flownet3d, wang2020flownet3d++, wu2019pointpwc, qi2018frustum, liu2019meteornet} achieve remarkable progress for 3D computer vision. This inspires us to process point clouds with a point-based branch, which can extract the fine 3D geometric information without any voxelization or projection.

It is worth noting that there are two challenges for the fusion of the image branch and the point branch. First, image features are organized in a dense grid structure, while point clouds do not conform to the regular grid and are sparsely distributed in the continuous domain. As a result, there is no guarantee of one-to-one correspondence between pixels and points. Second, LiDAR point cloud possesses the property of varying density, where nearby regions have much greater density than farther-away regions. To tackle the first problem, we propose a new learnable fusion operator, named \textit{bidirectional camera-LiDAR fusion module} (Bi-CLFM), which fuses image/point features for both directions via learnable interpolation and sampling. As for the second problem, we propose a new transformation operator, named \textit{inverse depth scaling} (IDS), which balances the distribution of points by scaling them non-linearly according to the inverse depth.

Experiments demonstrate that our approach achieves better performance with much fewer parameters. On FlyingThings3D \cite{mayer2016things3d}, we achieve up to a {\bf48.4\%} reduction in end-point-error over RAFT-3D with only 1/6 parameters. On KITTI \cite{menze2015osf}, CamLiFlow achieves an error of {\bf4.43\%}, outperforming the previous art \cite{yang2021rigidmask} with only 1/7 parameters. The leaderboard is shown in Fig. \ref{fig:kitti-leaderboard}.

\section{Related Work}

\paragraph{Optical Flow.} Optical flow estimation aims to predict dense 2D motion for each pixel from a pair of frames. Traditional methods \cite{horn1981determining, black1996robust, brox2009large, weinzaepfel2013deepflow, brox2004warping, bruhn2005lucas} often formulate optical flow as an energy minimization problem. FlowNet \cite{dosovitskiy2015flownet} is the first end-to-end trainable CNN for optical flow estimation, which adopts an encoder-decoder architecture. FlowNet2 \cite{ilg2017flownet2} stacks several FlowNets into a larger one. PWC-Net \cite{sun2018pwc} and some other methods \cite{ranjan2017spynet, hur2019iterative, hui2018liteflownet, yang2019volumetric} apply iterative refinement using coarse-to-fine pyramids. RAFT \cite{teed2020raft} constructs 4D cost volumes for all pairs of pixels and updates the flow iteratively. Although achieving state-of-the-art performance, RAFT runs much slower than PWC-Net. Hence, our two branches are built upon the PWC architecture to achieve a better trade-off between accuracy and speed.

\vspace{-10pt}
\paragraph{Scene Flow from RGB-D Frames.} RGB-D scene flow is the problem of estimating dense 3D motion for each pixel from a pair of stereo or RGB-D frames. Like optical flow, traditional methods \cite{menze2015osf, quiroga2014dense, jaimez2015primal, jaimez2015motion} explore variational optimization and discrete optimization and treat scene flow as an energy minimization problem. Recent methods \cite{behl2017isf, ma2019drisf, yang2020opticalexp, yang2021rigidmask} divide scene flow estimation into multiple subtasks and build a modular network with one or more submodules for each subtask. Although achieving remarkable progress, their submodules are independent of each other, which can not exploit the complementary characteristics of different modalities. RAFT-3D \cite{teed2021raft3d} concatenate images and depth maps to RGB-D frames at an early stage, followed by a unified 2D network which iteratively updates a dense field of pixel-wise SE3 motion. However, this kind of ``early fusion'' makes it hard for 2D CNNs to take advantage of the rich 3D structural information.

\vspace{-10pt}
\paragraph{Scene Flow from Point Clouds.} Recently, researchers start to study scene flow estimation in 3D point clouds (e.g. from LiDAR) \cite{liu2019flownet3d, wang2020flownet3d++, gu2019hplflownet, wu2019pointpwc, puy2020flot, liu2019meteornet, wei2021pvraft, kittenplon2021flowstep3d}. Based on \cite{qi2017pointnet++}, FlowNet3D \cite{liu2019flownet3d} uses a flow embedding layer to represent the motion of points. FlowNet3D++ \cite{wang2020flownet3d++} achieves better performance by adding geometric constraints. Inspired by Bilateral Convolutional Layers, HPLFlowNet \cite{gu2019hplflownet} projects the points onto a permutohedral lattice. PointPWC-Net \cite{wu2019pointpwc} introduces a learnable cost volume for point clouds and estimates scene flow in a coarse-to-fine fashion. However, these methods do not exploit color features provided by images. As we demonstrate in our experiments, fusing point clouds with images can bring significant improvements.

\begin{figure*} 
    \centering
    \includegraphics[width=\linewidth]{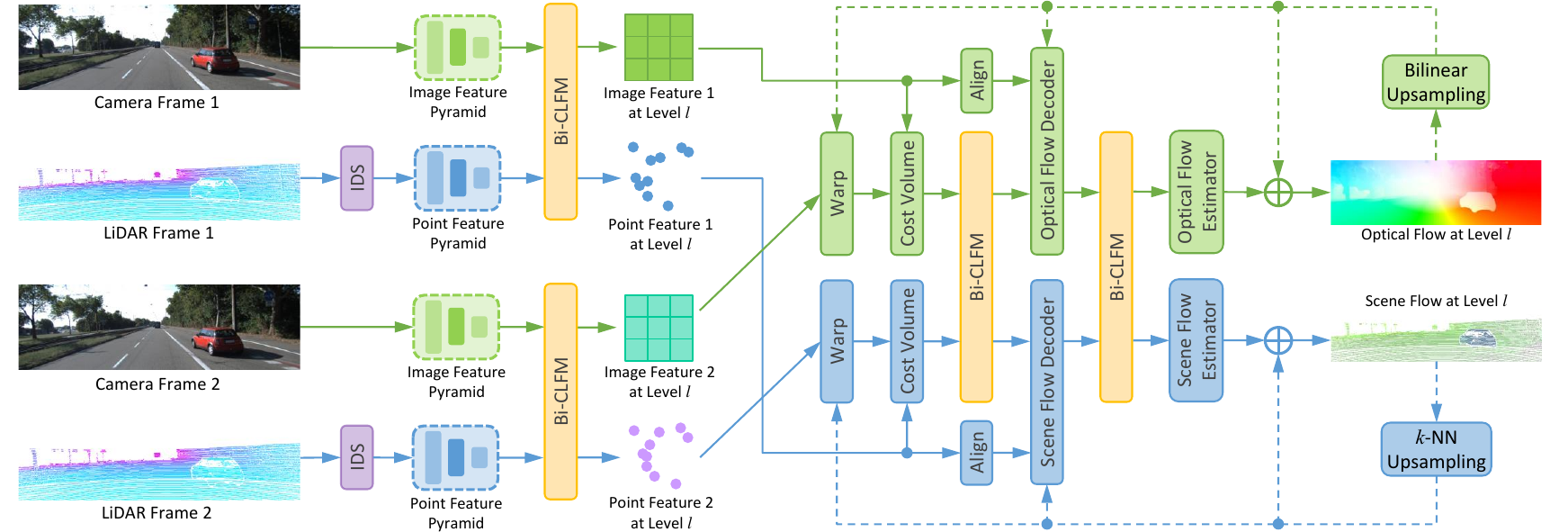}
    \vspace{-15pt}
    \caption{The architecture of CamLiFlow. Synchronized camera and LiDAR frames are taken as input, from which dense optical flow and sparse scene flow are estimated respectively. CamLiFlow is a two-branch network with multiple bidirectional fusion connections (Bi-CLFM) between them. Note that our method can also take advantage of other reliable depth sensors (such as stereo cameras).}
    \vspace{-10pt}
    \label{fig:arch}
\end{figure*}

\vspace{-10pt}
\paragraph{Camera-LiDAR Fusion.} Cameras and LiDARs have complementary characteristics, facilitating many computer vision tasks, such as depth estimation \cite{ma2018sparse, you2019pseudo,feng2021advancing}, scene flow estimation \cite{battrawy2019lidarflow, rishav2020deeplidarflow}, 3D object detection \cite{qi2018frustum, liang2018continuous, xu2018pointfusion, vora2020pointpainting, chen2017multi}, etc. Some researchers \cite{you2019pseudo, battrawy2019lidarflow, qi2018frustum, vora2020pointpainting} build a modular network and perform result-level fusion, while the others \cite{ma2018sparse, feng2021advancing, rishav2020deeplidarflow, liang2018continuous, xu2018pointfusion} explore feature-level fusion schemes including early-fusion and late-fusion. Instead, we propose a multi-stage and bidirectional fusion pipeline, which not only fully utilizes the characteristic of each modality, but maximizes the inter-modality complementarity as well.

\section{CamLiFlow}

Given a pair of the synchronized camera and LiDAR frames, CamLiFlow jointly estimates dense optical flow for camera frames and sparse scene flow for LiDAR frames. As illustrated in Fig. \ref{fig:arch}, CamLiFlow consists of two symmetric branches, named image branch and point branch, for 2D and 3D data respectively. Both branches are built on top of the PWC architecture \cite{sun2018pwc, wu2019pointpwc} where flow computed at the coarse level is upsampled and warped to a finner level. Features are fused in a bidirectional manner at multiple levels and stages.

In the following sections, we first introduce the \textit{bidirectional camera-LiDAR fusion module} along with the multi-stage fusion pipeline. Next, we introduce \textit{inverse depth scaling}, which makes the distribution of points more even across different regions. Finally, a multi-task loss for joint optimization is also introduced.

\subsection{Bidirectional Camera-LiDAR Fusion Module}

\begin{figure}[t]
    \centering
    \includegraphics[width=\linewidth]{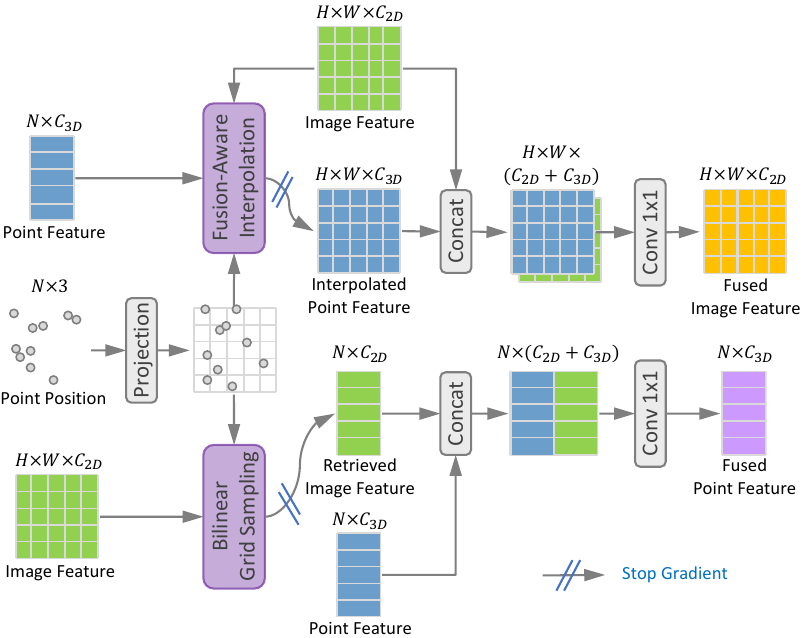}
    \vspace{-18pt}
    \caption{Details of Bidirectional Camera-LiDAR Fusion Module (Bi-CLFM). Features from two different modalities are fused in a bidirectional way, so that both modalities can benefit each other.}
    \vspace{-15pt}
    \label{fig:clfm}
\end{figure}

As mentioned above, the fusion between camera and LiDAR is challenging, since the data structures of image features and point features do not match. To overcome this, we introduce \textit{bidirectional camera-LiDAR fusion module} (Bi-CLFM), which can fuse dense image features and sparse point features in a bidirectional manner. 

As illustrated in Fig. \ref{fig:clfm}, Bi-CLFM takes image features $F \in \mathbb{R}^{H \times W \times C_{2D}}$, point features $G = \{g_i | i = 1, ..., N\} \in \mathbb{R}^{N \times C_{3D}}$ and point positions $P = \{ p_i | i = 1, ..., N \} \in \mathbb{R}^{N \times 3}$ as input, where $N$ denotes the number of points. Features are fused for both directions so that both modalities can benefit each other. Note that we stop the gradient at specific locations to prevent one modality from dominating and stabilize the training (please refer to the supplementary material for more details).

\vspace{-10pt}
\paragraph{2D $\Rightarrow$ 3D.} First, points are projected to the image plane (denoted as $X = \{ x_i | i = 1, ..., N \} \in \mathbb{R}^{N \times 2}$) to retrieve the corresponding 2D feature:
\begin{equation}
    H = \{ F(x_i) | i = 1, ..., N \} \in \mathbb{R}^{N \times C_{2D}},
\end{equation}
where $F(x)$ denotes the image feature at $x$ and can be retrieved by bilinear interpolation if the coordinate is not a integer. Next, the retrieved feature $H$ is concatenated with the input 3D feature $G$. Finally, a $1 \times 1$ convolution is employed to reduce the dimension of the fused 3D feature.

\begin{figure}[t]
    \centering
    \includegraphics[width=\linewidth]{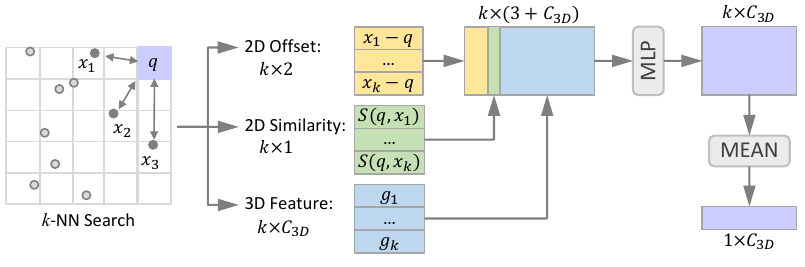}
    \vspace{-20pt}
    \caption{Details of Fusion-Aware Interpolation. For each target pixel, we find the $k$ nearest points around it. A learnable MLP followed by MEAN is employed to aggregate features.}
    \vspace{-10pt}
    \label{fig:fa-interpolation}
\end{figure}

\vspace{-10pt}
\paragraph{3D $\Rightarrow$ 2D.} Similarly, points are first projected to the image plane (denoted as $X = \{ x_i | i = 1, ..., N \} \in \mathbb{R}^{N \times 2}$). Since point clouds are sparse, we propose \textit{fusion-aware interpolation} (detailed in the following paragraphs) to create a dense feature map $D \in \mathbb{R}^{H \times W \times C_{3D}}$ from sparse 3D features. Next, the ``interpolated'' point features are concatenated with the input image features, followed by a $1 \times 1$ convolution to reduce the feature dimension.

\vspace{-10pt}
\paragraph{Fusion-Aware Interpolation.} To solve the problem of fusing sparse point features into dense image features, we propose a learnable \textit{fusion-aware interpolation}. As illustrated in Fig. \ref{fig:fa-interpolation}, for each target pixel $q$ in the dense map, we find its $k$ nearest neighbors among the projected points over the image plane. An MLP followed by \texttt{MEAN} is used to aggregate features, which can be formulated as:
\begin{equation}
    D(q) = \frac{1}{k} \sum_{x_i \in \mathcal{N}_q } \text{MLP}([x_i - q, \mathcal{S}(q, x_i), g_i]),
\end{equation}
where $\mathcal{N}_q$ denotes all the neighborhood points, $g_i$ is the 3D feature of point $i$ and $[\cdot]$ denotes concatenation. The inputs of our MLP also include 2D similarity measurements between the $q$ and its neighbors, which is defined as:
\begin{equation}
    \mathcal{S}(q, x_i) = F(q) \cdot F(x_i).
\end{equation}

Introducing 2D similarity measurements into the interpolation module makes it more robust in complex scenes with overlapping objects, since dense 2D features can be used to guide the densification of the sparse 3D features. We empirically test it in the ablation study (Fig. \ref{fig:ablation-sim2d}).

\subsection{Multi-stage Fusion Pipeline}

In this section, we build a multi-stage and bidirectional fusion pipeline with Bi-CLFM. Our backbone is based on the PWC architecture, which consists of multiple stages including feature extraction, warping, cost volume, and flow estimation. Within each stage, the two modalities are learned in separate branches using modality-specific architecture. At the end of each stage, a Bi-CLFM connects the two branches to pass complementary information.

\vspace{-10pt}
\paragraph{Feature Pyramid.} Given a pair of images and point clouds, we generate a feature pyramid for the image branch and the point branch respectively (the configuration details are included in the supplementary material). For each level $l$, image features are downsampled by a factor of 2 using residual blocks, while points are downsampled by the same factor using \textit{furthest point sampling}, followed by a PointConv \cite{wu2019pointconv} to aggregate features. The image pyramid encodes textural information, while the point pyramid encodes geometric information. Thus, features are fused by a Bi-CLFM at multiple levels for complementarity.

\vspace{-10pt}
\paragraph{Warping.} At each pyramid level $l$, both image features and point clouds are warped towards the reference frame using the upsampled flow from the lower level. Since the warping layer does not introduce any learnable parameters, we do not perform feature fusion after this stage.

\vspace{-10pt}
\paragraph{Cost Volume.} Cost volume stores the matching costs between the reference frame and the warped target frame. For the image branch, we follow \cite{sun2018pwc} to construct a partial cost volume by limiting the search range to 4 pixels around each pixel. For the point branch, we follow \cite{wu2019pointpwc} to construct a learnable cost volume layer. The pixel-based 2D cost volume maintains a fixed range of neighborhoods, while the point-based 3D cost volume searches for a dynamic range. Hence, we fuse the two cost volumes with a Bi-CLFM. 

\vspace{-10pt}
\paragraph{Flow Estimator.} We build a flow estimator for each branch. The input of the flow estimator includes the cost volume, the features of the reference frame, and the upsampled flow. Our optical flow estimator follows \cite{sun2018pwc}, which employs a multi-layer CNN with DenseNet \cite{huang2017densenet} connections. Our scene flow estimator follows \cite{wu2019pointpwc}, which is built as multiple layers of PointConv \cite{wu2019pointconv}. Features from the second last layer of the two estimators are fused. For clarity, we refer to the last layer as the ``flow estimator'' and the other layers as the ``flow decoder'' in Fig. \ref{fig:arch}.

\subsection{Inverse Depth Scaling}
\label{sec:ids}

As mentioned above, the distribution of LiDAR point clouds is not balanced, where nearby region has much greater density than farther-away region. Here, we propose a transformation operator for point clouds to address the problem, named \textit{inverse depth scaling} (IDS). Formally, let $(P_x, P_y, P_z)$ and $(P_x', P_y', P_z')$ be the coordinate of a point before and after the transformation respectively. IDS scales all three dimensions equally by the inverse depth $\frac{1}{P_z}$:
\begin{equation}
    \frac{\delta P_x'}{\delta P_x} = \frac{\delta P_y'}{\delta P_y} = \frac{\delta P_z'}{\delta P_z} = \frac{1}{P_z}.
\end{equation}

The transformed coordinates $(P_x', P_y', P_z')$ can be inferred by integrating the above formula:
\begin{align}
    \label{eq:ids_x}
    P_x' &= \int \frac{1}{P_z} dP_x = \frac{P_x}{P_z} + C_x, \\
    \label{eq:ids_y}
    P_y' &= \int \frac{1}{P_z} dP_y = \frac{P_y}{P_z} + C_y, \\
    \label{eq:ids_z}
    P_z' &= \int \frac{1}{P_z} dP_z = \log{P_z} + C_z,
\end{align}
where both $C_x$ and $C_y$ are set to 0, and $C_z$ is set to 1 to avoid zero depth.

\begin{figure}[t]
    \vspace{-5pt}
    \centering
    \begin{subfigure}[b]{0.5\linewidth}
        \includegraphics[width=\linewidth]{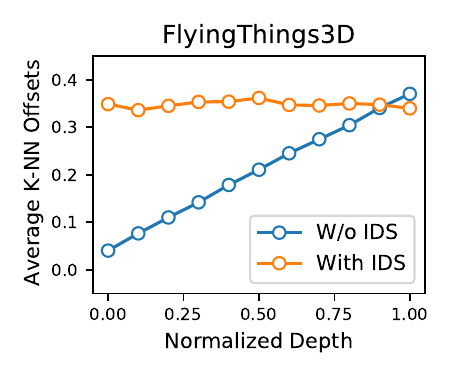}%
    \end{subfigure}%
    \hfill
    \begin{subfigure}[b]{0.5\linewidth}
        \includegraphics[width=\linewidth]{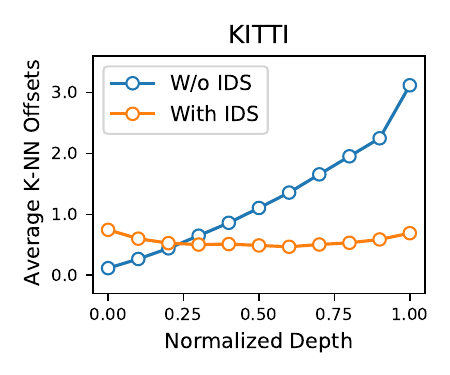}%
    \end{subfigure}%
    \vspace{-10pt}
    \caption{The density of point clouds across different distances with/without IDS. We measure the local density around a point by averaging the offsets of its $k$ nearest neighbors. IDS makes the distribution of points more even across different regions.}
    \vspace{-10pt}
    \label{fig:ids-stat}
\end{figure}

In Fig. \ref{fig:ids-stat}, we perform a statistic on FlyingThings3D and the raw Velodyne data of KITTI to show the density of points across different distances with/without IDS. The local density around a point is measured by averaging the offsets of its $k$ nearest neighbors ($k=16$). As we can see, IDS makes the distribution of points more even across different regions. In this paper, point clouds are transformed by IDS before being sent to the neural network. 

\subsection{Multi-task Loss}

Although the estimation of optical flow and scene flow are highly relevant (the projection of scene flow onto the image plane becomes optical flow), we formulate them as two different tasks. We supervise the 2D and 3D branches respectively and design a multi-task loss for joint optimization. Let $\hat{f}_{2D}^l$ and $\hat{f}_{3D}^l$ be the ground truth optical flow and scene flow at the $l$th level respectively. The regression loss for each branch is defined as follows:
\begin{align}
    \mathcal{L}_{2D} &= \sum_{l=l_0}^{L} \alpha_l \sum_{\mathbf{x}} \Vert f_{2D}^l(\mathbf{x}) - \hat{f}_{2D}^l(\mathbf{x}) \Vert_2,\\
    \mathcal{L}_{3D} &= \sum_{l=l_0}^{L} \alpha_l \sum_{\mathbf{p}} \Vert f_{3D}^l(\mathbf{p}) - \hat{f}_{3D}^l(\mathbf{p}) \Vert_2,
\end{align}
where $\Vert \cdot \Vert_2$ computes the $L_2$ norm. For fine-tuning, we use the following robust training loss:
\begin{align}
    \mathcal{L}_{2D} &= \sum_{l=l_0}^{L} \alpha_l \sum_{\mathbf{x}} (| f_{2D}^l(\mathbf{x}) - \hat{f}_{2D}^l(\mathbf{x}) | + \epsilon)^q,\\
    \mathcal{L}_{3D} &= \sum_{l=l_0}^{L} \alpha_l \sum_{\mathbf{x}} (| f_{3D}^l(\mathbf{x}) - \hat{f}_{3D}^l(\mathbf{x}) | + \epsilon)^q,
\end{align}
where $| \cdot |$ computes the $L_1$ norm, $q=0.4$ gives less penalty to outliers and $\epsilon$ is set to $0.01$. The final loss is a weighted sum of the losses defined above:
\begin{equation}
    \mathcal{L} = \mathcal{L}_{2D} + \lambda \mathcal{L}_{3D},  
\end{equation}
where $\lambda$ is set to 1.0 for all our experiments.

\section{Experiments}

\begin{figure*}
    \captionsetup[subfigure]{labelformat=empty}
    
    % optical flow
    \begin{subfigure}[b]{0.165\linewidth}
        \caption{Reference Frame}
    \end{subfigure}%
    \hfill
    \begin{subfigure}[b]{0.165\linewidth}
        \caption{Ground Truth}
    \end{subfigure}%
    \hfill
    \begin{subfigure}[b]{0.165\linewidth}
        \caption{RAFT}
    \end{subfigure}%
    \hfill
    \begin{subfigure}[b]{0.165\linewidth}
        \caption{RAFT-3D}
    \end{subfigure}%
    \hfill
    \begin{subfigure}[b]{0.165\linewidth}
        \caption{Ours (2D-branch Only)}
    \end{subfigure}%
    \hfill
    \begin{subfigure}[b]{0.165\linewidth}
        \caption{Ours (Full Model)}
    \end{subfigure}%
    \hfill
    
    \vspace{-3pt}
    \begin{subfigure}[b]{0.165\linewidth}
        \includegraphics[width=\linewidth]{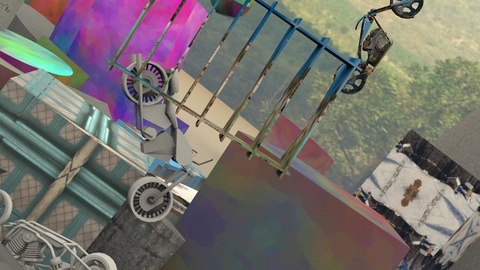}
    \end{subfigure}%
    \hfill
    \begin{subfigure}[b]{0.165\linewidth}
        \includegraphics[width=\linewidth]{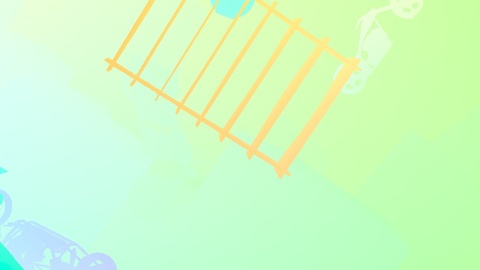}
    \end{subfigure}%
    \hfill
    \begin{subfigure}[b]{0.165\linewidth}
        \includegraphics[width=\linewidth]{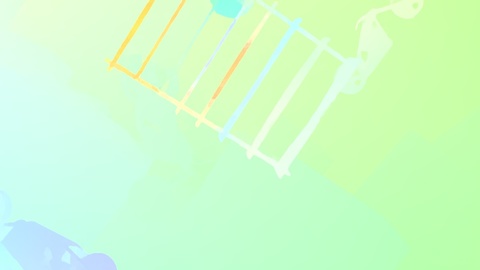}
    \end{subfigure}%
    \hfill
    \begin{subfigure}[b]{0.165\linewidth}
        \includegraphics[width=\linewidth]{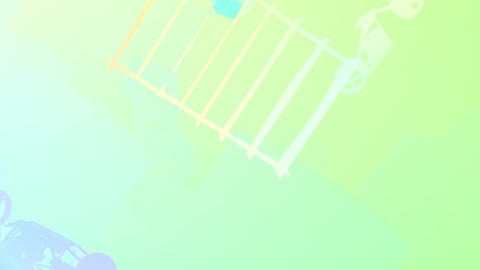}
    \end{subfigure}%
    \hfill
    \begin{subfigure}[b]{0.165\linewidth}
        \includegraphics[width=\linewidth]{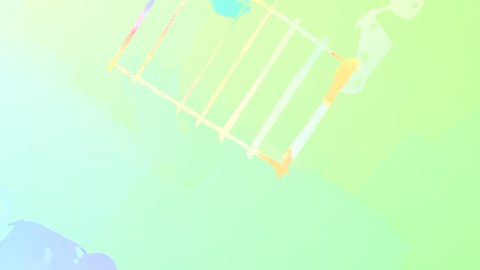}
    \end{subfigure}%
    \hfill
    \begin{subfigure}[b]{0.165\linewidth}
        \includegraphics[width=\linewidth]{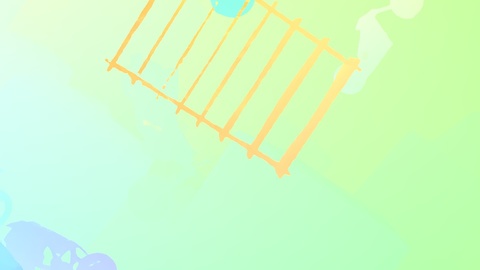}
    \end{subfigure}%
    \hfill

    \begin{subfigure}[b]{0.165\linewidth}
        \includegraphics[width=\linewidth]{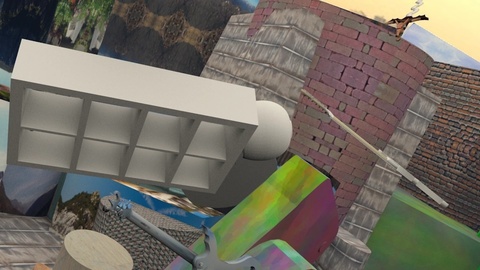}
    \end{subfigure}%
    \hfill
    \begin{subfigure}[b]{0.165\linewidth}
        \includegraphics[width=\linewidth]{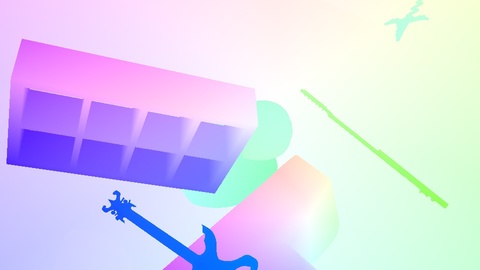}
    \end{subfigure}%
    \hfill
    \begin{subfigure}[b]{0.165\linewidth}
        \includegraphics[width=\linewidth]{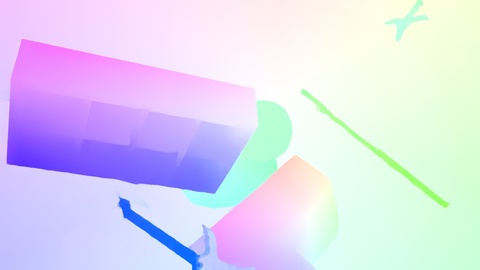}
    \end{subfigure}%
    \hfill
    \begin{subfigure}[b]{0.165\linewidth}
        \includegraphics[width=\linewidth]{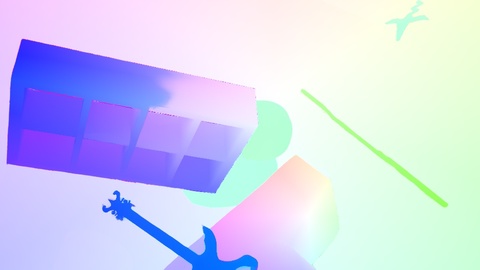}
    \end{subfigure}%
    \hfill
    \begin{subfigure}[b]{0.165\linewidth}
        \includegraphics[width=\linewidth]{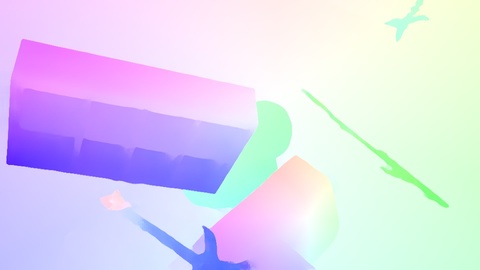}
    \end{subfigure}%
    \hfill
    \begin{subfigure}[b]{0.165\linewidth}
        \includegraphics[width=\linewidth]{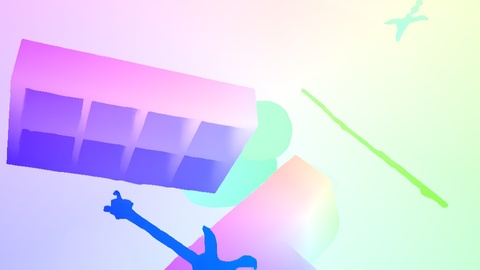}
    \end{subfigure}%
    \hfill
    
    % scene flow
    \vspace{3pt}
    \begin{subfigure}[b]{0.198\linewidth}
        \caption{Input}
    \end{subfigure}%
    \hfill
    \begin{subfigure}[b]{0.198\linewidth}
        \caption{Ground Truth}
    \end{subfigure}%
    \hfill
    \begin{subfigure}[b]{0.198\linewidth}
        \caption{RAFT-3D}
    \end{subfigure}%
    \hfill
    \begin{subfigure}[b]{0.198\linewidth}
        \caption{Ours (3D-branch Only)}
    \end{subfigure}%
    \hfill
    \begin{subfigure}[b]{0.198\linewidth}
        \caption{Ours (Full Model)}
    \end{subfigure}%
    \hfill
    
    \vspace{-3pt}
    \begin{subfigure}[b]{0.198\linewidth}
        \includegraphics[width=\linewidth]{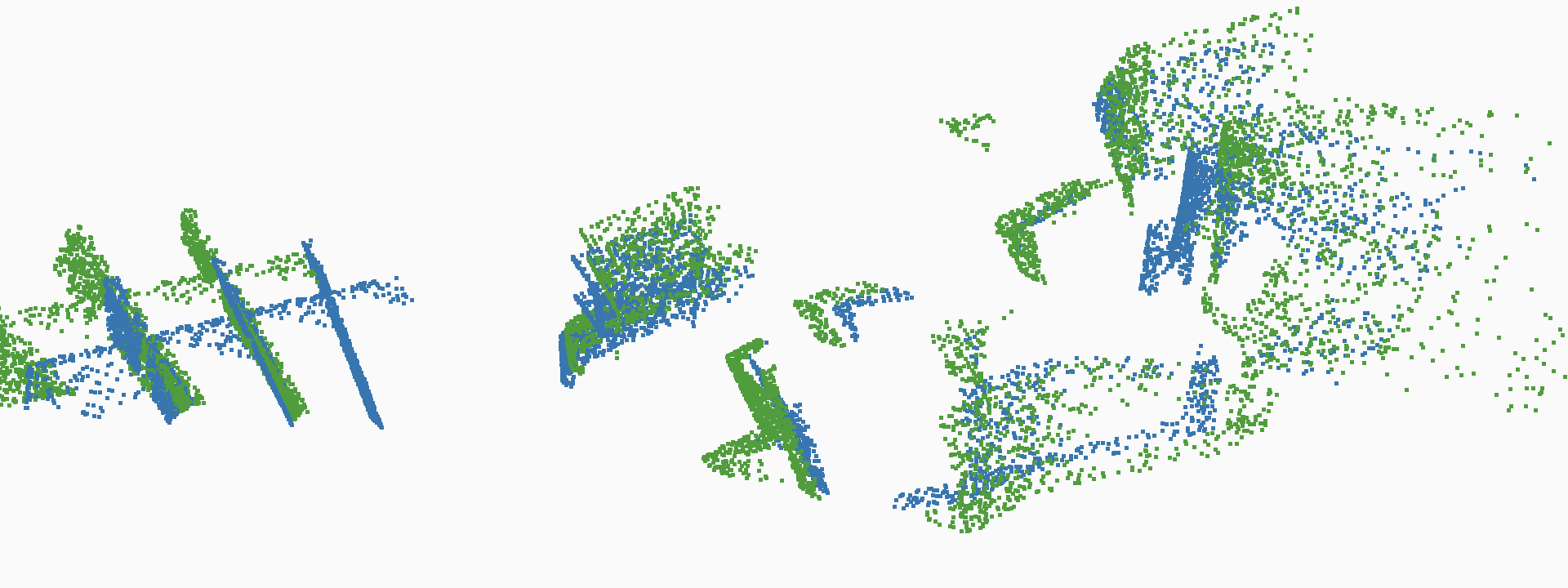}
    \end{subfigure}%
    \hfill
    \begin{subfigure}[b]{0.198\linewidth}
        \includegraphics[width=\linewidth]{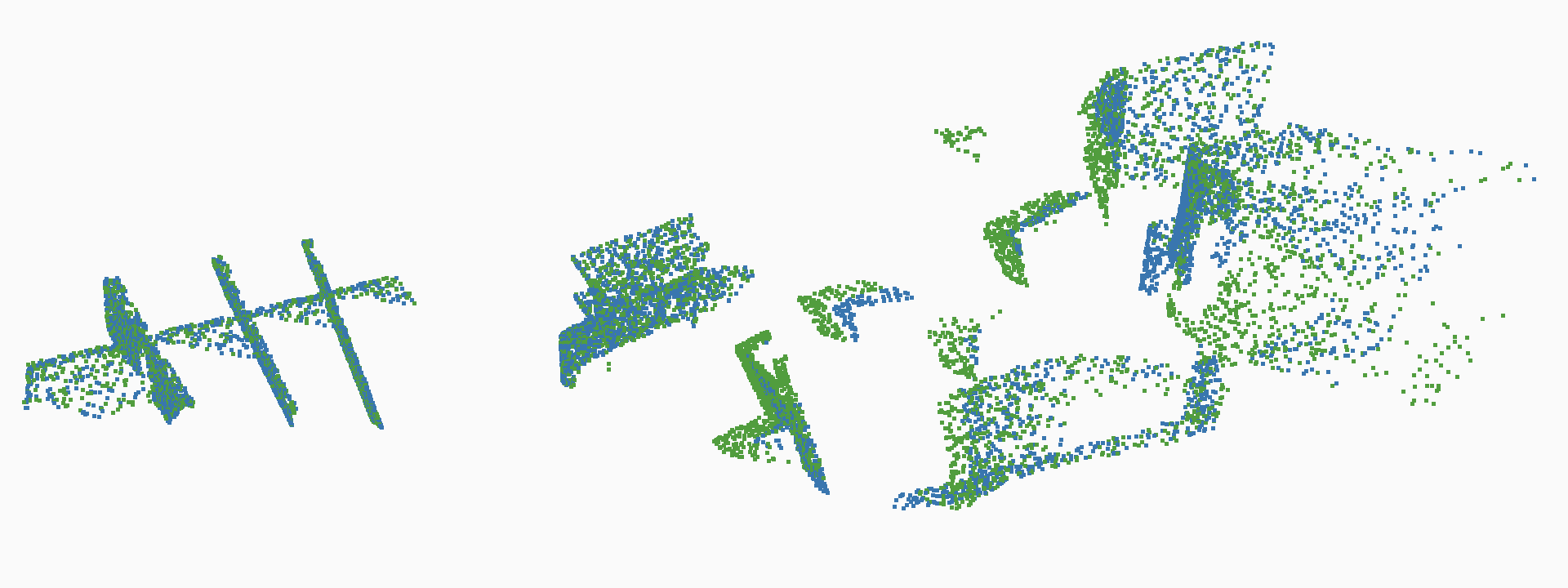}
    \end{subfigure}%
    \hfill
    \begin{subfigure}[b]{0.198\linewidth}
        \includegraphics[width=\linewidth]{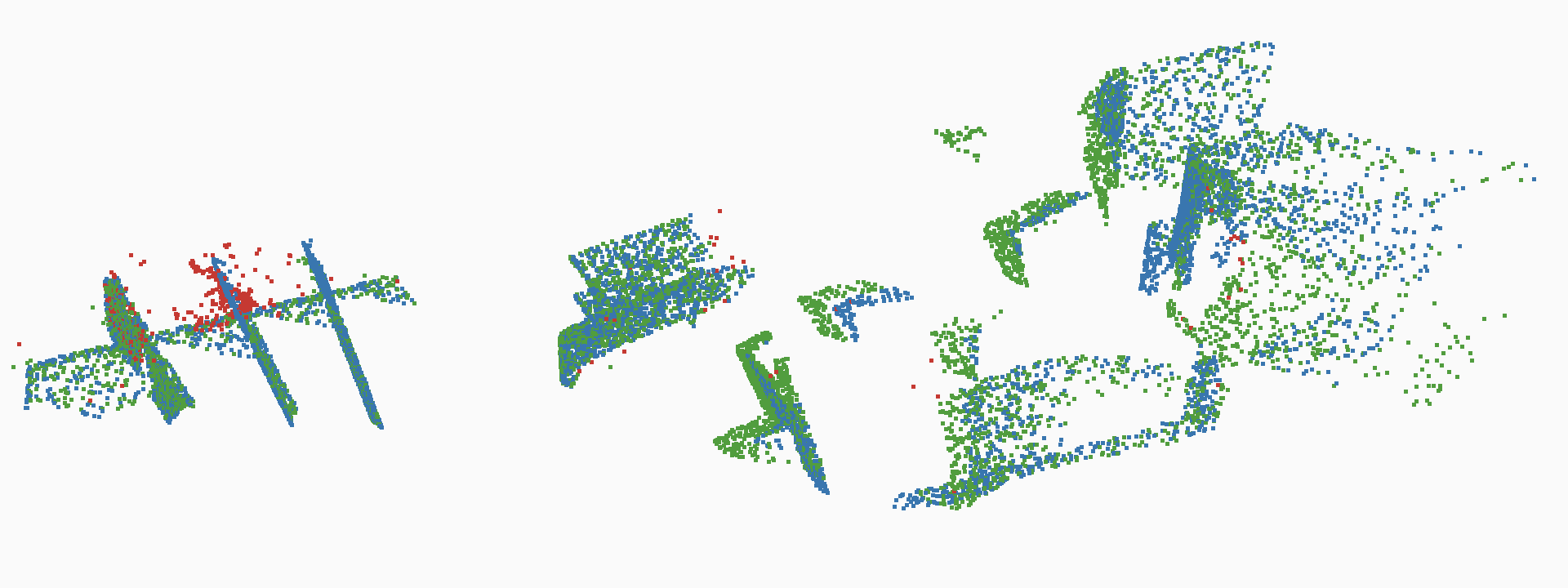}
    \end{subfigure}%
    \hfill
    \begin{subfigure}[b]{0.198\linewidth}
        \includegraphics[width=\linewidth]{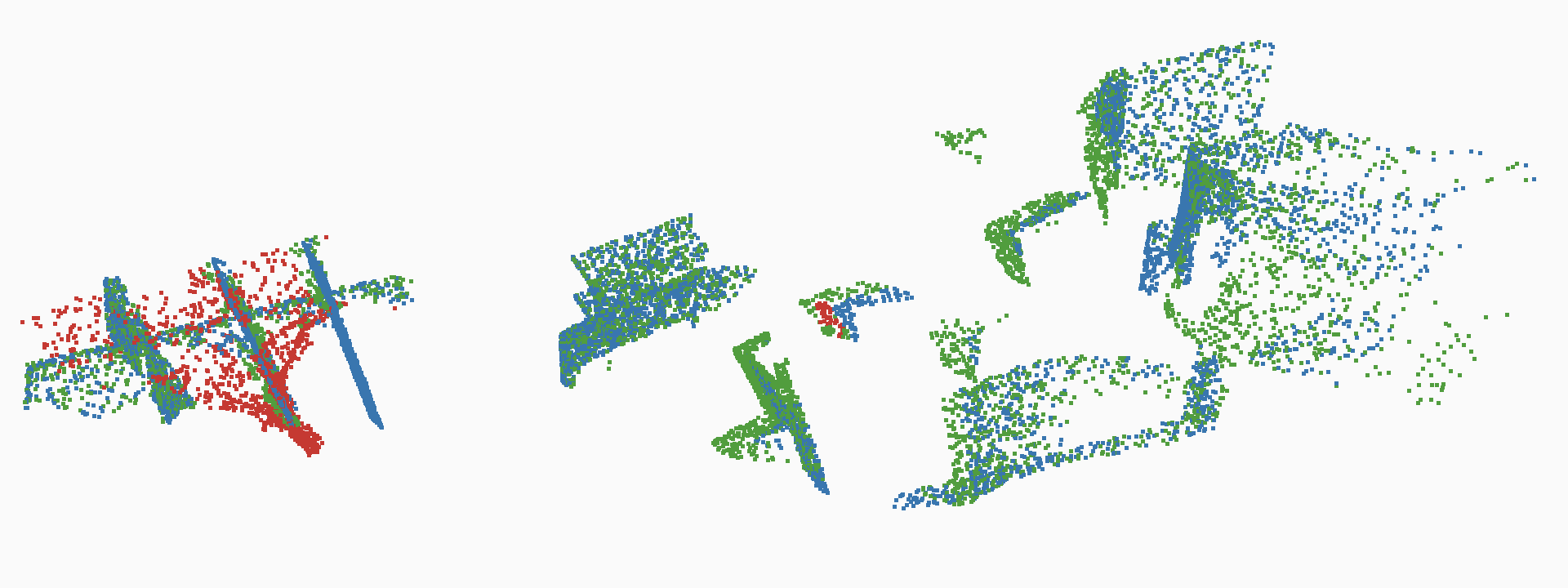}
    \end{subfigure}%
    \hfill
    \begin{subfigure}[b]{0.198\linewidth}
        \includegraphics[width=\linewidth]{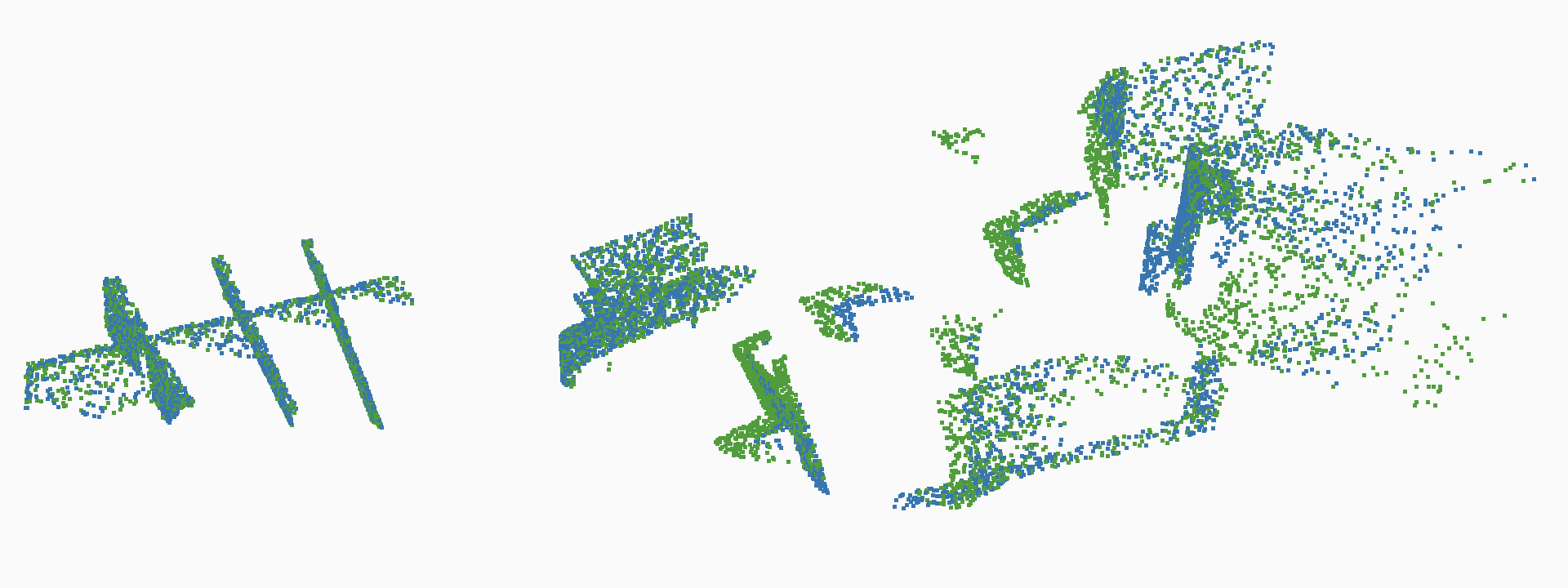}
    \end{subfigure}%
    \hfill
    
    \begin{subfigure}[b]{0.198\linewidth}
        \includegraphics[width=\linewidth]{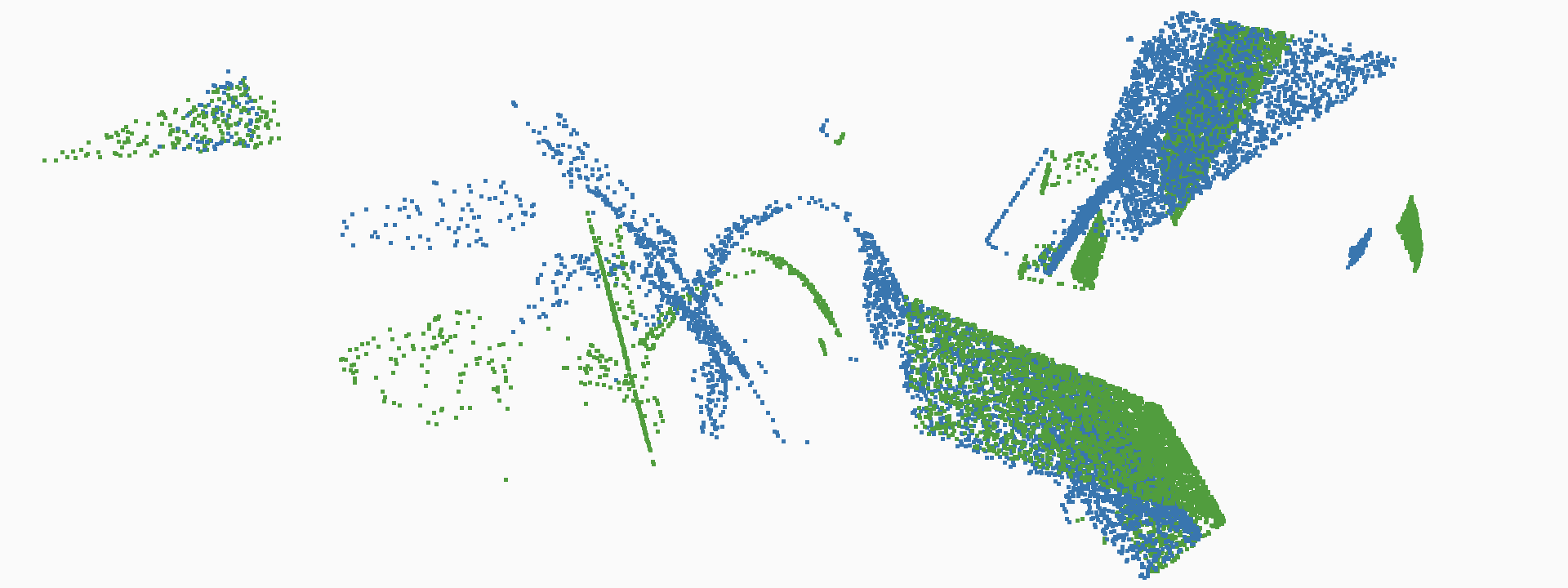}
    \end{subfigure}%
    \hfill
    \begin{subfigure}[b]{0.198\linewidth}
        \includegraphics[width=\linewidth]{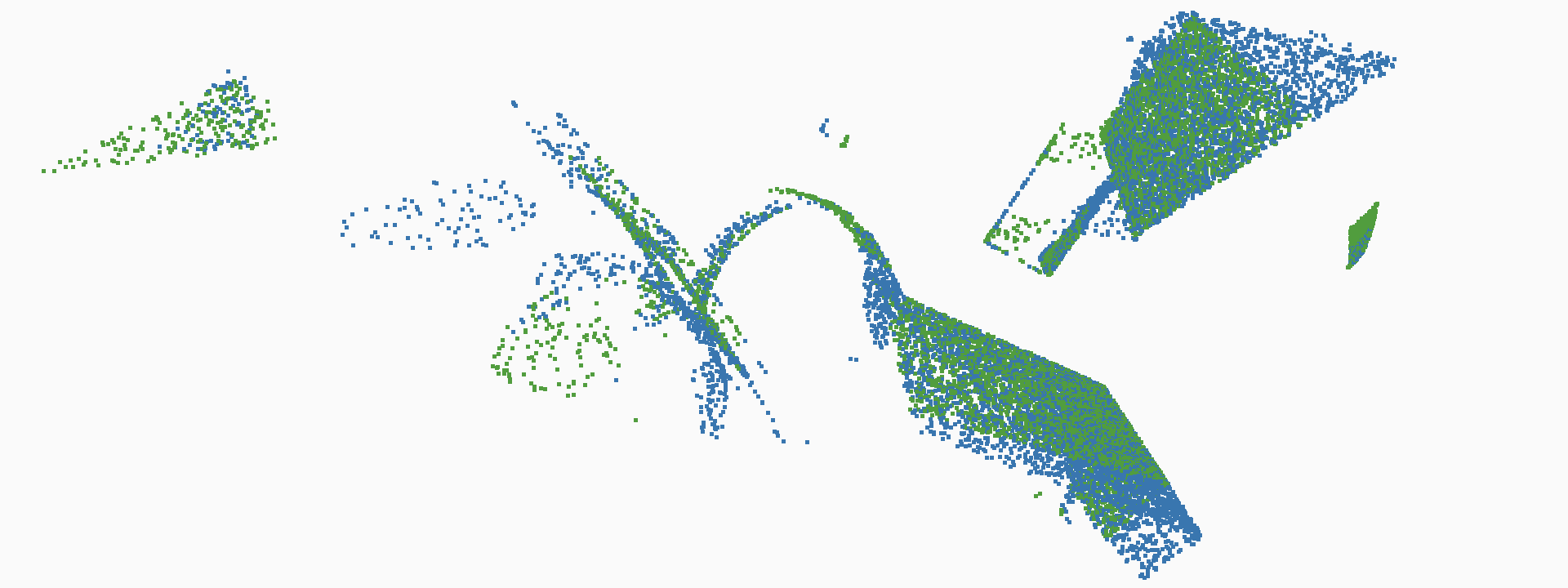}
    \end{subfigure}%
    \hfill
    \begin{subfigure}[b]{0.198\linewidth}
        \includegraphics[width=\linewidth]{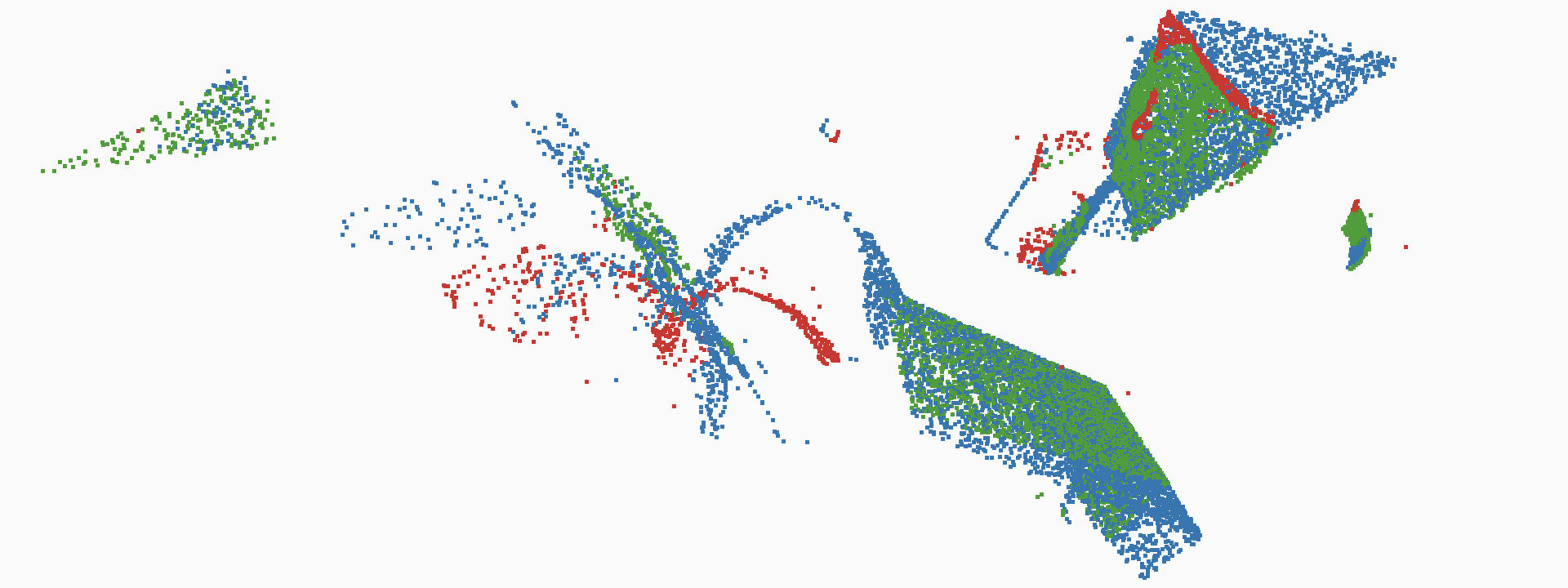}
    \end{subfigure}%
    \hfill
    \begin{subfigure}[b]{0.198\linewidth}
        \includegraphics[width=\linewidth]{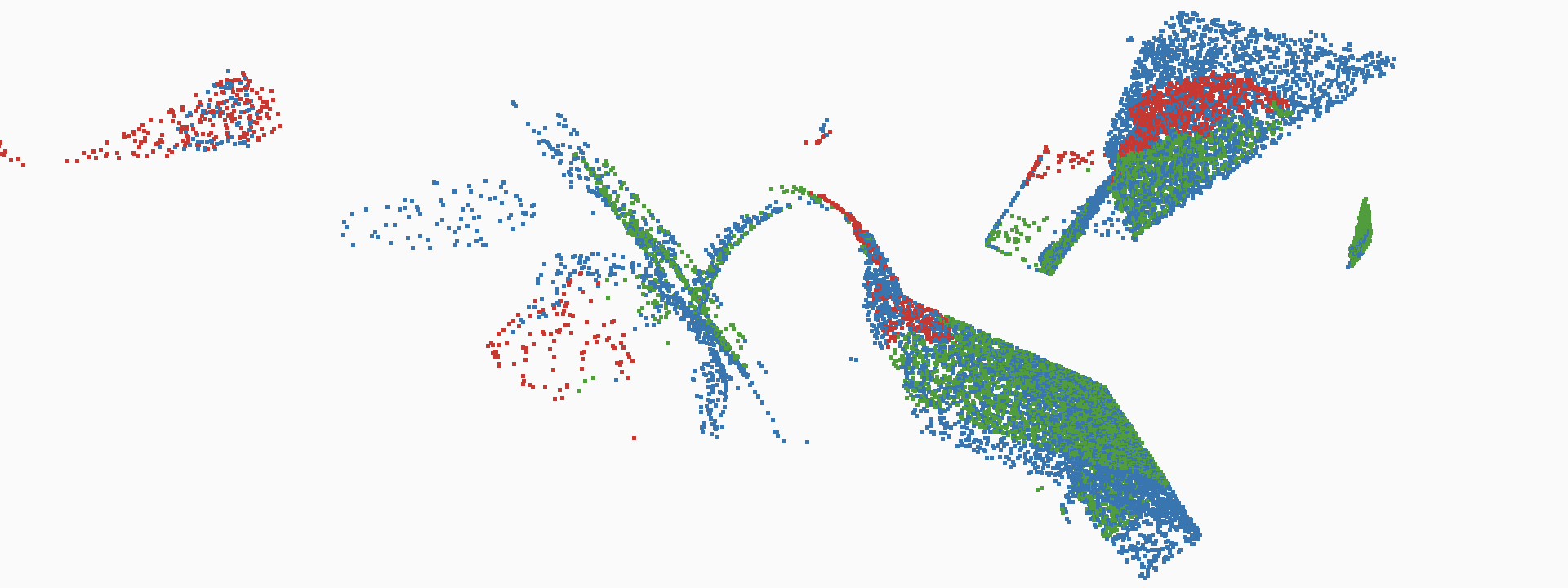}
    \end{subfigure}%
    \hfill
    \begin{subfigure}[b]{0.198\linewidth}
        \includegraphics[width=\linewidth]{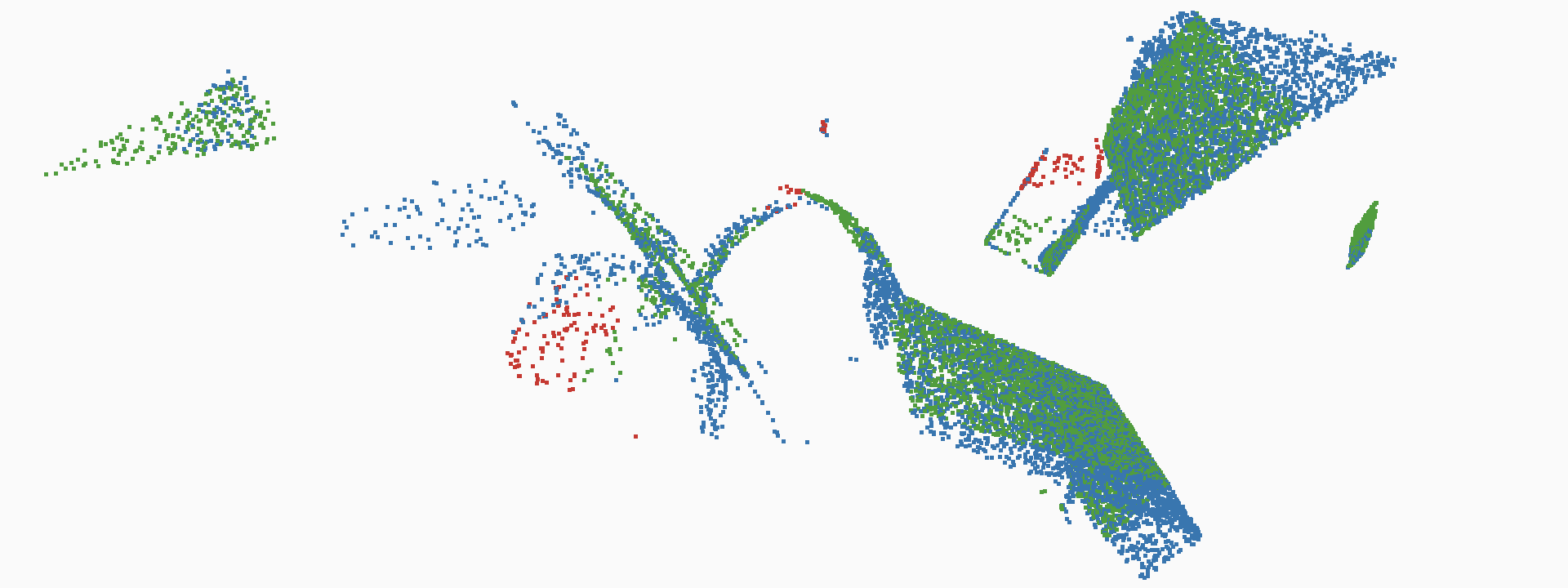}
    \end{subfigure}%
    \hfill

    \vspace{-5pt}
    \caption{Visualized optical flow and scene flow estimation on the ``val'' split of the FlyingThings3D subset. The outliers are marked as \textcolor{red}{red} for scene flow estimation. Our full model better handles objects with repetitive structures and texture-less regions.}
    \vspace{-5pt}
    
    \label{fig:main-things}
\end{figure*}

\begin{table*}
    \centering
    \begin{tabular}{c|c|cc|cc|c}
    \hline
    \multirow{2}{*}{Method} & \multirow{2}{*}{Input} & \multicolumn{2}{c|}{2D Metrics} & \multicolumn{2}{c|}{3D Metrics} & \multirow{2}{*}{Parameters} \\
    & & EPE\textsubscript{2D} & ACC\textsubscript{1px} & EPE\textsubscript{3D} & ACC\textsubscript{.05} \\
    \hline
    FlowNet2.0 \cite{ilg2017flownet2} & Image & 5.05 & 72.8\% & - & - & 162.5M \\
    PWC-Net \cite{sun2018pwc}         & Image & 6.55 & 64.3\% & - & - & 9.4M \\
    RAFT \cite{teed2020raft}          & Image & 3.12 & 81.1\% & - & - & 5.3M \\
    \hline
    FlowNet3D \cite{liu2019flownet3d} & LiDAR & - & - & 0.151 & 20.7\% & 1.2M \\
    PointPWC \cite{wu2019pointpwc}    & LiDAR & - & - & 0.112 & 51.8\% & 5.3M \\
    FLOT \cite{puy2020flot}           & LiDAR & - & - & 0.170 & 23.4\% & 0.1M \\
    \hline
    DeepLiDARFlow \cite{rishav2020deeplidarflow} & Image+LiDAR & 6.04 & 47.1\% & - & 27.2\% & 8.3M \\
    RAFT-3D \cite{teed2021raft3d}                & Image+Depth & 2.37 & 87.1\% & 0.062 & 84.5\% & 45M \\
    \hline
    Ours (W/o fine-tuning) & Image+LiDAR & \textbf{2.18} & 84.3\% & 0.033 & 91.4\% & 7.7M \\
    Ours & Image+LiDAR & 2.20 & \textbf{87.3\%} & \textbf{0.032} & \textbf{92.6\%} & 7.7M \\
    \end{tabular}
    
    \vspace{-5pt}
    \caption{Performance comparison on the ``val'' split of the FlyingThings3D subset. For 2D metrics, we evaluate on full images excluding extremely fast moving regions with flow $>250$px. For 3D metrics, we follow the setup of FlowNet3D in which only non-occluded points with depth $<35$m are considered for evaluation.}
    \vspace{-10pt}
    \label{tab:main-things}
\end{table*}

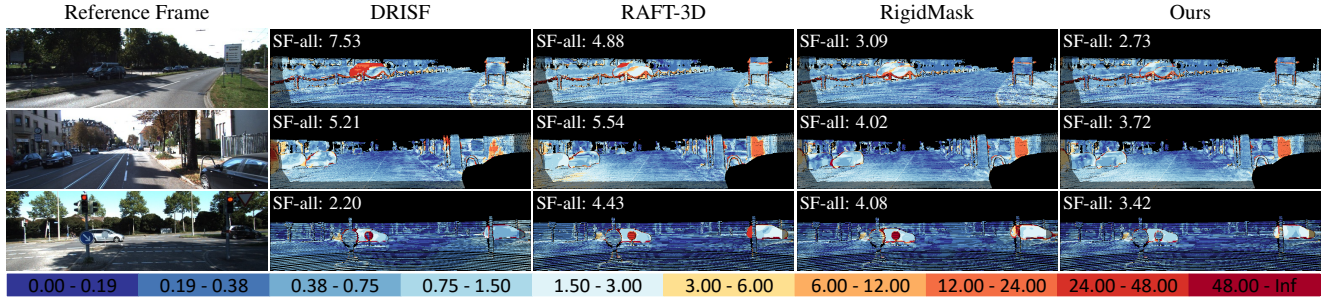
\begin{figure*}
    \captionsetup[subfigure]{labelformat=empty}
    
    \begin{subfigure}[b]{0.198\linewidth}
        \caption{Reference Frame}
    \end{subfigure}%
    \hfill
    \begin{subfigure}[b]{0.198\linewidth}
        \caption{DRISF}
    \end{subfigure}%
    \hfill
    \begin{subfigure}[b]{0.198\linewidth}
        \caption{RAFT-3D}
    \end{subfigure}%
    \hfill
    \begin{subfigure}[b]{0.198\linewidth}
        \caption{RigidMask}
    \end{subfigure}%
    \hfill
    \begin{subfigure}[b]{0.198\linewidth}
        \caption{Ours}
    \end{subfigure}%
    \hfill
    
    \vspace{4pt}
    \begin{subfigure}[b]{0.198\linewidth}
        \begin{picture}(10,10)
        \put(0,0){\includegraphics[width=\linewidth]{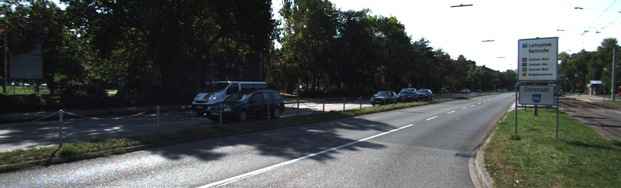}}
        \end{picture}
    \end{subfigure}%
    \hfill
    \begin{subfigure}[b]{0.198\linewidth}
        \begin{picture}(10,10)
        \put(0,0){\includegraphics[width=\linewidth]{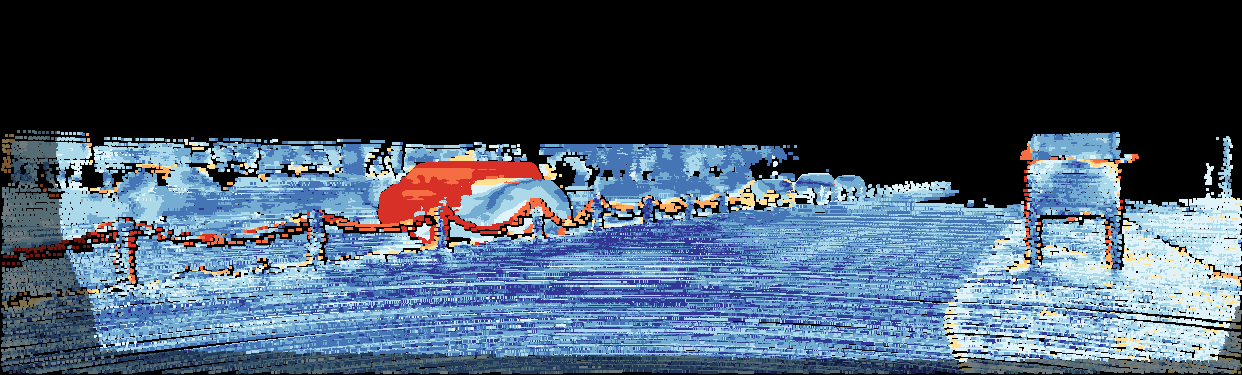}}
        \put(1,23){\scriptsize \textcolor{white}{SF-all: 7.53}}
        \end{picture}
    \end{subfigure}%
    \hfill
    \begin{subfigure}[b]{0.198\linewidth}
        \begin{picture}(10,10)
        \put(0,0){\includegraphics[width=\linewidth]{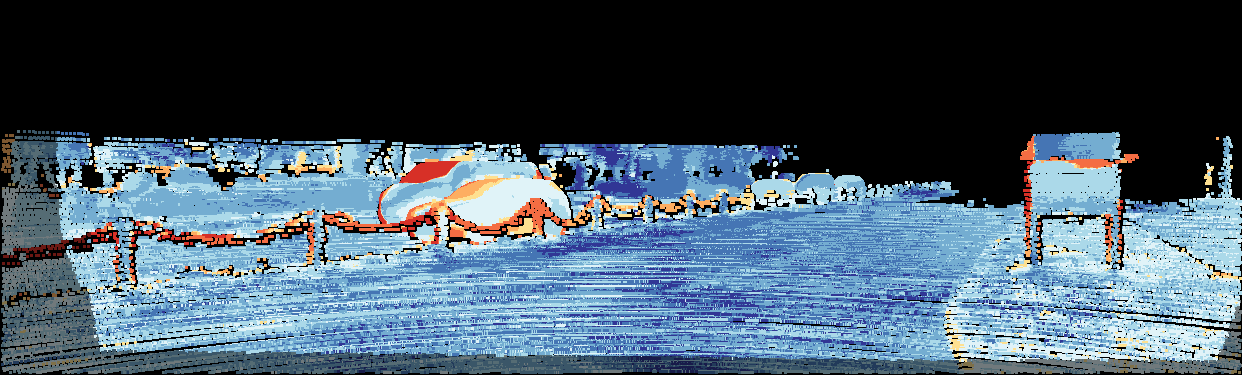}}
        \put(1,23){\scriptsize \textcolor{white}{SF-all: 4.88}}
        \end{picture}
    \end{subfigure}%
    \hfill
    \begin{subfigure}[b]{0.198\linewidth}
        \begin{picture}(10,10)
        \put(0,0){\includegraphics[width=\linewidth]{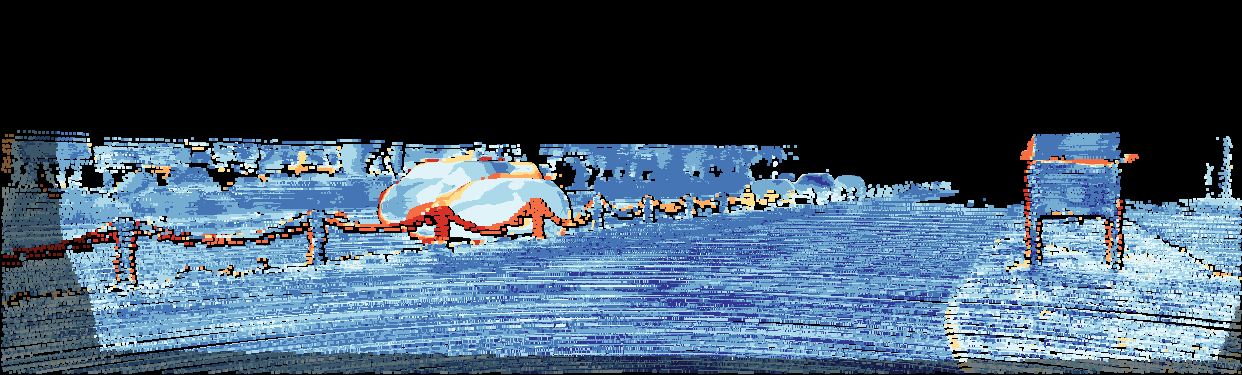}}
        \put(1,23){\scriptsize \textcolor{white}{SF-all: 3.09}}
        \end{picture}
    \end{subfigure}%
    \hfill
    \begin{subfigure}[b]{0.198\linewidth}
        \caption{Ours}
        \begin{picture}(10,10)
        \put(0,0){\includegraphics[width=\linewidth]{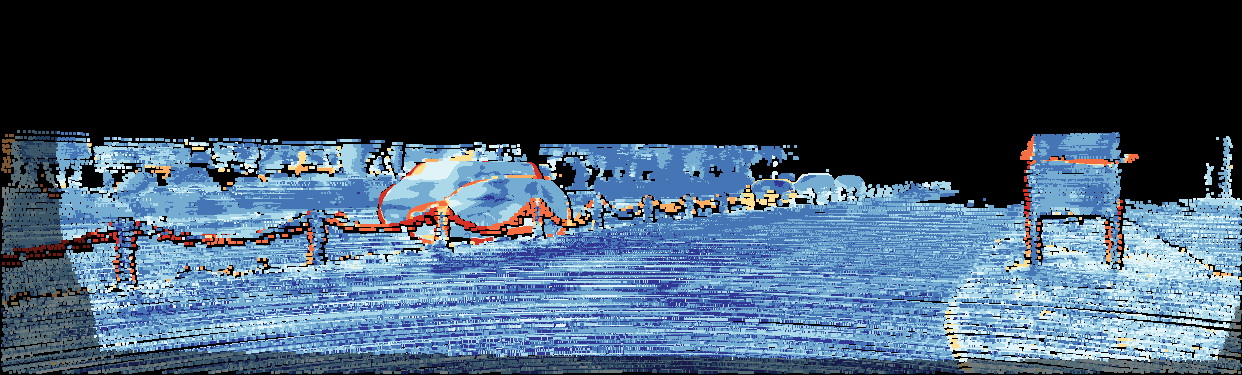}}
        \put(1,23){\scriptsize \textcolor{white}{SF-all: 2.73}}
        \end{picture}
    \end{subfigure}%
    \hfill
    
    \begin{subfigure}[b]{0.198\linewidth}
        \includegraphics[width=\linewidth]{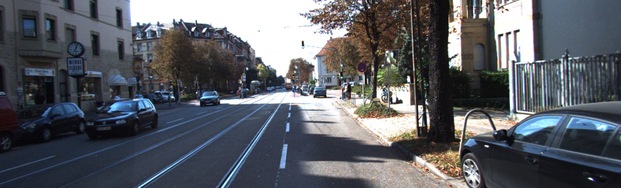}
    \end{subfigure}%
    \hfill
    \begin{subfigure}[b]{0.198\linewidth}
        \begin{picture}(10,10)
        \put(0,0){\includegraphics[width=\linewidth]{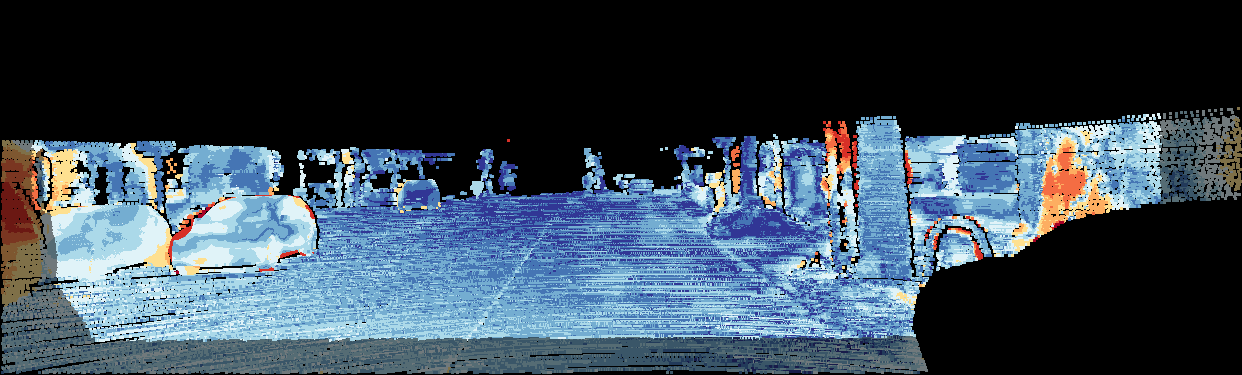}}
        \put(1,23){\scriptsize \textcolor{white}{SF-all: 5.21}}
        \end{picture}
    \end{subfigure}%
    \hfill
    \begin{subfigure}[b]{0.198\linewidth}
        \begin{picture}(10,10)
        \put(0,0){\includegraphics[width=\linewidth]{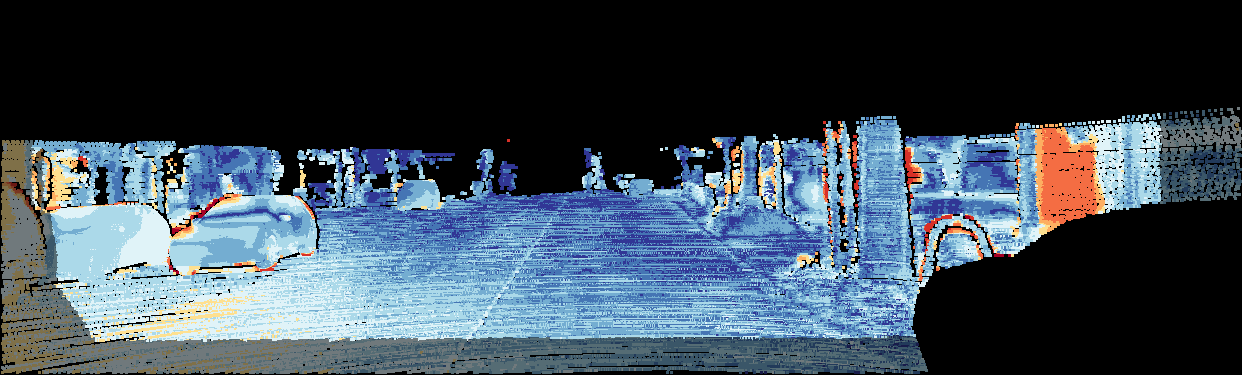}}
        \put(1,23){\scriptsize \textcolor{white}{SF-all: 5.54}}
        \end{picture}
    \end{subfigure}%
    \hfill
    \begin{subfigure}[b]{0.198\linewidth}
        \begin{picture}(10,10)
        \put(0,0){\includegraphics[width=\linewidth]{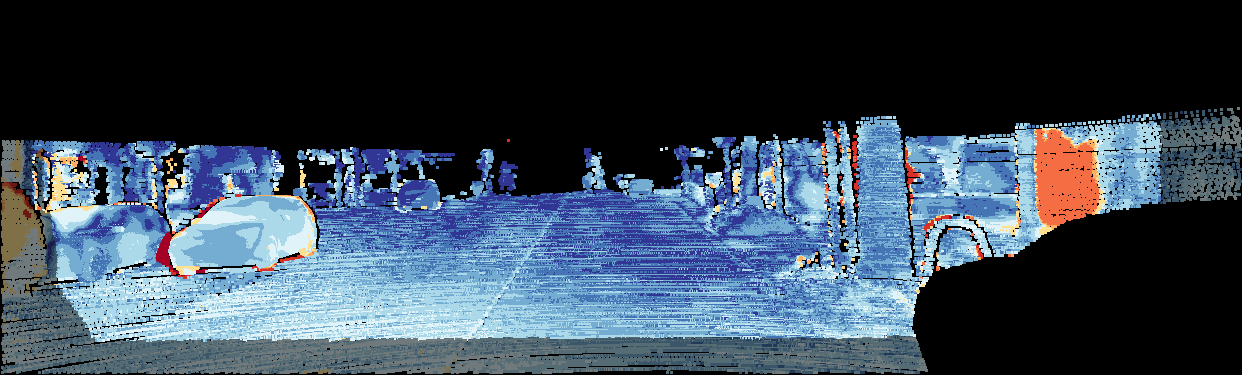}}
        \put(1,23){\scriptsize \textcolor{white}{SF-all: 4.02}}
        \end{picture}
    \end{subfigure}%
    \hfill
    \begin{subfigure}[b]{0.198\linewidth}
        \begin{picture}(10,10)
        \put(0,0){\includegraphics[width=\linewidth]{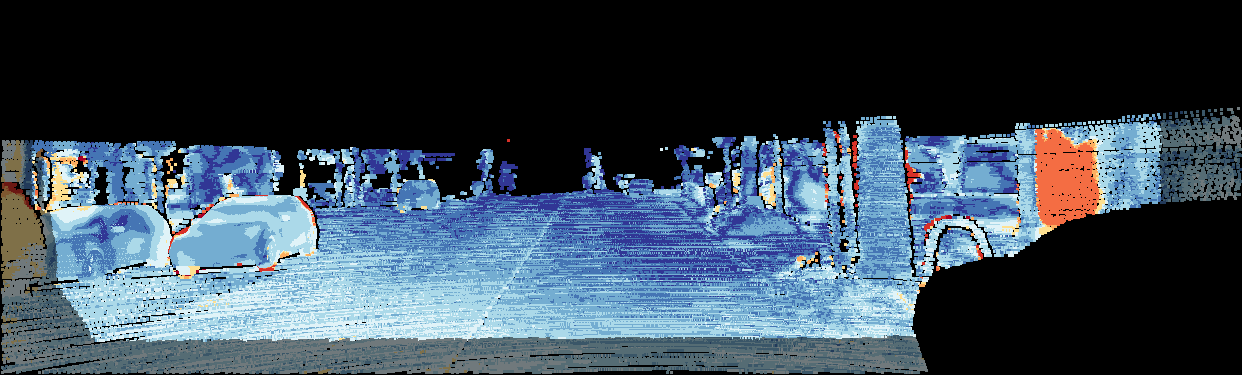}}
        \put(1,23){\scriptsize \textcolor{white}{SF-all: 3.72}}
        \end{picture}
    \end{subfigure}%
    \hfill
    
    \begin{subfigure}[b]{0.198\linewidth}
        \includegraphics[width=\linewidth]{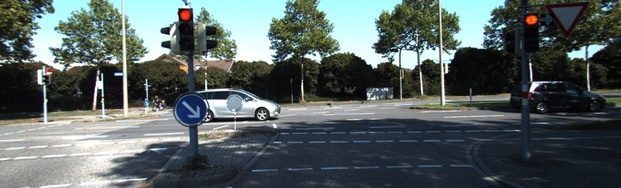}
    \end{subfigure}%
    \hfill
    \begin{subfigure}[b]{0.198\linewidth}
        \begin{picture}(10,10)
        \put(0,0){\includegraphics[width=\linewidth]{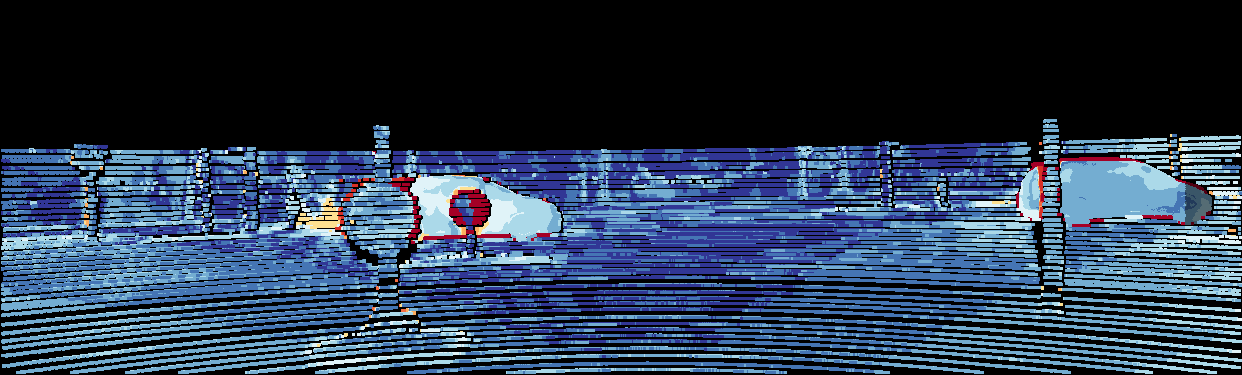}}
        \put(1,23){\scriptsize \textcolor{white}{SF-all: 2.20}}
        \end{picture}
    \end{subfigure}%
    \hfill
    \begin{subfigure}[b]{0.198\linewidth}
        \begin{picture}(10,10)
        \put(0,0){\includegraphics[width=\linewidth]{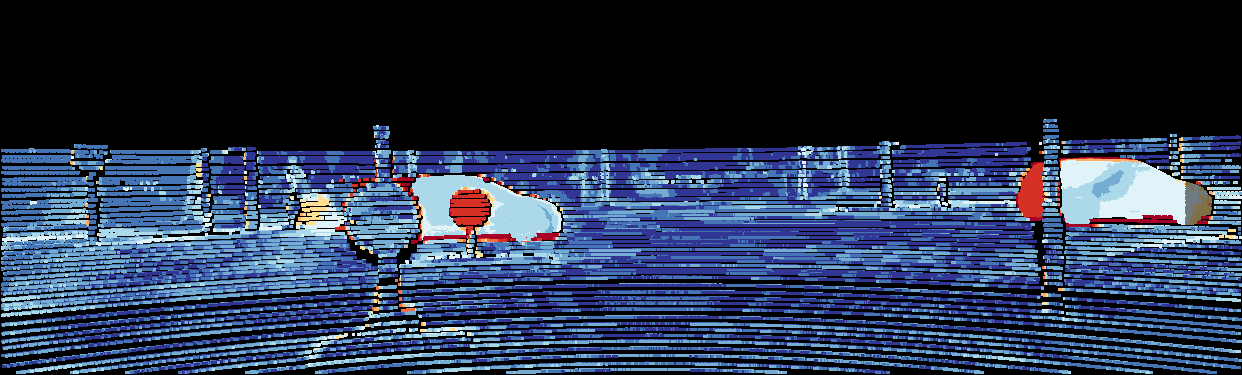}}
        \put(1,23){\scriptsize \textcolor{white}{SF-all: 4.43}}
        \end{picture}
    \end{subfigure}%
    \hfill
    \begin{subfigure}[b]{0.198\linewidth}
        \begin{picture}(10,10)
        \put(0,0){\includegraphics[width=\linewidth]{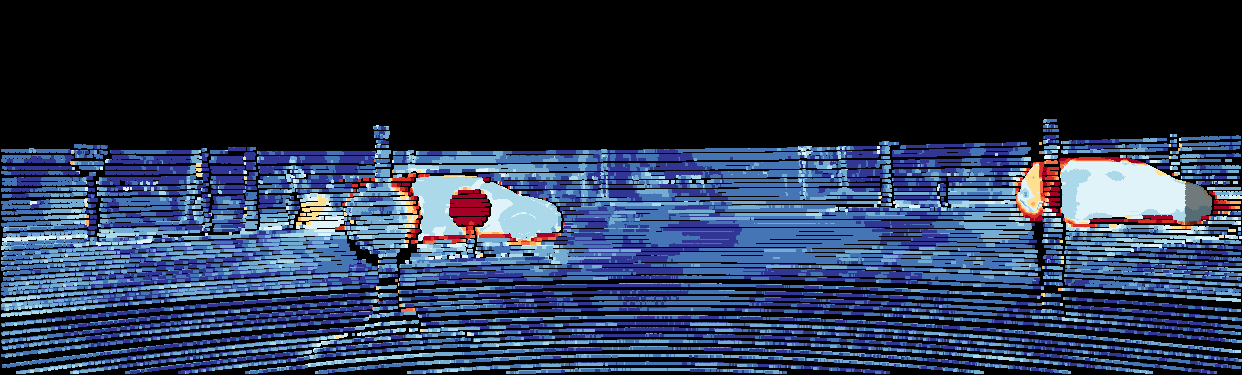}}
        \put(1,23){\scriptsize \textcolor{white}{SF-all: 4.08}}
        \end{picture}
    \end{subfigure}%
    \hfill
    \begin{subfigure}[b]{0.198\linewidth}
        \begin{picture}(10,10)
        \put(0,0){\includegraphics[width=\linewidth]{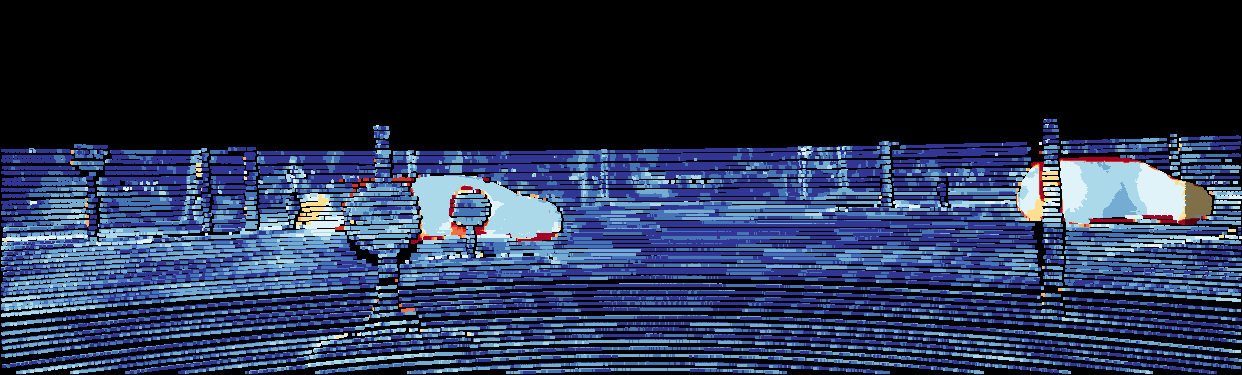}}
        \put(1,23){\scriptsize \textcolor{white}{SF-all: 3.42}}
        \end{picture}
    \end{subfigure}%
    \hfill
    
    \vspace{-2pt}
    \includegraphics[width=\linewidth]{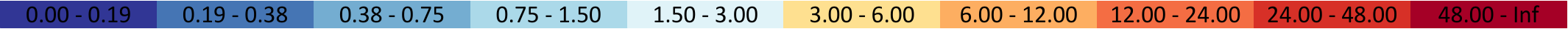}
    \vspace{-18pt}
    
    \caption{Qualitative results on the KITTI Scene Flow test set. Blue indicates a lower error, red indicates a higher error. Our approach improves accuracy near motion boundaries.}
    \vspace{-5pt}
    
    \label{fig:main-kitti}
\end{figure*}

\begin{table*}
    \centering
    \begin{tabular}{c|c|cc|cc|cc|cc|c}
    \hline
    \multirow{2}{*}{Method} & \multirow{2}{*}{Rigidity} & \multicolumn{2}{c|}{D1 (\%)} & \multicolumn{2}{c|}{D2 (\%)} & \multicolumn{2}{c|}{OF (\%)} & \multicolumn{2}{c|}{SF (\%)} & \multirow{2}{*}{Parameters} \\
    & & noc & all & noc & all & noc & all & noc & all & \\
    \hline
    SENSE \cite{jiang2019sense} & Background & 2.05 & 2.22 & 3.87 & 5.89 & 5.98 & 7.64 & 7.30 & 9.55 & 13.4M \\
    Binary TTC \cite{badki2021binaryttc} & None & \textbf{1.63} & \textbf{1.81} & 2.72 & 4.76 & 3.89 & 6.31 & 5.29 & 8.50 & - \\
    OpticalExp \cite{yang2020opticalexp} & None & \textbf{1.63} & \textbf{1.81} & 2.62 & 4.25 & 3.89 & 6.30 & 5.21 & 8.12 & 18.5M \\
    ISF \cite{behl2017isf} & Full Image & 4.02 & 4.46 & 4.95 & 5.95 & 4.69 & 6.22 & 6.45 & 8.08 & - \\
    ACOSF \cite{li2021acosf} & Full Image & 3.35 & 3.58 & 4.26 & 5.31 & 4.51 & 5.79 & 6.40 & 7.90 & - \\
    DRISF \cite{ma2019drisf} & Full Image & 2.35 & 2.55 & 3.14 & 4.04 & 3.58 & 4.73 & 4.99 & 6.31 & 58.9M \\
    RAFT-3D \cite{teed2021raft3d} & Full Image & \textbf{1.63} & \textbf{1.81} & 2.67 & 3.67 & 3.23 & 4.29 & 4.53 & 5.77 & 51.3M \\
    RigidMask \cite{yang2021rigidmask} & Full Image & 1.70 & 1.89 & 2.47 & 3.23 & 2.54 & 3.50 & 3.73 & 4.89 & 145.3M \\
    \hline
    Ours (non-rigid) & None & \textbf{1.63} & \textbf{1.81} & 2.39 & 3.19 & 2.77 & 4.05 & 4.03 & 5.62 & 14.0M \\
    Ours & Background & \textbf{1.63} & \textbf{1.81} & \textbf{2.37} & \textbf{2.95} & \textbf{2.40} & \textbf{3.10} & \textbf{3.55} & \textbf{4.43} & 19.7M \\
    \end{tabular}
    \vspace{-5pt}
    \caption{Leaderboard of the KITTI Scene Flow benchmark. ``D1'', ``D2'', and ``OF'' are predicted by GA-Net, our point branch and our image branch respectively. Our method ranks first on the leaderboard and can handle general non-rigid motions since we only treat the static background as rigid.}
    \vspace{-10pt}
    \label{tab:main-kitti}
\end{table*}

We implement our model using PyTorch \cite{paszke2019pytorch}. For all experiments we use the Adam optimizer \cite{kingma2014adam} with weight decay set to $10^{-6}$. The loss weights are set to $\alpha_0=8$, $\alpha_1=4$, $\alpha_2=2$, $\alpha_3=1$, and $\alpha_4=0.5$.

\subsection{Main Results}

We evaluate our method on the synthetic dataset FlyingThings3D \cite{mayer2016things3d} and the real-world dataset KITTI \cite{menze2015osf}. FlyingThings3D consists of stereo and RGB-D images rendered with multiple randomly moving objects from ShapeNet \cite{chang2015shapenet}, which is large-scale and challenging. KITTI Scene Flow is a real-world benchmark for autonomous driving, consisting of 200 training scenes and 200 test scenes.

\subsubsection{FlyingThings3D}

\paragraph{Data Preprocessing.} Following previous work \cite{ilg2017flownet2, gu2019hplflownet, wu2019pointpwc}, we use the subset of FlyingThings3D. The training and validation set respectively contains 19640 and 3824 pairs of camera-LiDAR frames. We follow FlowNet3D \cite{liu2019flownet3d} instead of HPLFlowNet \cite{gu2019hplflownet} to lift the depth images to point clouds, since HPLFlowNet only keeps non-occluded points which oversimplifies the problem. 

\vspace{-10pt}
\paragraph{Training.} The training consists of two stages. First, we train our model for 600 epochs with the $L_2$-norm loss function. The initial learning rate is set to $4 \times 10^{-4}$ and reduced by half at 400 and 500 epochs. Next, we fine-tune our model for another 800 epochs with the robust loss function and a fixed learning rate of $10^{-4}$. The batch size is set to 32.

\vspace{-10pt}
\paragraph{Evaluation Metrics.} Following RAFT-3D, we evaluate our network using 2D and 3D end-point error (EPE), as well as threshold metrics (ACC\textsubscript{1px} and ACC\textsubscript{.05}), which measure the portion of error within a threshold.

\vspace{-10pt}
\paragraph{Quantitative Results.} In Tab. \ref{tab:main-things}, we compare to several state-of-the-art methods which utilize different input modalities. By fusing the two modalities of camera and LiDAR, our method outperforms all image-only and LiDAR-only methods by a large margin. Our method also outperforms RAFT-3D, which has 45M parameters and takes dense RGB-D frames as input. In contrast, our model is much more lightweight with 7.7M parameters and only requires sparse depth measurements. Moreover, our model reduces the best published EPE\textsubscript{3D} from 0.062 to 0.032, which proves the superior performance of the point branch.

\vspace{-10pt}
\paragraph{Qualitative Results.} The visual comparison of optical flow and scene flow estimation is shown in Fig. \ref{fig:main-things}. We also add two single-modal variations of our method for comparison, which removes the 2D branch or the 3D branch. As we can see, our full model better handles objects with repetitive structures and complex scenes with overlapping objects.

\subsubsection{KITTI}

\paragraph{Training.} Using the weight pre-trained on FlyingThings3D and Driving \cite{mayer2016things3d}, we fine-tune our model on KITTI for 300 epochs with a fixed learning rate of $5 \times 10^{-5}$ and a batch size of 8. We follow \cite{yang2021rigidmask, yang2020opticalexp} and divide the 200 training images into \textit{train}, \textit{val} splits based on the 4:1 ratio. During training, we lift the ground-truth disparity maps into point clouds using the provided calibration parameters. Basic data augmentation strategies including color jitter, random horizontal flipping, and random cropping are applied.

\vspace{-10pt}
\paragraph{Testing.} During testing, since neither disparity maps nor point clouds are provided, we employ GA-Net \cite{zhang2019ganet} to estimate the disparity from stereo images, and generate point clouds with depth $<90$m. The sparse output of our point branch is interpolated to create a dense prediction.

\vspace{-10pt}
\paragraph{Refinement of Background Scene Flow.} Since most background objects in KITTI are rigid (e.g. ground, buildings, etc), we can refine the background scene flow using a rigidity refinement step. Specifically, we employ DDRNet-Slim \cite{hong2021ddrnet}, a light-weight 2D semantic segmentation network, to determine the rigid background. DDRNet-Slim is pre-trained on Cityscapes \cite{cordts2016cityscapes} and fine-tuned on KITTI. Next, we estimate ego-motion by fitting and decomposing essential matrices from the background flow map using a neural-guided RANSAC \cite{brachmann2019ransac}. Finally, the background scene flow is refined using the ego-motion and the disparity of the first frame.

\vspace{-10pt}
\paragraph{Comparison with State-of-the-art Methods.} We submit our approach to the website of KITTI Scene Flow benchmark and report the leaderboard in Tab. \ref{tab:main-kitti}. A visualized comparison is shown in Fig. \ref{fig:main-kitti}. Our approach outperforms all published methods, including RigidMask \cite{yang2021rigidmask} (SF-all: 4.43\% vs. 4.89\%), which employs more than 140M parameters. In contrast, our method is much more lightweight with only 19.7M parameters (6.3M GA-Net + 7.7M CamLiFlow + 5.7M DDRNet-Slim). Moreover, previous methods leverage more strict rigid-body assumptions by assigning rigid motions to all objects, while our method can handle general non-rigid motions since we only apply rigid motion refinement to the static background.

If the rigidity refinement step of the background scene flow is removed (corresponding to our ``non-rigid'' variation in Tab. \ref{tab:main-kitti}), our method still ranks second on the leaderboard (SF-all: 5.62\%). In this setting, our method does not require the background segmentation labels and can deal with any non-rigid motion (no matter foreground or background). Instead, RigidMask fails to handle non-rigid motions and suffers from the limitations of the motion segmentation network, since the whole pipeline depends on its results.

\subsection{Ablation Study}
\label{sec:ablation}

In this section, we conduct ablation studies on FlyingThings3D to confirm the effectiveness of each module. All variations are trained for the first stage without fine-tuning with the robust loss function.

\begin{table}[t]
    \centering
    \begin{tabular}{c|cc|cc}
    \hline
    Fusion & \multicolumn{2}{c|}{2D Metrics} & \multicolumn{2}{c}{3D Metrics} \\
    Direction & EPE\textsubscript{2D} & ACC\textsubscript{1px} & EPE\textsubscript{3D} & ACC\textsubscript{.05} \\
    \hline
    2D $\Rightarrow$ 3D      & 3.41 & 79.5\% & 0.036 & 89.6\% \\
    2D $\Leftarrow$ 3D       & 2.25 & 83.9\% & 0.067 & 74.0\% \\
    2D $\Leftrightarrow$ 3D  & \underline{\textbf{2.18}} & \underline{\textbf{84.3}\%} & \underline{\textbf{0.033}} & \underline{\textbf{91.4}\%} \\
    \hline
    \end{tabular}
    \vspace{-5pt}
    \caption{Unidirectional Fusion vs. Bidirectional Fusion. Bidirectional fusion provides the best results for all metrics since both modalities can benefit each other.}
    \vspace{-15pt}
    \label{tab:ablation-fusion-direction}
\end{table}

\vspace{-10pt}
\paragraph{Unidirectional Fusion vs. Bidirectional Fusion.} CamLiFlow fuses features in a bidirectional manner. Here, we train two variations where features are fused in a unidirectional manner (2D $\Rightarrow$ 3D or 2D $\Leftarrow$ 3D). As shown in Tab. \ref{tab:ablation-fusion-direction}, unidirectional fusion either improves 2D metrics or 3D metrics, while bidirectional fusion provides better results for both modalities. Moreover, compared with unidirectional fusion, bidirectional fusion improves the best EPE\textsubscript{2D} from 2.25 to 2.18 and EPE\textsubscript{3D} from 0.036 to 0.033, suggesting that the improvement of one modality can also benefit the other.

\begin{table}[t]
    \centering
    \begin{tabular}{ccc|cc|cc}
    \hline
    \multicolumn{3}{c|}{Stages} & \multicolumn{2}{c|}{2D Metrics} & \multicolumn{2}{c}{3D Metrics} \\
    $P$ & $C$ & $D$ & EPE\textsubscript{2D} & ACC\textsubscript{1px} & EPE\textsubscript{3D} & ACC\textsubscript{.05} \\
    \hline
    - & - & -                   & 3.42 & 79.5\% & 0.067 & 74.1\% \\
    $\surd$ & - & -             & 2.69 & 81.5\% & 0.047 & 87.6\% \\
    - & $\surd$ & -             & 2.42 & 82.3\% & 0.037 & 89.6\% \\
    - & - & $\surd$             & 2.40 & 82.7\% & 0.038 & 88.1\% \\
    $\surd$ & $\surd$ & -       & 2.29 & 83.2\% & 0.034 & 90.7\% \\
    $\surd$ & $\surd$ & $\surd$ & \underline{\textbf{2.18}} & \underline{\textbf{84.3}\%} & \underline{\textbf{0.033}} & \underline{\textbf{91.4}\%} \\
    \hline
    \end{tabular}
    \vspace{-5pt}
    \caption{Early/Late-Fusion vs. Multi-Stage Fusion. $P$, $C$, $D$ denotes features of pyramid, cost volume, and flow decoder respectively. Fusing at all three stages gives the best result.}
    \label{tab:ablation-multi-stage}
    \vspace{-5pt}
\end{table}

\begin{table}[t]
    \centering
    \begin{tabular}{ccc|cc}
    \hline
    \multicolumn{3}{c|}{Configurations} & \multicolumn{2}{c}{2D Metrics} \\
    $k$-NN & 2D Similarity & SOP & EPE\textsubscript{2D} & ACC\textsubscript{1px} \\
    \hline
    - & - & -                       & 2.30 & 83.3\% \\
    $k=1$ & - & -                   & 2.24 & 84.3\% \\
    $k=1$ & $\surd$ & -             & \underline{\textbf{2.18}} & \underline{84.3\%} \\
    $k=3$ & $\surd$ & \texttt{MEAN} & 2.19 & \textbf{84.5}\% \\
    $k=3$ & $\surd$ & \texttt{MAX}  & 2.19 & 84.4\% \\
    \hline
    \end{tabular}
    \vspace{-5pt}
    \caption{Ablations on the interpolation module of Bi-CLFM. We only report 2D metrics since 3D metrics are all similar.}
    \vspace{-5pt}
    \label{tab:ablation-clfm-interpolation}
\end{table}

\begin{table}[t]
    \centering
    \begin{tabular}{c|c|cc|cc}
    \hline
    \multirow{2}{*}{Setup} & \multirow{2}{*}{IDS} & \multicolumn{2}{c|}{2D Metrics} & \multicolumn{2}{c}{3D Metrics} \\
    & & EPE\textsubscript{2D} & ACC\textsubscript{1px} & EPE\textsubscript{3D} & ACC\textsubscript{.05} \\
    \hline
    C+L & -       & 2.24 & 83.2\% & 0.036 & 88.7\% \\
    C+L & $\surd$ & \textbf{2.18} & \textbf{84.3\%} & \textbf{0.033} & \textbf{91.4\%} \\
    \hline
    L & -         & - & - & 0.073 & 70.6\% \\
    L & $\surd$   & - & - & \textbf{0.068} & \textbf{74.3\%} \\
    \hline
    \end{tabular}
    \vspace{-5pt}
    \caption{Ablations on inverse depth scaling (IDS). ``L'' denotes a LiDAR-only variation of our method where the image branch is removed, while ``C+L'' denotes our full model. IDS improves performance for both LiDAR-only and Camera-LiDAR methods.}
    \vspace{-5pt}
    \label{tab:ablation-ids}
\end{table}

\vspace{-10pt}
\paragraph{Early/Late-Fusion vs. Multi-Stage Fusion.} As mentioned above, flow estimation typically consists of several stages including feature extraction, cost volume, and feature decoding. Here, we verify the effectiveness of feature fusion for each stage, as shown in Tab. \ref{tab:ablation-multi-stage}. The top row denotes the version where no fusion connection exists between the two branches. Both ``early-fusion'' and ``late-fusion'' (row 2, 3, 4) can only provide sub-optimal results. In contrast, fusing features at all three stages brings significant improvements compared to ``early/late-fusion'' (see the supplementary material for more details).

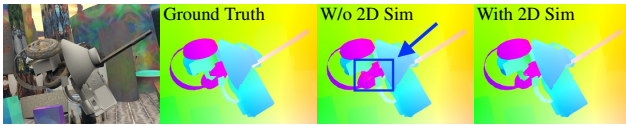
\begin{figure}[t]
    \captionsetup[subfigure]{labelformat=empty}

    \begin{subfigure}[b]{0.25\linewidth}
        \includegraphics[width=\linewidth]{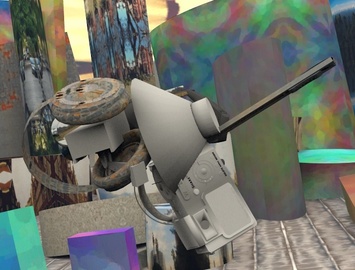}
    \end{subfigure}%
    \hfill
    \begin{subfigure}[b]{0.25\linewidth}
        \begin{picture}(10,10)
        \put(0,0){\includegraphics[width=\linewidth]{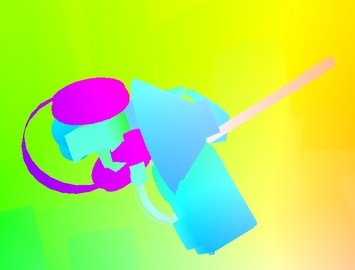}}
        \put(1,39){\scriptsize Ground Truth}
        \end{picture}
    \end{subfigure}%
    \hfill
    \begin{subfigure}[b]{0.25\linewidth}
        \begin{picture}(10,10)
        \put(0,0){\includegraphics[width=\linewidth]{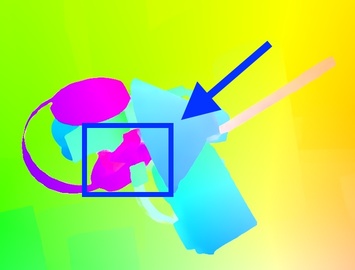}}
        \put(1,39){\scriptsize W/o 2D Sim}
        \end{picture}
    \end{subfigure}%
    \hfill
    \begin{subfigure}[b]{0.25\linewidth}
        \begin{picture}(10,10)
        \put(0,0){\includegraphics[width=\linewidth]{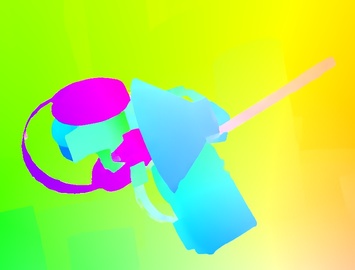}}
        \put(1,39){\scriptsize With 2D Sim}
        \end{picture}
    \end{subfigure}%
    
    \vspace{-5pt}
    \caption{Effects of introducing 2D similarities in fusion-aware interpolation, which better handles overlapping objects.}
    \vspace{-10pt}
    \label{fig:ablation-sim2d}
\end{figure}

\vspace{-10pt}
\paragraph{Fusion-Aware Interpolation.} In Tab. \ref{tab:ablation-clfm-interpolation}, we test the interpolation module of Bi-CLFM with different configurations. The top row denotes a naive implementation that simply projects 3D features onto the image plane without interpolation (empty locations are filled with zeros). By introducing learnable weights into a nearest-neighbor interpolation (the 2nd row), we reduce EPE\textsubscript{2D} from 2.30 to 2.24, and improve ACC\textsubscript{1px} from 83.3\% to 84.3\%. Integrating 2D similarity measurements into the interpolation module (the 3rd row) further reduces EPE\textsubscript{2D} from 2.24 to 2.18, and makes our model more robust in complex scenes with overlapping objects (as shown in Fig. \ref{fig:ablation-sim2d}). We also conduct two experiments by increasing $k$ (the number of nearest neighbors) from 1 to 3, followed by a symmetric operation such as \texttt{MEAN} and \texttt{MAX}. However, no significant improvement is observed, suggesting that $k=1$ is enough for interpolation.

\vspace{-10pt}
\paragraph{Inverse Depth Scaling.} We conduct several comparison experiments on FlyingThings3D to verify the effects of IDS. Since IDS does not require input images, we also test it on a variation of our method where the image branch is removed. As shown in Tab. \ref{tab:ablation-ids}, the performance is improved under both the camera-LiDAR and the LiDAR-only setup, suggesting that a more even distribution of points can facilitate the learning.

\begin{table}[t]
    \centering
    \begin{tabular}{l|c|c}
        \hline
        \multirow{2}{*}{Components} & Individual & Total \\
        & Timing & Timing \\
        \hline
        Image branch & 24ms & \multirow{6}{*}{118ms} \\
        Point branch & 45ms & \\
        Bi-CLFM (for feature pyramid) & 17ms & \\
        Bi-CLFM (for cost volume) & 7ms & \\
        Bi-CLFM (for flow decoder) & 11ms & \\
        Others & 14ms & \\
        \hline
    \end{tabular}
    \vspace{-5pt}
    \caption{Breakdown of the inference time. Timing results are computed on 960x540 images with a Tesla V100 GPU.}
    \vspace{-10pt}
    \label{tab:runtime}
\end{table}

\subsection{Analysis}

\paragraph{Timing.} In Tab. \ref{tab:runtime}, we provide a breakdown of the inference time on FlyingThings3D (960x540 images). Our model takes 118ms in total with a Tesla V100 GPU.

\vspace{-10pt}
\paragraph{Limitations.} CamLiFlow has two limitations. First, a synchronized camera and LiDAR are required for optimal performance. If the synchronization cannot be met by some applications, our method can also take advantage of other depth sensors, such as stereo cameras (but they are less accurate and less robust than LiDAR).
Another limitation is that since the two modalities are tightly coupled, the whole system will fail if one of them does not work. In the future, we plan to address this by introducing the attention mechanism to the fusion module so that the model can ``choose'' to ignore the modality if it does not work.

\section{Conclusion}

In this paper, we introduce CamLiFlow, a deep neural network for joint optical flow and scene flow estimation. It consists of 2D and 3D branches with multiple bidirectional connections between them in specific layers. Experiments show that CamLiFlow outperforms the previous art with fewer parameters.

\vspace{-10pt}
\paragraph{Acknowledgements.} This work is supported by National Natural Science Foundation of China (No.62076119, No.61921006), Program for Innovative Talents, Entrepreneur in Jiangsu Province, and Collaborative Innovation Center of Novel Software Technology and Industrialization.

%%%%%%%%% REFERENCES
{\small
\bibliographystyle{ieee_fullname}
\bibliography{egbib}

\begin{thebibliography}{10}\itemsep=-1pt

\bibitem{badki2021binaryttc}
Abhishek Badki, Orazio Gallo, Jan Kautz, and Pradeep Sen.
\newblock Binary ttc: A temporal geofence for autonomous navigation.
\newblock In {\em Proceedings of the IEEE/CVF Conference on Computer Vision and
  Pattern Recognition}, pages 12946--12955, 2021.

\bibitem{battrawy2019lidarflow}
Ramy Battrawy, Ren{\'e} Schuster, Oliver Wasenm{\"u}ller, Qing Rao, and Didier
  Stricker.
\newblock Lidar-flow: Dense scene flow estimation from sparse lidar and stereo
  images.
\newblock In {\em 2019 IEEE/RSJ International Conference on Intelligent Robots
  and Systems (IROS)}, pages 7762--7769. IEEE, 2019.

\bibitem{behl2017isf}
Aseem Behl, Omid Hosseini~Jafari, Siva Karthik~Mustikovela, Hassan Abu~Alhaija,
  Carsten Rother, and Andreas Geiger.
\newblock Bounding boxes, segmentations and object coordinates: How important
  is recognition for 3d scene flow estimation in autonomous driving scenarios?
\newblock In {\em Proceedings of the IEEE International Conference on Computer
  Vision}, pages 2574--2583, 2017.

\bibitem{black1996robust}
Michael~J Black and Paul Anandan.
\newblock The robust estimation of multiple motions: Parametric and
  piecewise-smooth flow fields.
\newblock {\em Computer vision and image understanding}, 63(1):75--104, 1996.

\bibitem{brachmann2019ransac}
Eric Brachmann and Carsten Rother.
\newblock Neural-guided ransac: Learning where to sample model hypotheses.
\newblock In {\em Proceedings of the IEEE/CVF International Conference on
  Computer Vision}, pages 4322--4331, 2019.

\bibitem{brox2009large}
Thomas Brox, Christoph Bregler, and Jitendra Malik.
\newblock Large displacement optical flow.
\newblock In {\em 2009 IEEE Conference on Computer Vision and Pattern
  Recognition}, pages 41--48. IEEE, 2009.

\bibitem{brox2004warping}
Thomas Brox, Andr{\'e}s Bruhn, Nils Papenberg, and Joachim Weickert.
\newblock High accuracy optical flow estimation based on a theory for warping.
\newblock In {\em European conference on computer vision}, pages 25--36.
  Springer, 2004.

\bibitem{bruhn2005lucas}
Andr{\'e}s Bruhn, Joachim Weickert, and Christoph Schn{\"o}rr.
\newblock Lucas/kanade meets horn/schunck: Combining local and global optic
  flow methods.
\newblock {\em International journal of computer vision}, 61(3):211--231, 2005.

\bibitem{chang2015shapenet}
Angel~X Chang, Thomas Funkhouser, Leonidas Guibas, Pat Hanrahan, Qixing Huang,
  Zimo Li, Silvio Savarese, Manolis Savva, Shuran Song, Hao Su, et~al.
\newblock Shapenet: An information-rich 3d model repository.
\newblock {\em arXiv preprint arXiv:1512.03012}, 2015.

\bibitem{chen2017multi}
Xiaozhi Chen, Huimin Ma, Ji Wan, Bo Li, and Tian Xia.
\newblock Multi-view 3d object detection network for autonomous driving.
\newblock In {\em Proceedings of the IEEE conference on Computer Vision and
  Pattern Recognition}, pages 1907--1915, 2017.

\bibitem{cordts2016cityscapes}
Marius Cordts, Mohamed Omran, Sebastian Ramos, Timo Rehfeld, Markus Enzweiler,
  Rodrigo Benenson, Uwe Franke, Stefan Roth, and Bernt Schiele.
\newblock The cityscapes dataset for semantic urban scene understanding.
\newblock In {\em Proceedings of the IEEE conference on computer vision and
  pattern recognition}, pages 3213--3223, 2016.

\bibitem{dosovitskiy2015flownet}
Alexey Dosovitskiy, Philipp Fischer, Eddy Ilg, Philip Hausser, Caner Hazirbas,
  Vladimir Golkov, Patrick Van Der~Smagt, Daniel Cremers, and Thomas Brox.
\newblock Flownet: Learning optical flow with convolutional networks.
\newblock In {\em Proceedings of the IEEE international conference on computer
  vision}, pages 2758--2766, 2015.

\bibitem{feng2021advancing}
Ziyue Feng, Longlong Jing, Peng Yin, Yingli Tian, and Bing Li.
\newblock Advancing self-supervised monocular depth learning with sparse
  li{DAR}.
\newblock In {\em 5th Annual Conference on Robot Learning}, 2021.

\bibitem{gu2019hplflownet}
Xiuye Gu, Yijie Wang, Chongruo Wu, Yong~Jae Lee, and Panqu Wang.
\newblock Hplflownet: Hierarchical permutohedral lattice flownet for scene flow
  estimation on large-scale point clouds.
\newblock In {\em Proceedings of the IEEE/CVF Conference on Computer Vision and
  Pattern Recognition}, pages 3254--3263, 2019.

\bibitem{he2015resnet}
Kaiming He, Xiangyu Zhang, Shaoqing Ren, and Jian Sun.
\newblock Deep residual learning for image recognition, 2015.

\bibitem{hong2021ddrnet}
Yuanduo Hong, Huihui Pan, Weichao Sun, Yisong Jia, et~al.
\newblock Deep dual-resolution networks for real-time and accurate semantic
  segmentation of road scenes.
\newblock {\em arXiv preprint arXiv:2101.06085}, 2021.

\bibitem{horn1981determining}
Berthold~KP Horn and Brian~G Schunck.
\newblock Determining optical flow.
\newblock {\em Artificial intelligence}, 17(1-3):185--203, 1981.

\bibitem{huang2017densenet}
Gao Huang, Zhuang Liu, Laurens Van Der~Maaten, and Kilian~Q Weinberger.
\newblock Densely connected convolutional networks.
\newblock In {\em Proceedings of the IEEE conference on computer vision and
  pattern recognition}, pages 4700--4708, 2017.

\bibitem{hui2018liteflownet}
Tak-Wai Hui, Xiaoou Tang, and Chen~Change Loy.
\newblock Liteflownet: A lightweight convolutional neural network for optical
  flow estimation.
\newblock In {\em Proceedings of the IEEE conference on computer vision and
  pattern recognition}, pages 8981--8989, 2018.

\bibitem{hur2019iterative}
Junhwa Hur and Stefan Roth.
\newblock Iterative residual refinement for joint optical flow and occlusion
  estimation.
\newblock In {\em Proceedings of the IEEE/CVF Conference on Computer Vision and
  Pattern Recognition}, pages 5754--5763, 2019.

\bibitem{ilg2017flownet2}
Eddy Ilg, Nikolaus Mayer, Tonmoy Saikia, Margret Keuper, Alexey Dosovitskiy,
  and Thomas Brox.
\newblock Flownet 2.0: Evolution of optical flow estimation with deep networks.
\newblock In {\em Proceedings of the IEEE conference on computer vision and
  pattern recognition}, pages 2462--2470, 2017.

\bibitem{ioffe2015batchnorm}
Sergey Ioffe and Christian Szegedy.
\newblock Batch normalization: Accelerating deep network training by reducing
  internal covariate shift, 2015.

\bibitem{jaimez2015primal}
Mariano Jaimez, Mohamed Souiai, Javier Gonzalez-Jimenez, and Daniel Cremers.
\newblock A primal-dual framework for real-time dense rgb-d scene flow.
\newblock In {\em 2015 IEEE international conference on robotics and automation
  (ICRA)}, pages 98--104. IEEE, 2015.

\bibitem{jaimez2015motion}
Mariano Jaimez, Mohamed Souiai, J{\"o}rg St{\"u}ckler, Javier Gonzalez-Jimenez,
  and Daniel Cremers.
\newblock Motion cooperation: Smooth piece-wise rigid scene flow from rgb-d
  images.
\newblock In {\em 2015 International Conference on 3D Vision}, pages 64--72.
  IEEE, 2015.

\bibitem{jiang2019sense}
Huaizu Jiang, Deqing Sun, Varun Jampani, Zhaoyang Lv, Erik Learned-Miller, and
  Jan Kautz.
\newblock Sense: A shared encoder network for scene-flow estimation.
\newblock In {\em Proceedings of the IEEE/CVF International Conference on
  Computer Vision}, pages 3195--3204, 2019.

\bibitem{kingma2014adam}
Diederik~P Kingma and Jimmy Ba.
\newblock Adam: A method for stochastic optimization.
\newblock {\em arXiv preprint arXiv:1412.6980}, 2014.

\bibitem{kittenplon2021flowstep3d}
Yair Kittenplon, Yonina~C Eldar, and Dan Raviv.
\newblock Flowstep3d: Model unrolling for self-supervised scene flow
  estimation.
\newblock In {\em Proceedings of the IEEE/CVF Conference on Computer Vision and
  Pattern Recognition}, pages 4114--4123, 2021.

\bibitem{li2021acosf}
Congcong Li, Haoyu Ma, and Qingmin Liao.
\newblock Two-stage adaptive object scene flow using hybrid cnn-crf model.
\newblock In {\em 2020 25th International Conference on Pattern Recognition
  (ICPR)}, pages 3876--3883. IEEE, 2021.

\bibitem{liang2018continuous}
Ming Liang, Bin Yang, Shenlong Wang, and Raquel Urtasun.
\newblock Deep continuous fusion for multi-sensor 3d object detection.
\newblock In {\em Proceedings of the European Conference on Computer Vision
  (ECCV)}, pages 641--656, 2018.

\bibitem{liu2019flownet3d}
Xingyu Liu, Charles~R Qi, and Leonidas~J Guibas.
\newblock Flownet3d: Learning scene flow in 3d point clouds.
\newblock In {\em Proceedings of the IEEE/CVF Conference on Computer Vision and
  Pattern Recognition}, pages 529--537, 2019.

\bibitem{liu2019meteornet}
Xingyu Liu, Mengyuan Yan, and Jeannette Bohg.
\newblock Meteornet: Deep learning on dynamic 3d point cloud sequences.
\newblock In {\em Proceedings of the IEEE/CVF International Conference on
  Computer Vision}, pages 9246--9255, 2019.

\bibitem{ma2018sparse}
Fangchang Ma and Sertac Karaman.
\newblock Sparse-to-dense: Depth prediction from sparse depth samples and a
  single image.
\newblock In {\em 2018 IEEE international conference on robotics and automation
  (ICRA)}, pages 4796--4803. IEEE, 2018.

\bibitem{ma2019drisf}
Wei-Chiu Ma, Shenlong Wang, Rui Hu, Yuwen Xiong, and Raquel Urtasun.
\newblock Deep rigid instance scene flow.
\newblock In {\em Proceedings of the IEEE/CVF Conference on Computer Vision and
  Pattern Recognition}, pages 3614--3622, 2019.

\bibitem{mayer2016things3d}
Nikolaus Mayer, Eddy Ilg, Philip Hausser, Philipp Fischer, Daniel Cremers,
  Alexey Dosovitskiy, and Thomas Brox.
\newblock A large dataset to train convolutional networks for disparity,
  optical flow, and scene flow estimation.
\newblock In {\em Proceedings of the IEEE conference on computer vision and
  pattern recognition}, pages 4040--4048, 2016.

\bibitem{menze2015osf}
Moritz Menze and Andreas Geiger.
\newblock Object scene flow for autonomous vehicles.
\newblock In {\em Proceedings of the IEEE conference on computer vision and
  pattern recognition}, pages 3061--3070, 2015.

\bibitem{paszke2019pytorch}
Adam Paszke, Sam Gross, Francisco Massa, Adam Lerer, James Bradbury, Gregory
  Chanan, Trevor Killeen, Zeming Lin, Natalia Gimelshein, Luca Antiga, et~al.
\newblock Pytorch: An imperative style, high-performance deep learning library.
\newblock {\em Advances in neural information processing systems},
  32:8026--8037, 2019.

\bibitem{puy2020flot}
Gilles Puy, Alexandre Boulch, and Renaud Marlet.
\newblock Flot: Scene flow on point clouds guided by optimal transport.
\newblock In {\em Computer Vision--ECCV 2020: 16th European Conference,
  Glasgow, UK, August 23--28, 2020, Proceedings, Part XXVIII 16}, pages
  527--544. Springer, 2020.

\bibitem{qi2018frustum}
Charles~R Qi, Wei Liu, Chenxia Wu, Hao Su, and Leonidas~J Guibas.
\newblock Frustum pointnets for 3d object detection from rgb-d data.
\newblock In {\em Proceedings of the IEEE conference on computer vision and
  pattern recognition}, pages 918--927, 2018.

\bibitem{qi2017pointnet}
Charles~R Qi, Hao Su, Kaichun Mo, and Leonidas~J Guibas.
\newblock Pointnet: Deep learning on point sets for 3d classification and
  segmentation.
\newblock In {\em Proceedings of the IEEE conference on computer vision and
  pattern recognition}, pages 652--660, 2017.

\bibitem{qi2017pointnet++}
Charles~R Qi, Li Yi, Hao Su, and Leonidas~J Guibas.
\newblock Pointnet++: Deep hierarchical feature learning on point sets in a
  metric space.
\newblock {\em arXiv preprint arXiv:1706.02413}, 2017.

\bibitem{quiroga2014dense}
Julian Quiroga, Thomas Brox, Fr{\'e}d{\'e}ric Devernay, and James Crowley.
\newblock Dense semi-rigid scene flow estimation from rgbd images.
\newblock In {\em European Conference on Computer Vision}, pages 567--582.
  Springer, 2014.

\bibitem{ranjan2017spynet}
Anurag Ranjan and Michael~J Black.
\newblock Optical flow estimation using a spatial pyramid network.
\newblock In {\em Proceedings of the IEEE conference on computer vision and
  pattern recognition}, pages 4161--4170, 2017.

\bibitem{ren2017ssf}
Zhile Ren, Deqing Sun, Jan Kautz, and Erik Sudderth.
\newblock Cascaded scene flow prediction using semantic segmentation.
\newblock In {\em 2017 International Conference on 3D Vision (3DV)}, pages
  225--233. IEEE, 2017.

\bibitem{rishav2020deeplidarflow}
Rishav Rishav, Ramy Battrawy, Ren{\'e} Schuster, Oliver Wasenm{\"u}ller, and
  Didier Stricker.
\newblock Deeplidarflow: A deep learning architecture for scene flow estimation
  using monocular camera and sparse lidar.
\newblock In {\em 2020 IEEE/RSJ International Conference on Intelligent Robots
  and Systems (IROS)}, pages 10460--10467. IEEE, 2020.

\bibitem{sun2018pwc}
Deqing Sun, Xiaodong Yang, Ming-Yu Liu, and Jan Kautz.
\newblock Pwc-net: Cnns for optical flow using pyramid, warping, and cost
  volume.
\newblock In {\em Proceedings of the IEEE conference on computer vision and
  pattern recognition}, pages 8934--8943, 2018.

\bibitem{teed2020raft}
Zachary Teed and Jia Deng.
\newblock Raft: Recurrent all-pairs field transforms for optical flow.
\newblock In {\em European conference on computer vision}, pages 402--419.
  Springer, 2020.

\bibitem{teed2021raft3d}
Zachary Teed and Jia Deng.
\newblock Raft-3d: Scene flow using rigid-motion embeddings.
\newblock In {\em Proceedings of the IEEE/CVF Conference on Computer Vision and
  Pattern Recognition}, pages 8375--8384, 2021.

\bibitem{vogel2015prsm}
Christoph Vogel, Konrad Schindler, and Stefan Roth.
\newblock 3d scene flow estimation with a piecewise rigid scene model.
\newblock {\em International Journal of Computer Vision}, 115(1):1--28, 2015.

\bibitem{vora2020pointpainting}
Sourabh Vora, Alex~H Lang, Bassam Helou, and Oscar Beijbom.
\newblock Pointpainting: Sequential fusion for 3d object detection.
\newblock In {\em Proceedings of the IEEE/CVF conference on computer vision and
  pattern recognition}, pages 4604--4612, 2020.

\bibitem{wang2020flownet3d++}
Zirui Wang, Shuda Li, Henry Howard-Jenkins, Victor Prisacariu, and Min Chen.
\newblock Flownet3d++: Geometric losses for deep scene flow estimation.
\newblock In {\em Proceedings of the IEEE/CVF Winter Conference on Applications
  of Computer Vision}, pages 91--98, 2020.

\bibitem{wei2021pvraft}
Yi Wei, Ziyi Wang, Yongming Rao, Jiwen Lu, and Jie Zhou.
\newblock Pv-raft: Point-voxel correlation fields for scene flow estimation of
  point clouds.
\newblock In {\em Proceedings of the IEEE/CVF Conference on Computer Vision and
  Pattern Recognition}, pages 6954--6963, 2021.

\bibitem{weinzaepfel2013deepflow}
Philippe Weinzaepfel, Jerome Revaud, Zaid Harchaoui, and Cordelia Schmid.
\newblock Deepflow: Large displacement optical flow with deep matching.
\newblock In {\em Proceedings of the IEEE international conference on computer
  vision}, pages 1385--1392, 2013.

\bibitem{wu2019pointconv}
Wenxuan Wu, Zhongang Qi, and Li Fuxin.
\newblock Pointconv: Deep convolutional networks on 3d point clouds.
\newblock In {\em Proceedings of the IEEE/CVF Conference on Computer Vision and
  Pattern Recognition}, pages 9621--9630, 2019.

\bibitem{wu2019pointpwc}
Wenxuan Wu, Zhiyuan Wang, Zhuwen Li, Wei Liu, and Li Fuxin.
\newblock Pointpwc-net: A coarse-to-fine network for supervised and
  self-supervised scene flow estimation on 3d point clouds.
\newblock {\em arXiv preprint arXiv:1911.12408}, 2019.

\bibitem{xu2018pointfusion}
Danfei Xu, Dragomir Anguelov, and Ashesh Jain.
\newblock Pointfusion: Deep sensor fusion for 3d bounding box estimation.
\newblock In {\em Proceedings of the IEEE conference on computer vision and
  pattern recognition}, pages 244--253, 2018.

\bibitem{yang2019volumetric}
Gengshan Yang and Deva Ramanan.
\newblock Volumetric correspondence networks for optical flow.
\newblock {\em Advances in neural information processing systems}, 32:794--805,
  2019.

\bibitem{yang2020opticalexp}
Gengshan Yang and Deva Ramanan.
\newblock Upgrading optical flow to 3d scene flow through optical expansion.
\newblock In {\em Proceedings of the IEEE/CVF Conference on Computer Vision and
  Pattern Recognition}, pages 1334--1343, 2020.

\bibitem{yang2021rigidmask}
Gengshan Yang and Deva Ramanan.
\newblock Learning to segment rigid motions from two frames.
\newblock In {\em Proceedings of the IEEE/CVF Conference on Computer Vision and
  Pattern Recognition}, pages 1266--1275, 2021.

\bibitem{you2019pseudo}
Yurong You, Yan Wang, Wei-Lun Chao, Divyansh Garg, Geoff Pleiss, Bharath
  Hariharan, Mark Campbell, and Kilian~Q Weinberger.
\newblock Pseudo-lidar++: Accurate depth for 3d object detection in autonomous
  driving.
\newblock {\em arXiv preprint arXiv:1906.06310}, 2019.

\bibitem{zhang2019ganet}
Feihu Zhang, Victor Prisacariu, Ruigang Yang, and Philip~HS Torr.
\newblock Ga-net: Guided aggregation net for end-to-end stereo matching.
\newblock In {\em Proceedings of the IEEE/CVF Conference on Computer Vision and
  Pattern Recognition}, pages 185--194, 2019.

\end{thebibliography}
}

\newpage
~
\newpage

\appendix
\section{Improving Training Stability}

\subsection{Gradient Detaching} Fusing two branches belonging to different modalities may encounter scale-mismatched gradients, making the training unstable and dominated by one modality. To solve the issue, we propose to detach the gradient from one branch to the other for each Bi-CLFM, as shown in Fig. \ref{fig:clfm} in the main paper.

\begin{figure}[t]
    \centering
    \begin{subfigure}[b]{0.5\linewidth}
        \includegraphics[width=\linewidth]{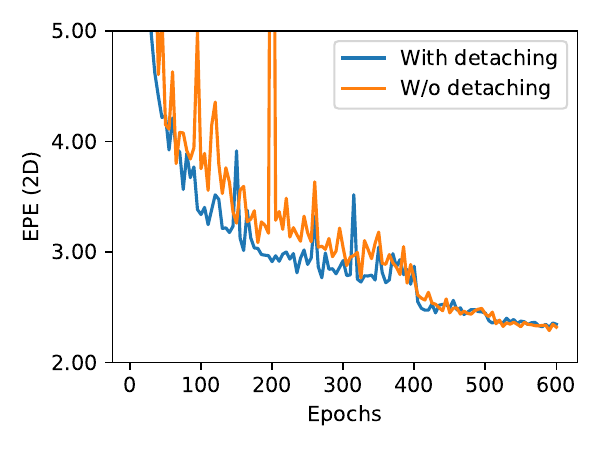}%
    \end{subfigure}%
    \hfill
    \begin{subfigure}[b]{0.5\linewidth}
        \includegraphics[width=\linewidth]{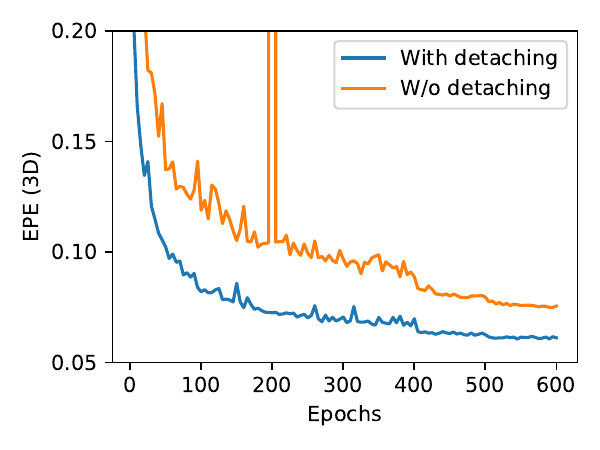}%
    \end{subfigure}%
    \vspace{-5pt}
    \caption{Performance on the ``val'' split of FlyingThings3D with/without our gradient detaching strategy. Detaching the gradient from one branch to the other can prevent one modality from dominating, so that each branch focuses on its own task.}
    \vspace{-5pt}
    \label{fig:detach}
\end{figure}

In Fig. \ref{fig:detach}, we conduct an ablation to demonstrate the effect of gradient detaching. Without gradient detaching, the image branch dominates the training and hurts the performance of the point branch. By detaching the gradient from one branch to the other, we prevent one modality from dominating so that each branch focuses on its task.

\subsection{Batch Normalization} Although the original implementation of PWC-Net \cite{sun2018pwc} does not use batch normalization (BN) \cite{ioffe2015batchnorm} in the whole network, we empirically find that integrating BN into the feature pyramid can improve training stability and speed up the convergence.

\section{Implementation Details}

\subsection{Feature Pyramid} 
\begin{figure*}
    \captionsetup[subfigure]{labelformat=empty}
    \includegraphics[width=\linewidth]{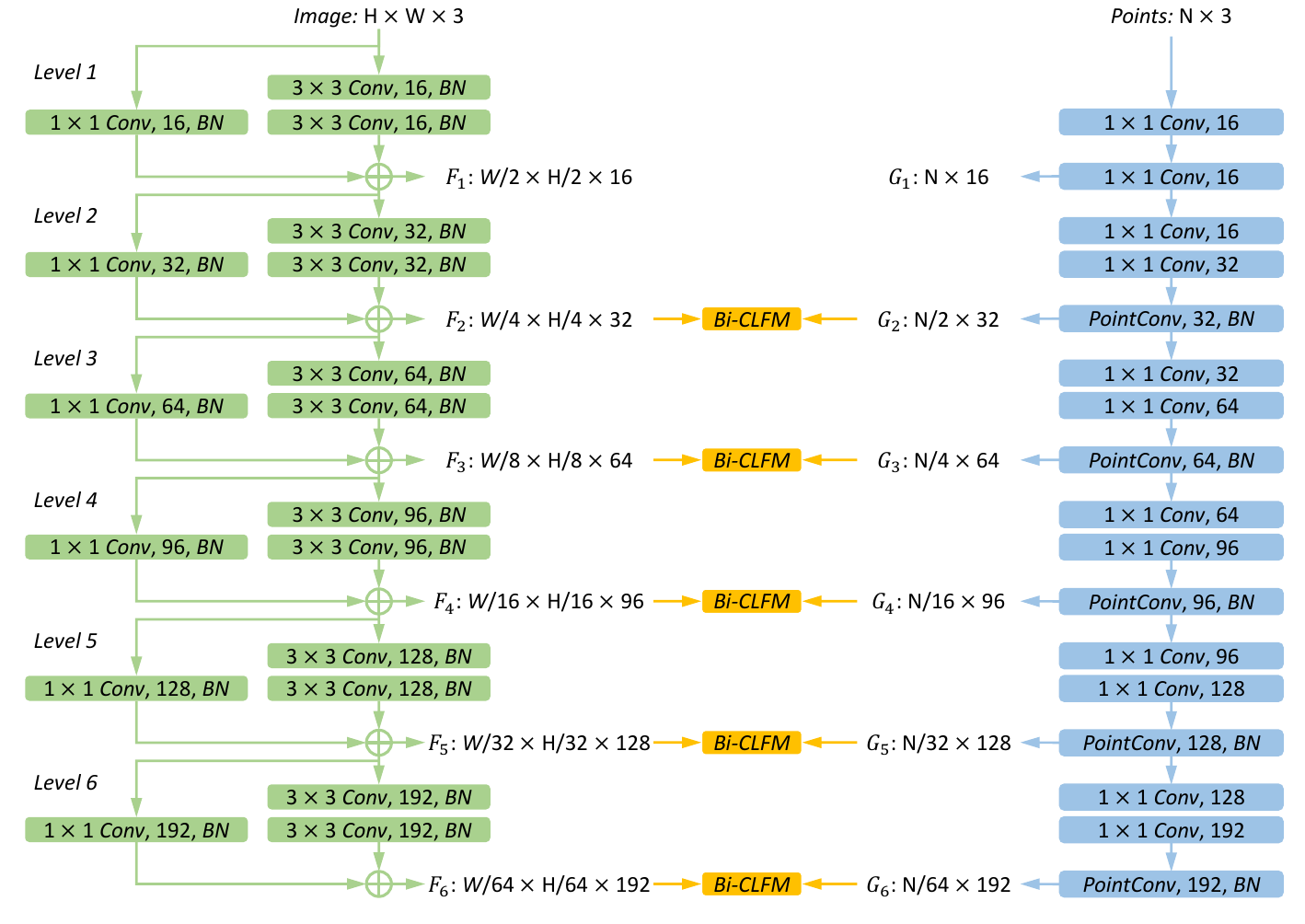}
    \caption{The feature pyramid of CamLiFlow. We build a feature pyramid for the image branch and the point branch respectively, with the top-level being the inputs. For the image branch, we extract features with residual blocks and perform batch normalization for each convolution layer. For the point branch, we build a six-level pyramid with a downsampling factor of 2 to match the levels of the image branch. Features from level 6 to level 2 are fused by a Bi-CLFM to pass complementary information.}
    \label{fig:pyramid}
\end{figure*}

As shown in Fig. \ref{fig:pyramid}, we generate a feature pyramid for the image branch and the point branch respectively, with the top-level being the inputs. Our feature pyramid follows the implementation of the original PWC-Net \cite{sun2018pwc} and PointPWC-Net \cite{wu2019pointpwc} but with some modifications. For the image branch, we replace the feedforward CNNs with residual blocks \cite{he2015resnet} and perform batch normalization for each convolutional layer. For the point branch, the original PointPWC-Net builds a three-level pyramid and downsamples the points by a factor of 4, while we build a six-level pyramid with a downsampling factor of 2 to match the levels of the image branch. Features from level 6 to level 2 are fused by a Bi-CLFM.

We start processing from the top level and perform the coarse-to-fine estimation scheme until level 2. Hence, CamLiFlow outputs optical flow at 1/4 resolution and scene flow at 1/2 resolution. We obtain full-scale optical flow and scene flow by convex upsampling \cite{teed2020raft} and $k$-NN upsampling respectively.

\subsection{Warping} For image features, warping can be directly implemented with \textit{bilinear grid sampling}. For point clouds, however, grid sampling no longer works since points do not conform to the regular grids as in images. In PointPWC-Net, the warping layer is simply an element-wise addition between the scene flow and the point clouds of the first frame. However, this does not work in our case, since PointPWC-Net warps the first frame towards the second frame, while our image branch warps the second frame towards the first frame. This mismatch can make it difficult for training to converge.

Hence, we warp the point clouds of the second frame towards the first frame using the backward scene flow, which can be computed using \textit{inverse distance weighted interpolation}. Formally, let $P_1^l$ and $P_2^l$ be the point clouds of the first frame and the second frame at level $l$ respectively. We first warp $P_1^l$ to the target frame using the upsampled scene flow from level $l+1$:
\begin{equation}
    \tilde{P}_1^l = P_1^l + \mathbf{U}(f_{3D}^{l+1}),
\end{equation}
where $\mathbf{U}$ denotes the upsampling operator based on $k$-NN interpolation and $f_{3D}^{l+1}$ denotes the coarse scene flow from level $l+1$. Then, for each point in $P_2^l$, we find its $k$ nearest neighbor ($k$-NN) in $\tilde{P}_1^l$. The backward flow $b_{3D}^l$ can be computed using inverse distance weighted interpolation:
\begin{equation}
    b_{3D}^l(i) = -\frac
    {\sum_{j=1}^k w_{ij} f_{3D}^{l+1}(j)}
    {\sum_{j=1}^k w_{ij}},
\end{equation}
where $w_{ij} = 1/d(P_2^l(i), \tilde{P}_1^l(j))$ and $d$ is a distance metric, i.e., Euclidean distance. Finally, $P_2^l$ can be warped to the first frame using the backward flow:
\begin{equation}
    P_w^l = P_2^l + b_{3D}^l.
\end{equation}

\subsection{Cost Volume} For the image branch, we follow PWC-Net to construct a partial cost volume by limiting the search range to a maximum displacement of $d_{max}=4$ pixels around each pixel. The resulting cost volume is organized as a 3D array of dimensions $D^2 \times H \times W$, where $H$ and $W$ are the height and width of the feature map respectively and $D = 2d_{max} + 1 = 9$. For the point branch, we follow PointPWC-Net to construct a learnable cost volume layer. Formally, for a point $p$, we form the point-to-patch cost by searching its $k$-NN neighborhoods in the target point cloud. Then the point-to-patch costs are further aggregated with PointConv to construct a patch-to-patch cost volume. The size of the neighborhoods is set to $k=16$ for all our experiments. Finally, we fuse the cost volumes of the two branches with a Bi-CLFM.

\subsection{Flow Decoder and Estimator} Our optical flow decoder follows the original PWC-Net which employs a multi-layer CNN with DenseNet \cite{huang2017densenet} connections. The numbers of the feature channels for each convolutional layer are 128, 128, 96, 64, and 32 respectively. Our scene flow estimator follows the original PointPWC-Net, which consists of two PointConv layers with 128 feature channels and a two-layer MLP with 128 and 64 feature channels respectively. Next, the outputs of the optical flow decoder and the scene flow decoder are fused by a Bi-CLFM. Finally, a single-layer perceptron (SLP) with two output channels serves as the optical flow estimator to estimate the optical flow at level $l$, and the scene flow estimator is an SLP with three output channels to estimate the scene flow at level $l$. For both the image branch and point branch, weights are shared across all pyramid levels.

\begin{figure*}
    \captionsetup[subfigure]{labelformat=empty}

    \begin{subfigure}[b]{0.02\linewidth}
        \centering
        \raisebox{3pt}{\rotatebox{90}{Input}}
    \end{subfigure}%
    \hfill
    \begin{subfigure}[b]{0.19\linewidth}
        \includegraphics[width=\linewidth]{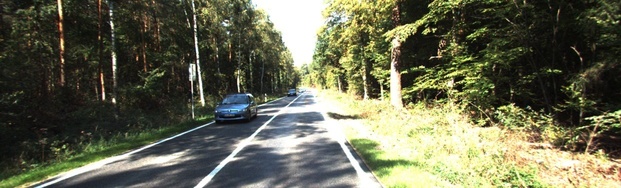}
    \end{subfigure}%
    \hfill
    \begin{subfigure}[b]{0.19\linewidth}
        \includegraphics[width=\linewidth]{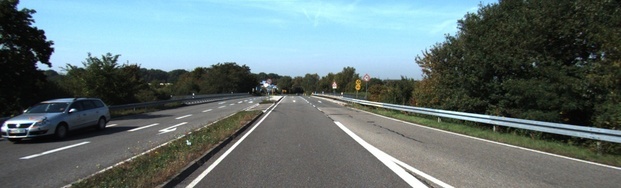}
    \end{subfigure}%
    \hfill
    \begin{subfigure}[b]{0.19\linewidth}
        \includegraphics[width=\linewidth]{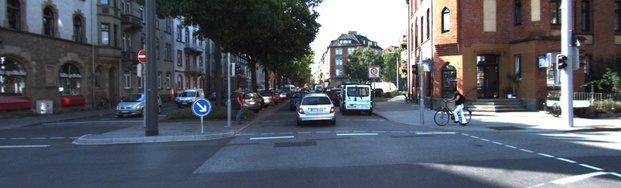}
    \end{subfigure}%
    \hfill
    \begin{subfigure}[b]{0.19\linewidth}
        \includegraphics[width=\linewidth]{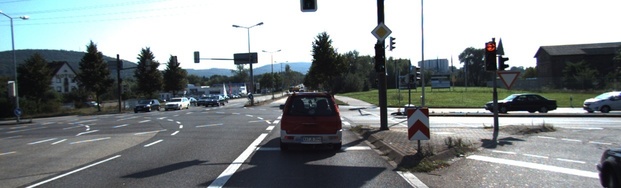}
    \end{subfigure}%
    \hfill
    \begin{subfigure}[b]{0.19\linewidth}
        \includegraphics[width=\linewidth]{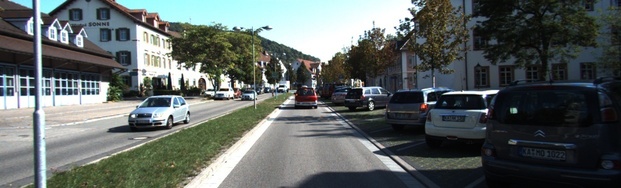}
    \end{subfigure}%
    \hfill
    \begin{subfigure}[b]{0.02\linewidth}
        \hspace{10pt}
    \end{subfigure}%
    \hfill

    \begin{subfigure}[b]{0.02\linewidth}
        \centering
        \raisebox{3pt}{\rotatebox{90}{Input}}
    \end{subfigure}%
    \hfill
    \begin{subfigure}[b]{0.19\linewidth}
        \includegraphics[width=\linewidth]{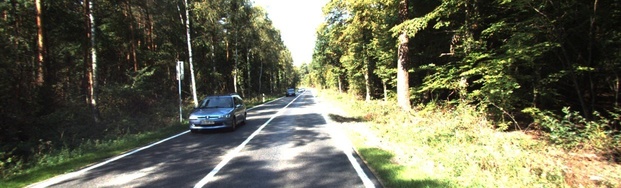}
    \end{subfigure}%
    \hfill
    \begin{subfigure}[b]{0.19\linewidth}
        \includegraphics[width=\linewidth]{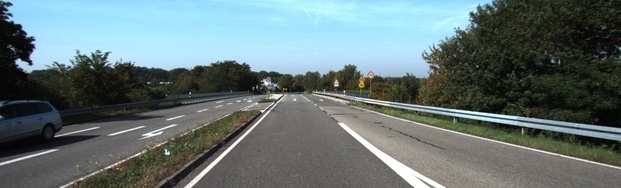}
    \end{subfigure}%
    \hfill
    \begin{subfigure}[b]{0.19\linewidth}
        \includegraphics[width=\linewidth]{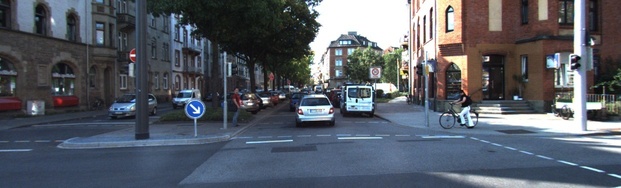}
    \end{subfigure}%
    \hfill
    \begin{subfigure}[b]{0.19\linewidth}
        \includegraphics[width=\linewidth]{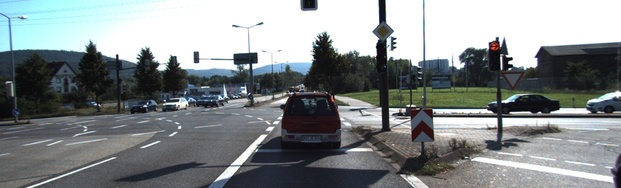}
    \end{subfigure}%
    \hfill
    \begin{subfigure}[b]{0.19\linewidth}
        \includegraphics[width=\linewidth]{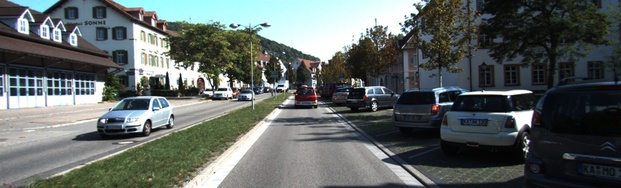}
    \end{subfigure}%
    \hfill
    \begin{subfigure}[b]{0.02\linewidth}
        \hspace{10pt}
    \end{subfigure}%
    \hfill

    \begin{subfigure}[b]{0.02\linewidth}
        \centering
        \raisebox{8pt}{\rotatebox{90}{(a)}}
    \end{subfigure}%
    \hfill
    \begin{subfigure}[b]{0.19\linewidth}
        \includegraphics[width=\linewidth]{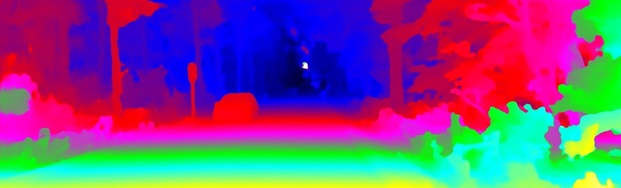}
    \end{subfigure}%
    \hfill
    \begin{subfigure}[b]{0.19\linewidth}
        \includegraphics[width=\linewidth]{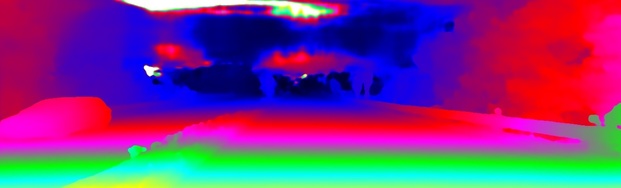}
    \end{subfigure}%
    \hfill
    \begin{subfigure}[b]{0.19\linewidth}
        \includegraphics[width=\linewidth]{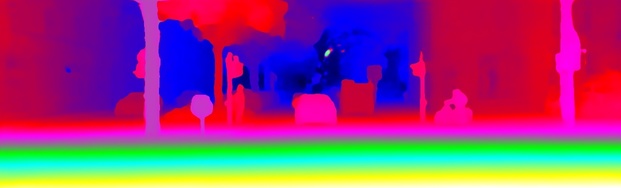}
    \end{subfigure}%
    \hfill
    \begin{subfigure}[b]{0.19\linewidth}
        \includegraphics[width=\linewidth]{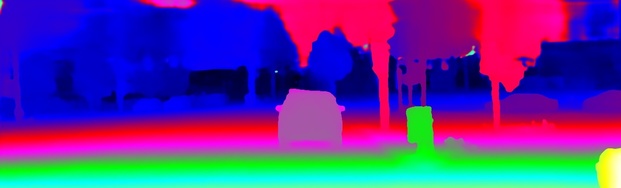}
    \end{subfigure}%
    \hfill
    \begin{subfigure}[b]{0.19\linewidth}
        \includegraphics[width=\linewidth]{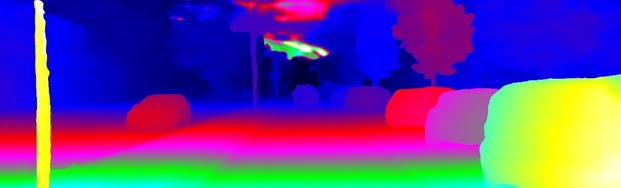}
    \end{subfigure}%
    \hfill
    \begin{subfigure}[b]{0.02\linewidth}
        \hspace{10pt}
    \end{subfigure}%
    \hfill

    \begin{subfigure}[b]{0.02\linewidth}
        \centering
        \raisebox{8pt}{\rotatebox{90}{(b)}}
    \end{subfigure}%
    \hfill
    \begin{subfigure}[b]{0.19\linewidth}
        \includegraphics[width=\linewidth]{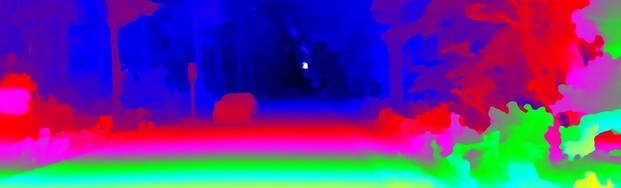}
    \end{subfigure}%
    \hfill
    \begin{subfigure}[b]{0.19\linewidth}
        \includegraphics[width=\linewidth]{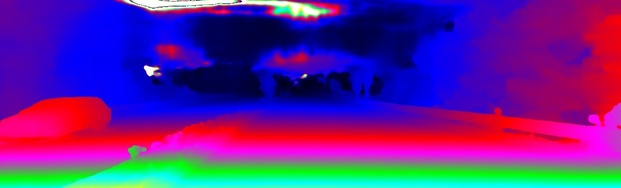}
    \end{subfigure}%
    \hfill
    \begin{subfigure}[b]{0.19\linewidth}
        \includegraphics[width=\linewidth]{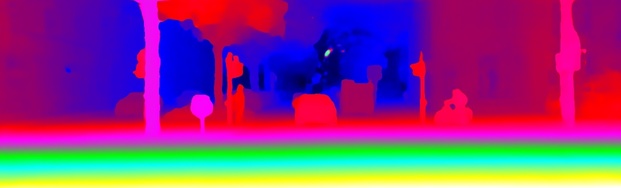}
    \end{subfigure}%
    \hfill
    \begin{subfigure}[b]{0.19\linewidth}
        \includegraphics[width=\linewidth]{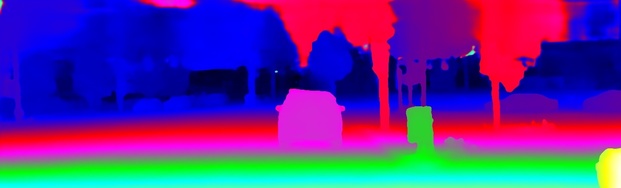}
    \end{subfigure}%
    \hfill
    \begin{subfigure}[b]{0.19\linewidth}
        \includegraphics[width=\linewidth]{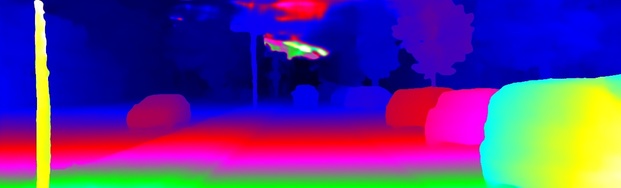}
    \end{subfigure}%
    \hfill
    \begin{subfigure}[b]{0.02\linewidth}
        \hspace{10pt}
    \end{subfigure}%
    \hfill

    \begin{subfigure}[b]{0.02\linewidth}
        \centering
        \raisebox{8pt}{\rotatebox{90}{(c)}}
    \end{subfigure}%
    \hfill
    \begin{subfigure}[b]{0.19\linewidth}
        \includegraphics[width=\linewidth]{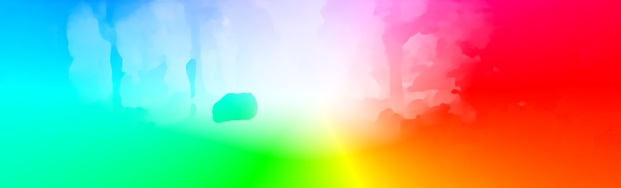}
    \end{subfigure}%
    \hfill
    \begin{subfigure}[b]{0.19\linewidth}
        \includegraphics[width=\linewidth]{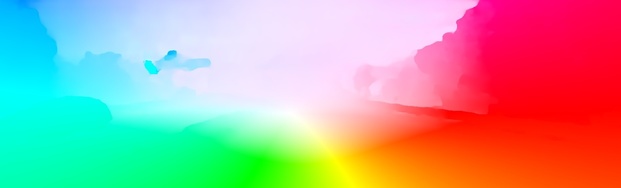}
    \end{subfigure}%
    \hfill
    \begin{subfigure}[b]{0.19\linewidth}
        \includegraphics[width=\linewidth]{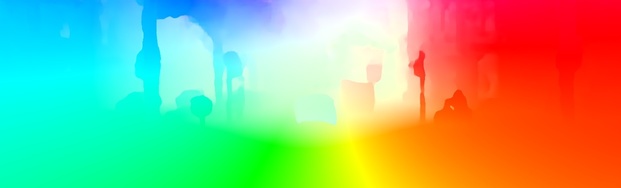}
    \end{subfigure}%
    \hfill
    \begin{subfigure}[b]{0.19\linewidth}
        \includegraphics[width=\linewidth]{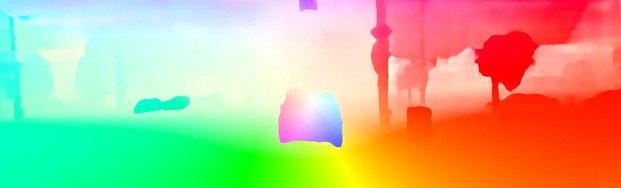}
    \end{subfigure}%
    \hfill
    \begin{subfigure}[b]{0.19\linewidth}
        \includegraphics[width=\linewidth]{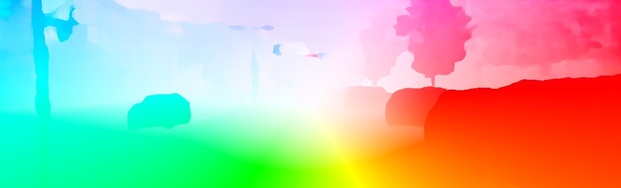}
    \end{subfigure}%
    \hfill
    \begin{subfigure}[b]{0.02\linewidth}
        \hspace{10pt}
    \end{subfigure}%
    \hfill

    \begin{subfigure}[b]{0.02\linewidth}
        \centering
        \raisebox{8pt}{\rotatebox{90}{(d)}}
    \end{subfigure}%
    \hfill
    \begin{subfigure}[b]{0.19\linewidth}
        \includegraphics[width=\linewidth]{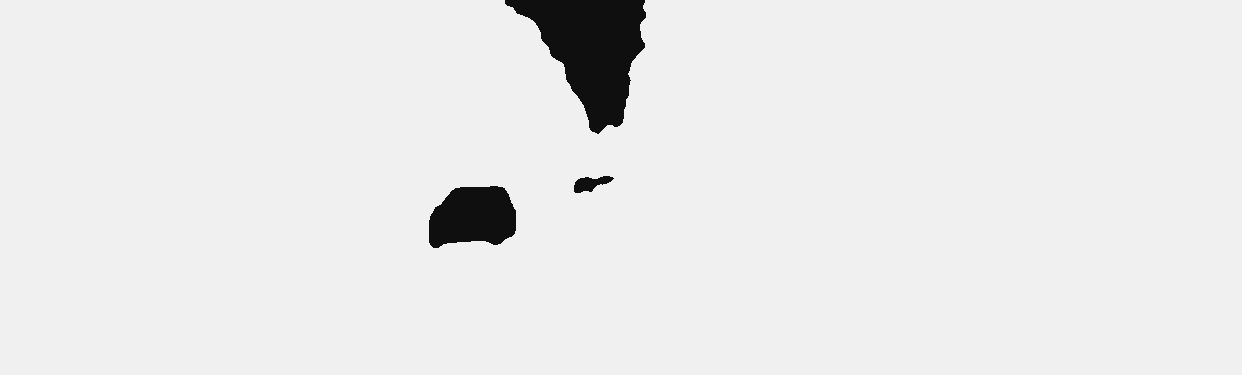}
    \end{subfigure}%
    \hfill
    \begin{subfigure}[b]{0.19\linewidth}
        \includegraphics[width=\linewidth]{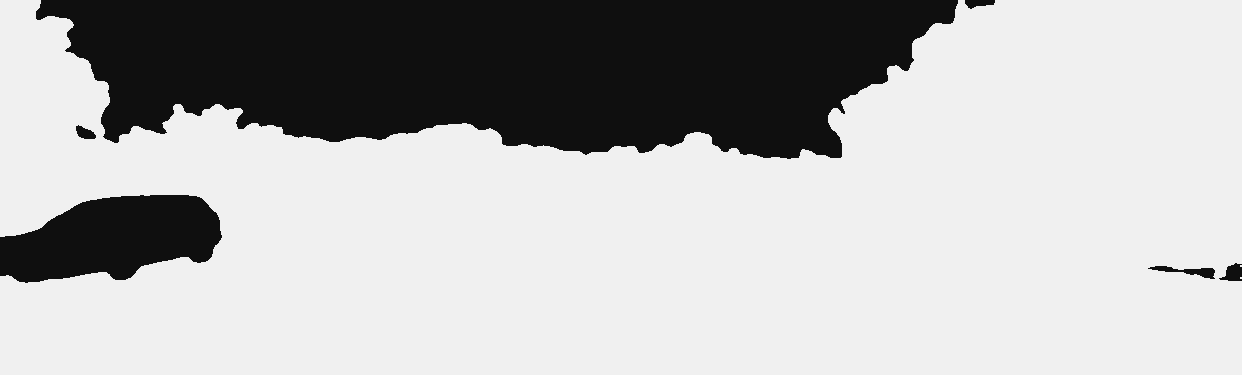}
    \end{subfigure}%
    \hfill
    \begin{subfigure}[b]{0.19\linewidth}
        \includegraphics[width=\linewidth]{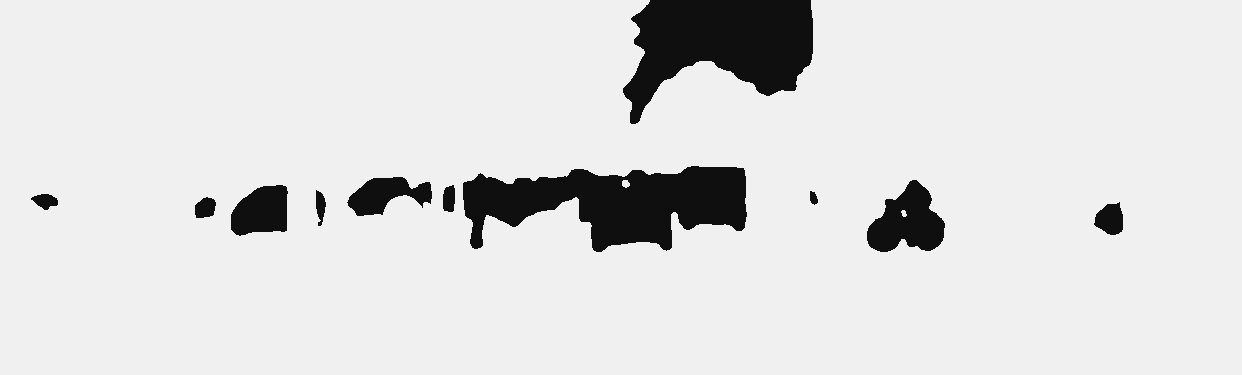}
    \end{subfigure}%
    \hfill
    \begin{subfigure}[b]{0.19\linewidth}
        \includegraphics[width=\linewidth]{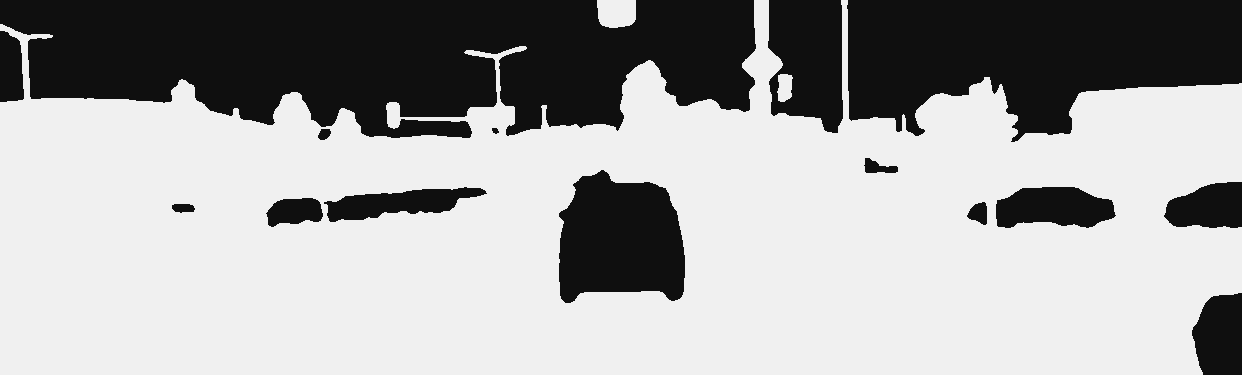}
    \end{subfigure}%
    \hfill
    \begin{subfigure}[b]{0.19\linewidth}
        \includegraphics[width=\linewidth]{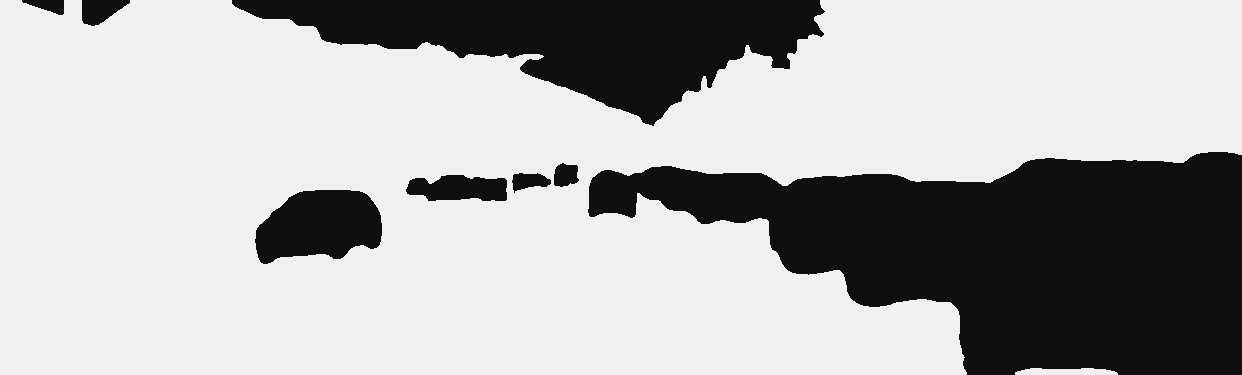}
    \end{subfigure}%
    \hfill
    \begin{subfigure}[b]{0.02\linewidth}
        \hspace{10pt}
    \end{subfigure}%
    \hfill

    \begin{subfigure}[b]{0.02\linewidth}
        \centering
        \raisebox{8pt}{\rotatebox{90}{(e)}}
    \end{subfigure}%
    \hfill
    \begin{subfigure}[b]{0.19\linewidth}
        \includegraphics[width=\linewidth]{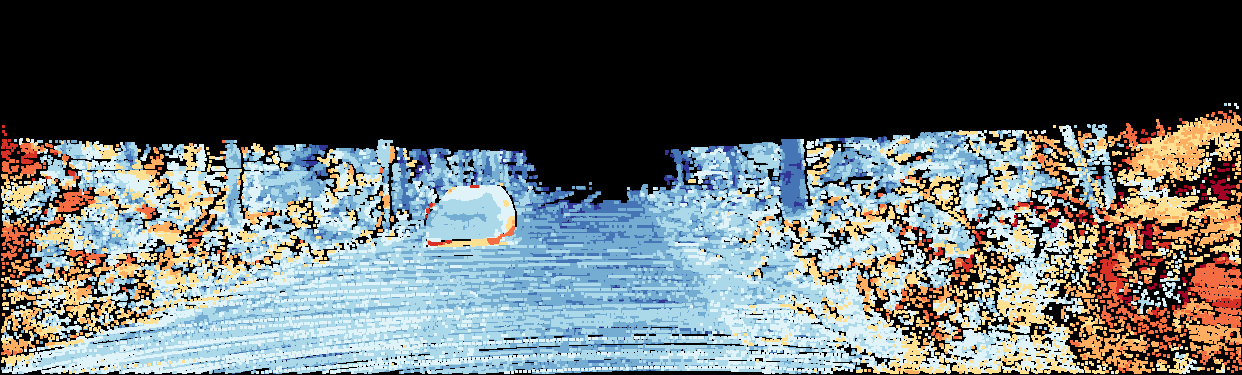}
    \end{subfigure}%
    \hfill
    \begin{subfigure}[b]{0.19\linewidth}
        \includegraphics[width=\linewidth]{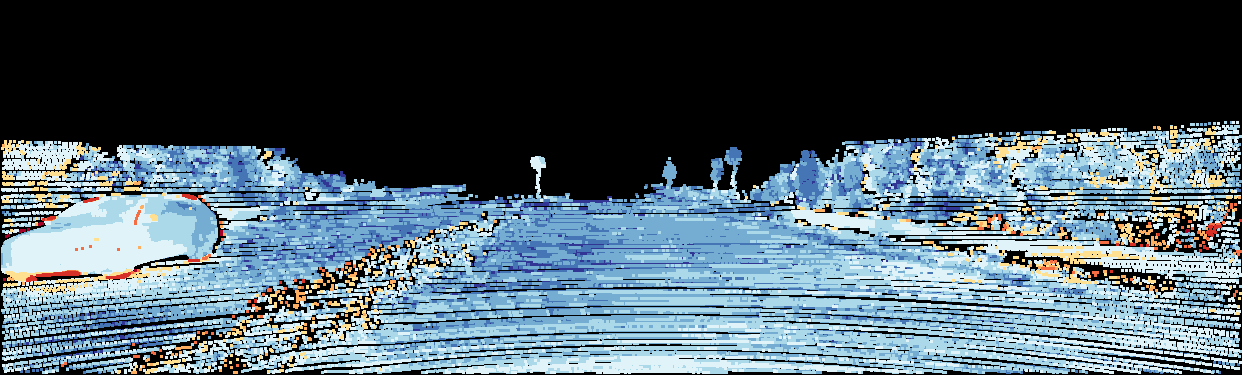}
    \end{subfigure}%
    \hfill
    \begin{subfigure}[b]{0.19\linewidth}
        \includegraphics[width=\linewidth]{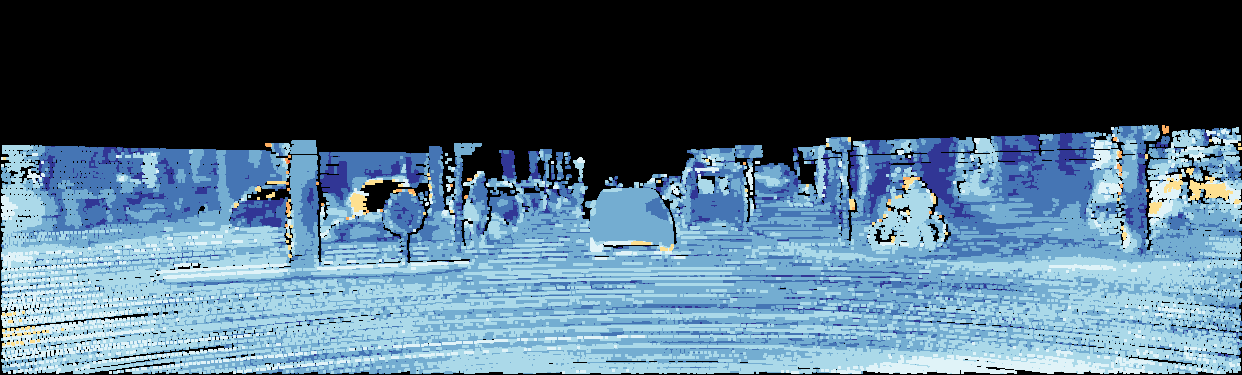}
    \end{subfigure}%
    \hfill
    \begin{subfigure}[b]{0.19\linewidth}
        \includegraphics[width=\linewidth]{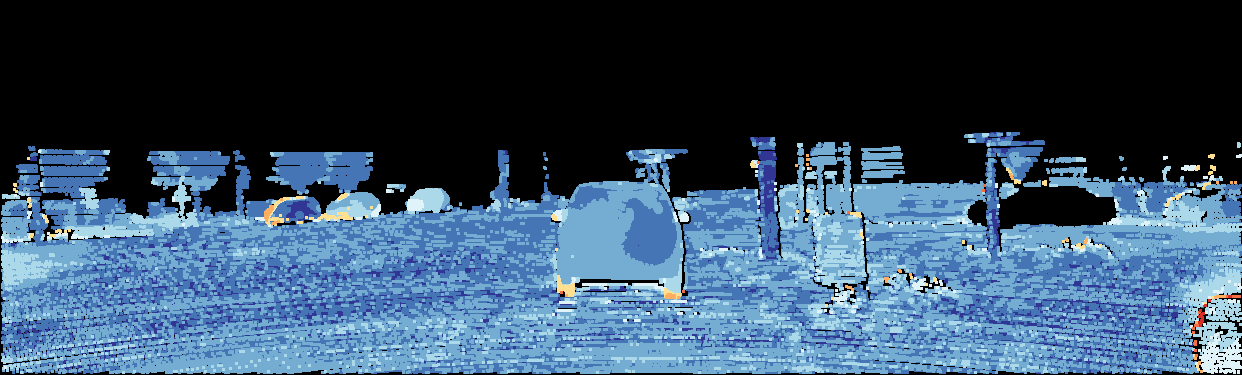}
    \end{subfigure}%
    \hfill
    \begin{subfigure}[b]{0.19\linewidth}
        \includegraphics[width=\linewidth]{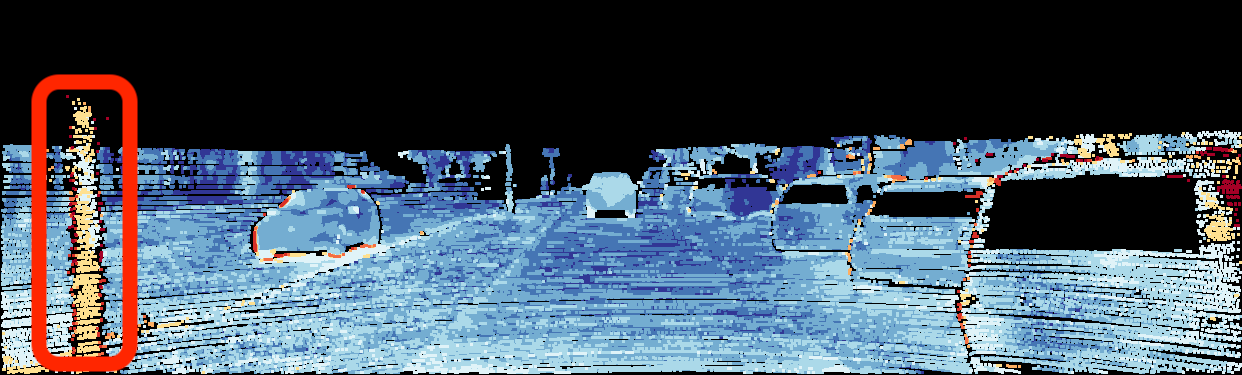}
    \end{subfigure}%
    \hfill
    \begin{subfigure}[b]{0.02\linewidth}
        \hspace{10pt}
    \end{subfigure}%
    \hfill

    \begin{subfigure}[b]{0.02\linewidth}
        \centering
        \raisebox{8pt}{\rotatebox{90}{(f)}}
    \end{subfigure}%
    \hfill
    \begin{subfigure}[b]{0.19\linewidth}
        \includegraphics[width=\linewidth]{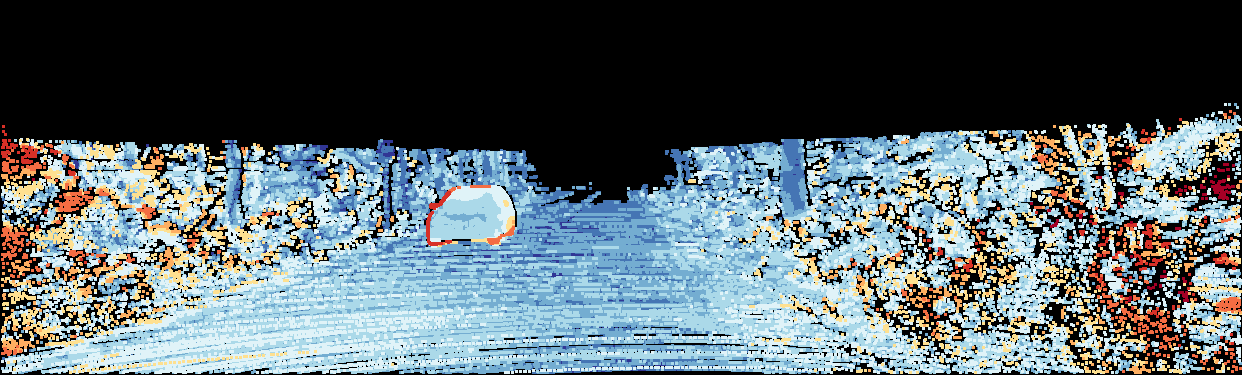}
    \end{subfigure}%
    \hfill
    \begin{subfigure}[b]{0.19\linewidth}
        \includegraphics[width=\linewidth]{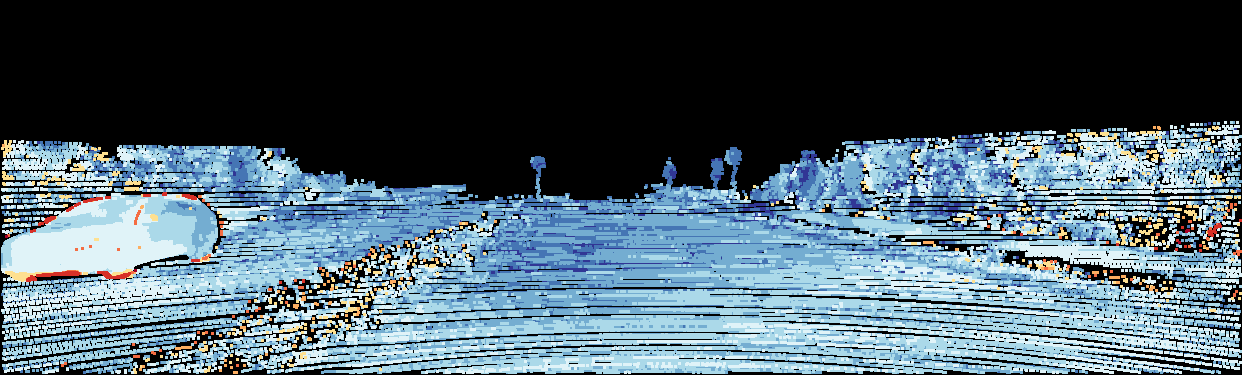}
    \end{subfigure}%
    \hfill
    \begin{subfigure}[b]{0.19\linewidth}
        \includegraphics[width=\linewidth]{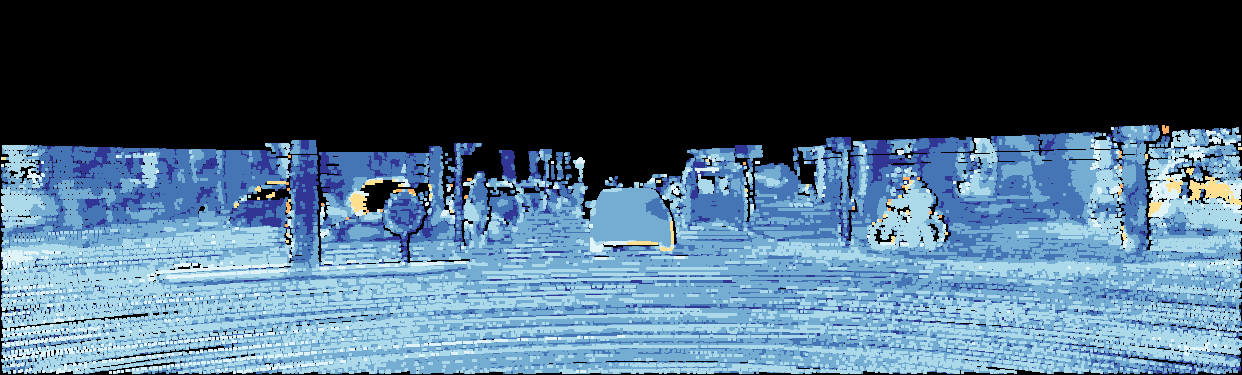}
    \end{subfigure}%
    \hfill
    \begin{subfigure}[b]{0.19\linewidth}
        \includegraphics[width=\linewidth]{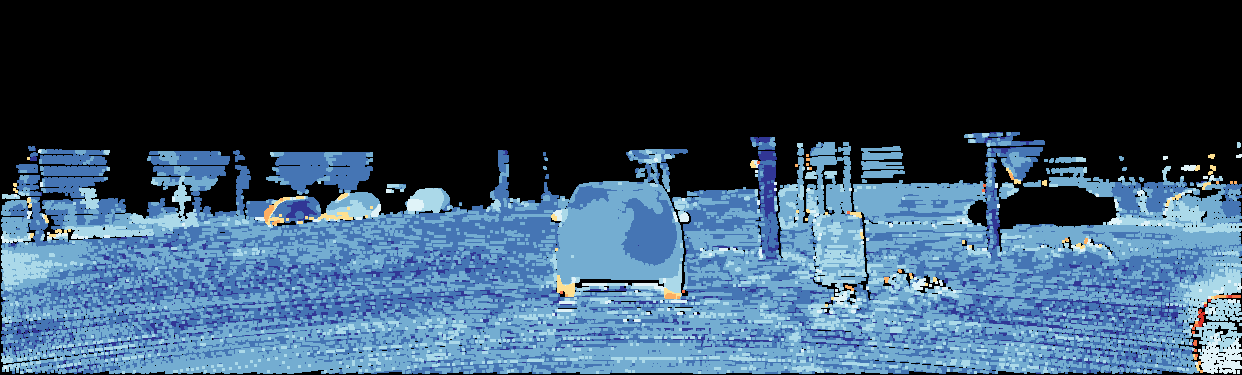}
    \end{subfigure}%
    \hfill
    \begin{subfigure}[b]{0.19\linewidth}
        \includegraphics[width=\linewidth]{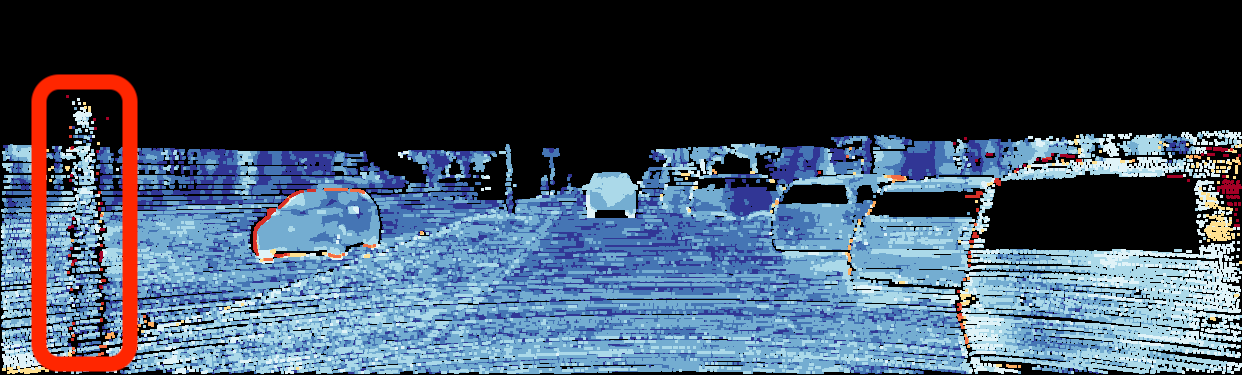}
    \end{subfigure}%
    \hfill
    \begin{subfigure}[b]{0.02\linewidth}
        \hspace{10pt}
    \end{subfigure}%
    \hfill

    \vspace{-3pt}
    \begin{subfigure}[b]{0.02\linewidth}
        \centering
    \end{subfigure}%
    \hfill
    \begin{subfigure}[b]{0.956\linewidth}
        \includegraphics[width=\linewidth]{images/kitti-error-bar.pdf}
    \end{subfigure}%
    \hfill
    \begin{subfigure}[b]{0.02\linewidth}
        \centering
    \end{subfigure}%
    \hfill

    \caption{Qualitative results on the validation set of the KITTI Scene Flow dataset: (a) disparity of the first frame by GA-Net \cite{zhang2019ganet}, (b) warped disparity of the second frame by our point branch, (c) optical flow by our image branch, (d) background segmentation by DDRNet-Slim \cite{hong2021ddrnet}, (e, f) scene flow error map with/without the background rigidity refinement.}
    \label{fig:more-kitti-val}
\end{figure*}

\section{Training Details}

\begin{figure}[t]
    \begin{subfigure}[b]{0.499\linewidth}
        \includegraphics[width=\linewidth]{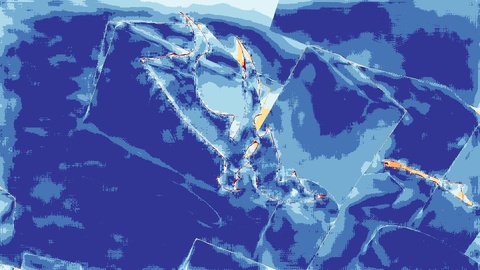}
    \end{subfigure}%%
    \hfill
    \begin{subfigure}[b]{0.499\linewidth}
        \includegraphics[width=\linewidth]{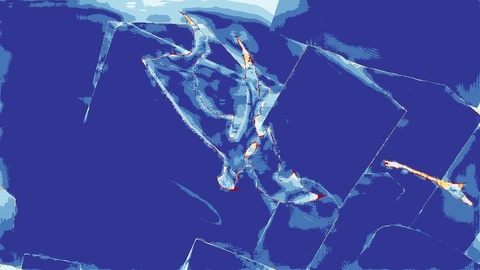}
    \end{subfigure}%%
    
    \vspace{-4.5pt}
    \begin{subfigure}[b]{\linewidth}
        \includegraphics[width=\linewidth]{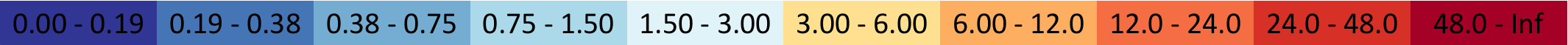}
    \end{subfigure}%%
    \vspace{3pt}

    \begin{subfigure}[b]{0.499\linewidth}
        \caption{W/o Finetune}
    \end{subfigure}%%
    \hfill
    \begin{subfigure}[b]{0.499\linewidth}
        \caption{With Finetune}
    \end{subfigure}%%

    \vspace{-5pt}
    \caption{Visualized error map of optical flow on FlyingThings3D with/without finetuning using the robust loss function. Fine-tuning with the robust loss fusion gives less penalty to outliers and can improve the threshold metrics.}
    \vspace{-5pt}

    \label{fig:things-finetune}
\end{figure}

\paragraph{FlyingThings3D.} We follow FlowNet3D \cite{liu2019flownet3d} to lift the disparity maps to point clouds with depth $<$35m. For data augmentation, we only perform random flipping (for both horizontal and vertical directions) since the number of the training samples is enough. We train our model with the $L_2$-norm loss function and then fine-tune it on the same dataset with the robust loss function. Fine-tuning with the robust loss fusion gives less penalty to outliers and can improve the threshold metrics (ACC\textsubscript{1px} and ACC\textsubscript{.05}). In Fig. \ref{fig:things-finetune}, we provide a visualized comparison of the error map to demonstrate the effect.

\paragraph{KITTI.} Using the weight pre-trained on FlyingThings3D, we finetune our model on Driving and KITTI in sequence. Basic data augmentation strategies including color jitter, random horizontal flipping, random scaling, and random cropping are applied. We use the \texttt{ColorJitter} from Torchvision \cite{paszke2019pytorch} with brightness 0.4, contrast 0.4, saturation 0.2, and hue $0.4/\pi$. For Driving, we randomly crop the image with a size of 768x384. For KITTI, we randomly rescale the image by the factor in the range $[1.0, 1.5]$.

\section{Additional Qualitative Examples}

\begin{figure*}
    \captionsetup[subfigure]{labelformat=empty}
        \begin{subfigure}[b]{0.166\linewidth}
        \caption{Reference Frame}
        \end{subfigure}%
        \hfill
        \begin{subfigure}[b]{0.166\linewidth}
        \caption{W/o Fusion}
        \end{subfigure}%
        \hfill
        \begin{subfigure}[b]{0.166\linewidth}
        \caption{+ Pyramid Fusion}
        \end{subfigure}%
        \hfill
        \begin{subfigure}[b]{0.166\linewidth}
        \caption{+ Cost Volume Fusion}
        \end{subfigure}%
        \hfill
        \begin{subfigure}[b]{0.166\linewidth}
        \caption{+ Decoder Fusion}
        \end{subfigure}%
        \hfill
        \begin{subfigure}[b]{0.166\linewidth}
        \caption{Ground Truth}
        \end{subfigure}%
        \hfill
    \vspace{-3pt}

    \begin{subfigure}[b]{0.166\linewidth}
    \includegraphics[width=\linewidth]{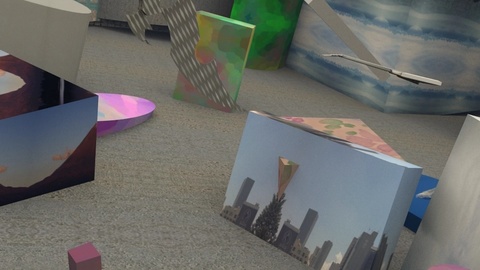}
    \end{subfigure}%%
    \hfill
    \begin{subfigure}[b]{0.166\linewidth}
    \includegraphics[width=\linewidth]{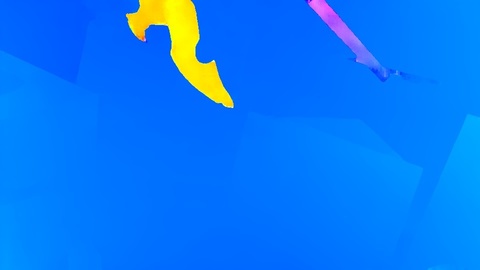}
    \end{subfigure}%%
    \hfill
    \begin{subfigure}[b]{0.166\linewidth}
    \includegraphics[width=\linewidth]{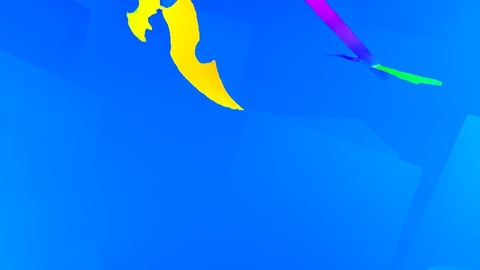}
    \end{subfigure}%%
    \hfill
    \begin{subfigure}[b]{0.166\linewidth}
    \includegraphics[width=\linewidth]{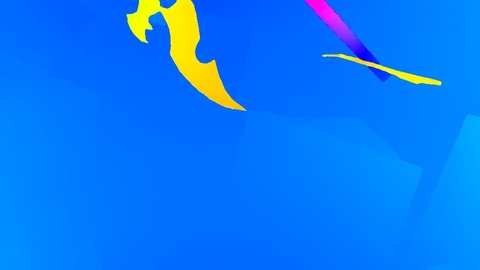}
    \end{subfigure}%%
    \hfill
    \begin{subfigure}[b]{0.166\linewidth}
    \includegraphics[width=\linewidth]{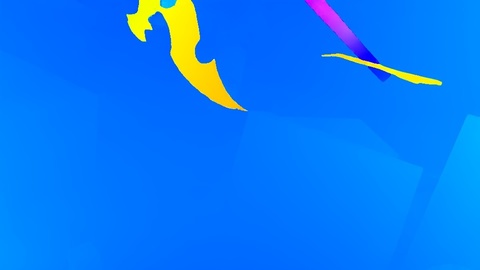}
    \end{subfigure}%%
    \hfill
    \begin{subfigure}[b]{0.166\linewidth}
    \includegraphics[width=\linewidth]{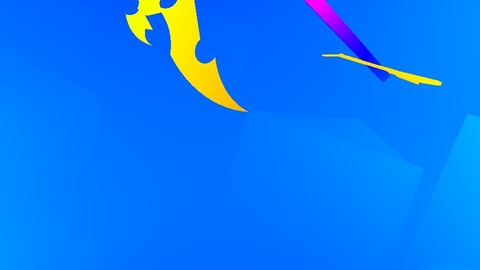}
    \end{subfigure}%%
    \hfill
    
    \begin{subfigure}[b]{0.166\linewidth}
    \includegraphics[width=\linewidth]{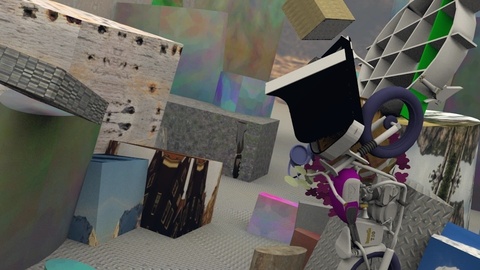}
    \end{subfigure}%%
    \hfill
    \begin{subfigure}[b]{0.166\linewidth}
    \includegraphics[width=\linewidth]{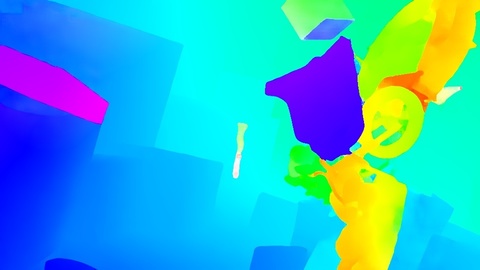}
    \end{subfigure}%%
    \hfill
    \begin{subfigure}[b]{0.166\linewidth}
    \includegraphics[width=\linewidth]{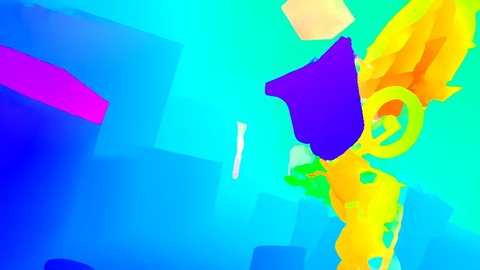}
    \end{subfigure}%%
    \hfill
    \begin{subfigure}[b]{0.166\linewidth}
    \includegraphics[width=\linewidth]{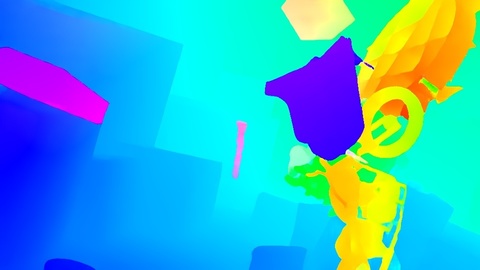}
    \end{subfigure}%%
    \hfill
    \begin{subfigure}[b]{0.166\linewidth}
    \includegraphics[width=\linewidth]{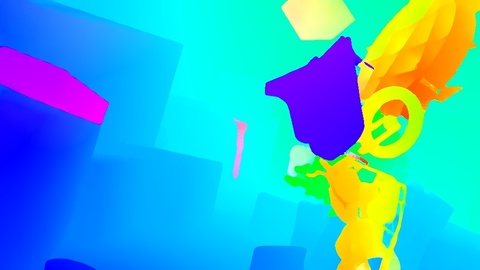}
    \end{subfigure}%%
    \hfill
    \begin{subfigure}[b]{0.166\linewidth}
    \includegraphics[width=\linewidth]{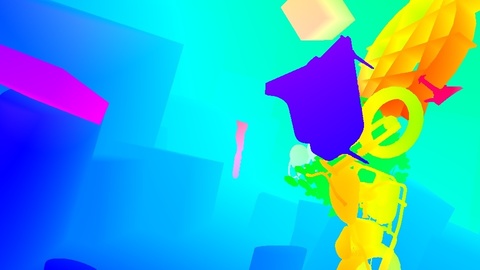}
    \end{subfigure}%%
    \hfill
    
    \begin{subfigure}[b]{0.166\linewidth}
    \includegraphics[width=\linewidth]{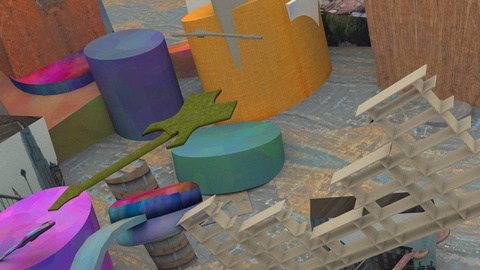}
    \end{subfigure}%%
    \hfill
    \begin{subfigure}[b]{0.166\linewidth}
    \includegraphics[width=\linewidth]{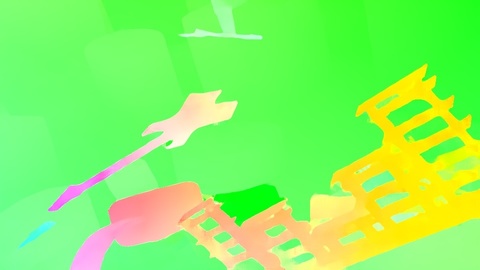}
    \end{subfigure}%%
    \hfill
    \begin{subfigure}[b]{0.166\linewidth}
    \includegraphics[width=\linewidth]{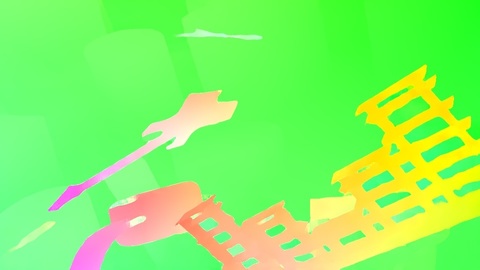}
    \end{subfigure}%%
    \hfill
    \begin{subfigure}[b]{0.166\linewidth}
    \includegraphics[width=\linewidth]{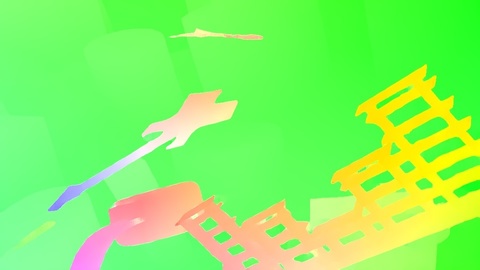}
    \end{subfigure}%%
    \hfill
    \begin{subfigure}[b]{0.166\linewidth}
    \includegraphics[width=\linewidth]{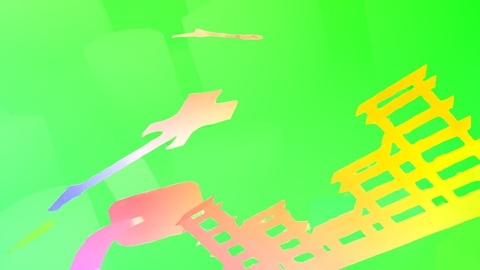}
    \end{subfigure}%%
    \hfill
    \begin{subfigure}[b]{0.166\linewidth}
    \includegraphics[width=\linewidth]{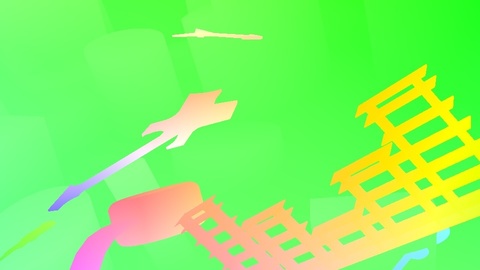}
    \end{subfigure}%%
    \hfill
    \caption{A visual ablation on our multi-stage fusion pipeline. Fusing pyramid features helps to recover the structure and boundary of complex objects. Fusing the cost volume enables the network to capture small and fast-moving objects. Fusing the features of the flow decoder improves the performance on complex scenes where objects overlap.}
    \label{fig:multi-stage-ablation}
\end{figure*}

\begin{figure*}
    \includegraphics[width=\linewidth]{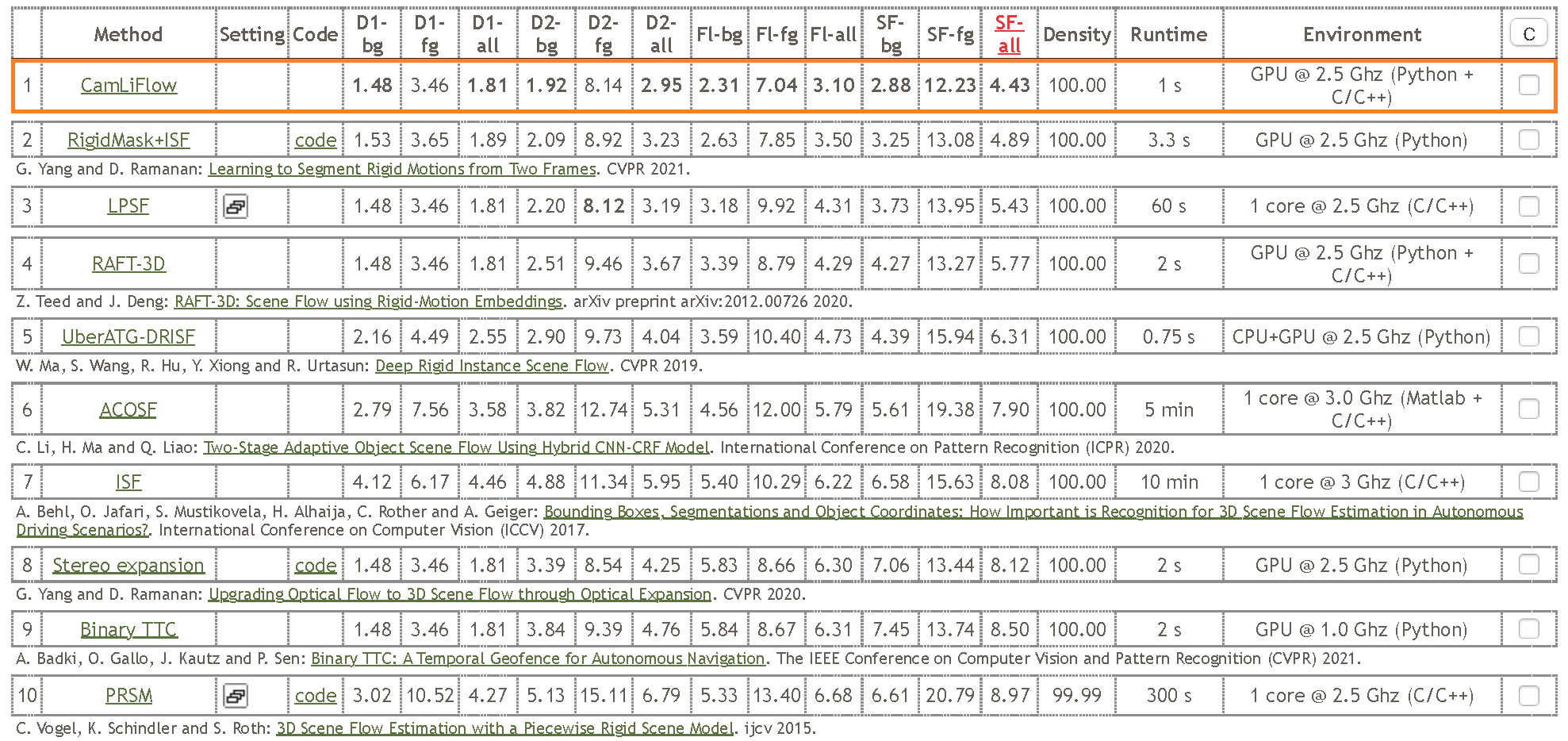}
    \caption{Screenshot of the KITTI Scene Flow 2015 benchmark on November 21th, 2021.}
    \label{fig:kitti-sf-screenshot}
\end{figure*}

\paragraph{KITTI.} Since the website of KITTI only visualizes a limited number of samples on the test set, we provide additional qualitative examples on the validation split in Fig. \ref{fig:more-kitti-val}. Row (a), (b), and (c) are respectively estimated by GA-Net \cite{zhang2019ganet}, our point branch, and our image branch before the rigidity refinement step. Our model can estimate the foreground motions and most background motions accurately as shown in Row (e). It fails when objects are entirely occluded (see the last column). By applying the rigidity refinement for the background (the segmentation masks are estimated by DDRNet-Slim \cite{hong2021ddrnet} and are visualized in Row (d)), our model can estimate both foreground and background motions accurately, as shown in Row (f).

\paragraph{Ablations for Multi-stage Fusion Pipeline.} CamLiFlow performs feature fusion in a multi-stage manner. In the main paper, we confirm the effectiveness of each stage by quantitative results in Tab. \ref{tab:ablation-multi-stage}. Here, we provide a more detailed qualitative analysis. In Fig. \ref{fig:multi-stage-ablation}, the second column denotes a variant of our model where no fusion connection exists between the two branches. The third column only fuses the pyramid feature, while the fourth column further fuses the cost volume. The fifth column denotes our full model which performs feature fusion at all three stages.

As we can see, fusing pyramid features makes the structure of the objects clearer, since point pyramid encodes geometric information which helps to recover the shape of complex objects. Fusing the cost volume enables the model to capture small and fast-moving objects, since the point-based 3D cost volume searches for a dynamic range of neighborhoods and can be complementary to the 2D cost volume which searches for a fixed range. Fusing the features of the flow decoder improves overall performance, especially on complex scenes where objects overlap.

\section{Screenshot on Online Benchmark}

We submit our method to the website of KITTI and rank first on the KITTI Scene Flow 2015 benchmark. The screenshot of the leaderboard is shown in Fig. \ref{fig:kitti-sf-screenshot}.

\end{document}